\documentclass[11pt, a4paper]{preprint}

\usepackage[numbers,sort&compress]{natbib}
\usepackage{xspace}

\theoremstyle{plain}

\newtheorem*{proposition*}{Proposition}

\theoremstyle{definition}

\theoremstyle{definition}

\def\eqref#1{equation~\ref{#1}}

\usepackage{graphicx}           
\usepackage{tikz}               
\usepackage[edges]{forest}      

\usepackage{url}                
\usepackage{xurl}               

\usepackage{array}              
\usepackage{longtable}          
\usepackage{multirow}           
\usepackage{makecell}           
\usepackage{ragged2e}           

\usepackage{mathtools}          
\usepackage{nicefrac}           

\usepackage{algorithm}          
\usepackage{algorithmicx}       
\usepackage{algpseudocode}      
\usepackage{listings}           

\usepackage{subcaption}         
\usepackage{wrapfig}            
\usepackage{placeins}           
\usepackage[export]{adjustbox}  

\usepackage{xspace}             
\usepackage[normalem]{ulem}     
\usepackage{CJKutf8}            

\usepackage[tikz]{bclogo}       
\usepackage[framemethod=tikz]{mdframed} 

\usepackage{lipsum}             
\usepackage{tocloft}            
\usepackage{afterpage}          
\usepackage{bbding}             
\usepackage{epigraph}           
\usepackage{minitoc}            
\usepackage{multicol}           
\usepackage{textgreek}          

\newcolumntype{P}[1]{>{\RaggedRight\arraybackslash}p{#1}}

\definecolor{uclablue}{RGB}{39, 116, 174}
\definecolor{bigaired}{RGB}{156, 0, 0}
\definecolor{myblue}{HTML}{598BE7}
\definecolor{mildblue}{RGB}{31,119,180}
\definecolor{sectionblue}{RGB}{70, 130, 180}
\definecolor{methodblue}{RGB}{0, 150, 136}
\definecolor{bgblue}{RGB}{245,243,253}
\definecolor{ttblue}{RGB}{91,194,224}
\definecolor{mygreen}{rgb}{0.64, 0.56, 0.88}
\definecolor{myyellow}{rgb}{0.68, 0.6, 0.1}
\definecolor{fancygreen}{rgb}{0.33, 0.68, 0.20}
\definecolor{salmon}{rgb}{0.94, 0.52, 0.49}
\definecolor{tablegreen}{rgb}{0.82, 0.94, 0.75}
\definecolor{tableblue}{rgb}{0.81, 0.90, 0.94}
\definecolor{tablered}{rgb}{0.97, 0.85, 0.85}
\definecolor{tableorange}{rgb}{0.96, 0.85, 0.81}
\definecolor{myorange}{rgb}{1.0, 0.49, 0.0}
\definecolor{casepurple}{RGB}{112,96,200}
\definecolor{tlgreen}{rgb}{0.33, 0.68, 0.20}
\definecolor{darkgreen}{RGB}{0,100,0}
\definecolor{darkred}{RGB}{200, 0, 0}
\definecolor{customyellow}{HTML}{FFFACD}
\definecolor{refinegreen}{RGB}{0, 128, 75}
\definecolor{scoregreen}{RGB}{34, 139, 34}
\definecolor{hidden-blue}{RGB}{194,232,247}
\definecolor{hidden-black}{RGB}{20,68,106}
\definecolor{yes}{HTML}{C6EFCE}
\definecolor{no}{HTML}{FFC7CE}
\definecolor{partial}{HTML}{FFEB9C}
\definecolor{external}{HTML}{D9E1F2}
\definecolor{hdr}{HTML}{F2F2F2}
\definecolor{GRPOrow}{gray}{0.96}
\definecolor{FlowRLrow}{RGB}{225,236,255}
\definecolor{FlowBlue}{RGB}{80,120,210}
\definecolor{GRPOGray}{gray}{0.35}

\hypersetup{
    colorlinks=true, 
    citecolor=uclablue, 
    linkcolor=bigaired,
    urlcolor=darkblue
}

\setlist[itemize]{leftmargin=20pt, noitemsep, topsep=0pt}


\NewDocumentCommand{\kaiyan}{mO{}}{\textcolor{purple}{\textsuperscript{\textit{kaiyan}}\textsf{\textbf{\small[#1]}}}}
\NewDocumentCommand{\yuxin}{mO{}}{\textcolor{cyan}{\textsuperscript{\textit{yuxin}}\textsf{\textbf{\small[#1]}}}}
\NewDocumentCommand{\bx}{mO{}}{\textcolor{green}{\textsuperscript{\textit{bx}}\textsf{\textbf{\small[#1]}}}}
\NewDocumentCommand{\at}{mO{}}{\textcolor{red}{\textsuperscript{\textit{AT}}\textsf{\textbf{\small[#1]}}}}
\NewDocumentCommand{\re}{mO{}}{\textcolor{blue}{\textsuperscript{\textit{RE}}\textsf{\textbf{\small[#1]}}}}
\NewDocumentCommand{\ybsun}{mO{}}{\textcolor{magenta}{\textsuperscript{\textit{youbang}}\textsf{\textbf{\small[#1]}}}}
\NewDocumentCommand{\runze}{mO{}}{\textcolor{orange}{\textsuperscript{\textit{runze}}\textsf{\textbf{\small[#1]}}}}
\NewDocumentCommand{\add}{mO{}}{\textcolor{darkgreen}{\textsuperscript{\textit{Maybe Consider Discuss}}\textsf{\textbf{[#1]}}}}

\newcommand{\cmark}{\textcolor{darkgreen}{\boldmath$\checkmark$}}
\newcommand{\xmark}{\textcolor{darkred}{\boldmath$\times$}}

\newif\ifshowrev
\showrevfalse
\definecolor{revcolor}{RGB}{200,25,25}
\newcommand{\rev}[1]{\ifshowrev\textcolor{revcolor}{#1}\else#1\fi}

\newenvironment{itemize*}%
 {\leftmargini=10pt\begin{itemize}%
  \setlength{\itemsep}{0pt}%
  \setlength{\parskip}{0pt}%
  }%
 {\end{itemize}}

\newenvironment{enumerate*}%
 {\begin{enumerate}%
  \setlength{\itemsep}{0pt}%
  \setlength{\parskip}{0pt}}%
 {\end{enumerate}}

\newcommand{\cellstatus}[1]{%
  \begingroup
  \StrTrim{#1}[\statusval]%
  \IfStrEq{\statusval}{Yes}{\cellcolor{yes}\cmark}{}%
  \IfStrEq{\statusval}{No}{\cellcolor{no}\xmark}{}%
  \IfBeginWith{\statusval}{Yes (}{\cellcolor{yes}\cmark~\textit{\statusval\unskip}}{}%
  \IfStrEq{\statusval}{Partial}{\cellcolor{partial}\textbf{Partial}}{}%
  \IfStrEq{\statusval}{External}{\cellcolor{external}\textbf{External}}{}%
  \endgroup
}

\newtcolorbox{myboxi}[1][]{
  breakable,
  title=#1,
  colback=red!5,
  colbacktitle=red!5,
  coltitle=black,
  fonttitle=\bfseries,
  bottomrule=0pt,
  toprule=0pt,
  leftrule=2pt,
  rightrule=2pt,
  titlerule=0pt,
  arc=0pt,
  outer arc=0pt,
  colframe=red,
}

\newtcolorbox{myboxnote}[1][]{
  breakable,
  title=#1,
  colback=orange!0,
  colbacktitle=orange!0,
  coltitle=black,
  fonttitle=\bfseries,
  bottomrule=0pt,
  toprule=0pt,
  leftrule=2pt,
  rightrule=2pt,
  titlerule=0pt,
  arc=0pt,
  outer arc=0pt,
  colframe=orange,
}

\newtcolorbox{myboxii}[1][]{
  breakable,
  freelance,
  title=#1,
  colback=white,
  colbacktitle=white,
  coltitle=black,
  fonttitle=\bfseries,
  bottomrule=0pt,
  boxrule=0pt,
  colframe=white,
  overlay unbroken and first={
  \draw[red!75!black,line width=3pt]
    ([xshift=5pt]frame.north west) -- 
    (frame.north west) -- 
    (frame.south west);
  \draw[red!75!black,line width=3pt]
    ([xshift=-5pt]frame.north east) -- 
    (frame.north east) -- 
    (frame.south east);
  },
  overlay unbroken app={
  \draw[red!75!black,line width=3pt,line cap=rect]
    (frame.south west) -- 
    ([xshift=5pt]frame.south west);
  \draw[red!75!black,line width=3pt,line cap=rect]
    (frame.south east) -- 
    ([xshift=-5pt]frame.south east);
  },
  overlay middle and last={
  \draw[red!75!black,line width=3pt]
    (frame.north west) -- 
    (frame.south west);
  \draw[red!75!black,line width=3pt]
    (frame.north east) -- 
    (frame.south east);
  },
  overlay last app={
  \draw[red!75!black,line width=3pt,line cap=rect]
    (frame.south west) --
    ([xshift=5pt]frame.south west);
  \draw[red!75!black,line width=3pt,line cap=rect]
    (frame.south east) --
    ([xshift=-5pt]frame.south east);
  },
}

\tcbset{
  takeawaysbox/.style={
    title=Takeaways,
    colback=lightblue!80,
    colframe=black,
    fonttitle=\bfseries\small,
    coltitle=white,
    colbacktitle=black,
    enhanced,
    attach boxed title to top left={xshift=2.5mm,yshift=-2.5mm},
    boxed title style={rounded corners, size=small, colframe=black, colback=black},
    width=\linewidth,
    arc=3.5mm
  }
}

\mdfdefinestyle{mystyle}{%
  rightline=true,
  innerleftmargin=10,
  innerrightmargin=10,
  outerlinewidth=3pt,
  topline=false,
  rightline=true,
  bottomline=false,
  skipabove=\topsep,
  skipbelow=\topsep
}

\tikzset{%
    every node/.style={font=\tiny},
    parent/.style =          {align=center,text width=2cm,rounded corners=3pt, line width=0.3mm, fill=gray!10,draw=gray!80},
    child/.style =           {align=center,text width=2.0cm,rounded corners=3pt, fill=blue!10,draw=blue!80,line width=0.3mm},
    grandchild/.style =      {align=center,text width=2cm,rounded corners=3pt},
    greatgrandchild/.style = {align=center,text width=1.5cm,rounded corners=3pt},
    greatgrandchild2/.style = {align=center,text width=1.5cm,rounded corners=3pt},    
    referenceblock/.style =  {align=center,text width=1.5cm,rounded corners=2pt},
    pretrain/.style =           {align=center,text width=2.0cm,rounded corners=3pt, fill=blue!10,draw=blue!80,line width=0.3mm},   
    pretrain_work/.style =           {align=center, text width=8.5cm,rounded corners=3pt, fill=blue!10,draw=blue!0,line width=0.3mm},  
    template/.style =           {align=center,text width=2.0cm,rounded corners=3pt, fill=red!10,draw=red!80,line width=0.3mm},   
    template_work/.style =           {align=center,text width=8.5cm,rounded corners=3pt, fill=red!10,draw=red!0,line width=0.3mm},    
    answer/.style =           {align=center,text width=2.0cm,rounded corners=3pt, fill= cyan!10,draw= cyan!80,line width=0.3mm},   
    answer_work/.style =           {align=center,text width=8.5cm,rounded corners=3pt, fill= cyan!10,draw= cyan!0,line width=0.3mm},      
    multiple/.style =           {align=center,text width=2.0cm,rounded corners=3pt, fill= orange!10,draw= orange!80,line width=0.3mm},   
    multiple_work/.style =           {align=center,text width=8.5cm,rounded corners=3pt, fill= orange!10,draw= orange!0,line width=0.3mm},        
    tuning/.style =           {align=center,text width=2.0cm,rounded corners=3pt, fill= magenta!10,draw= magenta!80,line width=0.3mm},   
    tuning_work/.style =           {align=center,text width=8.5cm,rounded corners=3pt, fill= magenta!10,draw= magenta!0,line width=0.3mm},          
}

\newcommand{\lstbg}[3][0pt]{{\fboxsep#1\colorbox{#2}{\strut #3}}}

\lstdefinelanguage{diff}{
  basicstyle=\ttfamily\small,
  morecomment=[f][\lstbg{red!20}]-,
  morecomment=[f][\lstbg{green!20}]+,
}

\lstdefinelanguage{diffpython}{
  language=diff,
  morekeywords={def, if, else, for, while, return, import, from, as, class, with, try, except, finally, raise, lambda, and, or, not, in, is, None, True, False},
  morecomment=[l]{\#},
  morestring=[b]",
  morestring=[b]',
}

\definecolor{gaincolor}{RGB}{0,128,0}
\definecolor{dropcolor}{RGB}{190,40,40}
\definecolor{neutralcolor}{RGB}{110,110,110}

\setheadertext{D\&CG Lab}

\correspondingemail{$^\ddagger$ Corresponding authors. \\
\emailicon{} changjianz@student.unimelb.edu.au, \quad negin.yousefpour@unimelb.edu.au \\
\faGithub\ \href{https://github.com/Data-Driven-Computational-Geotechnics/TRACE}{Data-Driven-Computational-Geotechnics/TRACE}}

\renewcommand{\titlefont}{\color{darkblue}\normalfont\bfseries\fontsize{17}{20}\selectfont}

\renewcommand{\footerfont}{\color{black}\normalfont\fontsize{10}{12}\selectfont}

\fancypagestyle{firststyle}{%
  \fancyhead[L]{%
    \includegraphics[height=36pt]{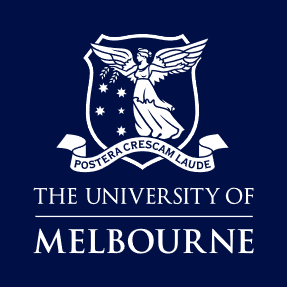}\hspace{12pt}%
    \includegraphics[height=36pt]{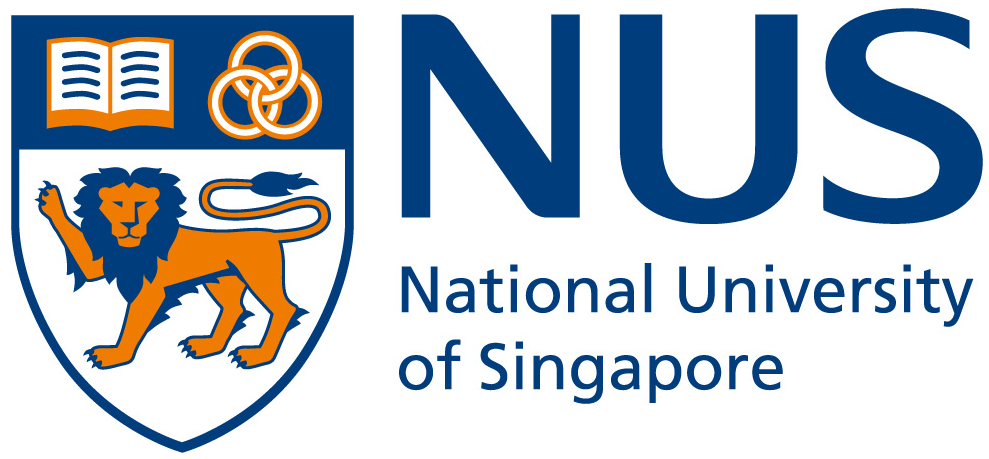}%
    \includegraphics[height=36pt]{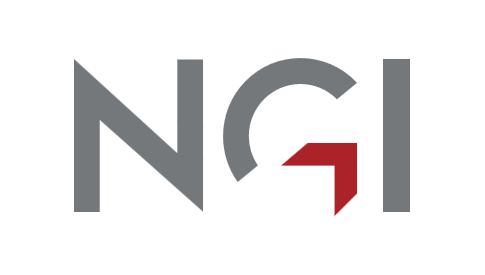}%
  }%
  \fancyhead[R]{%
    \includegraphics[height=36pt]{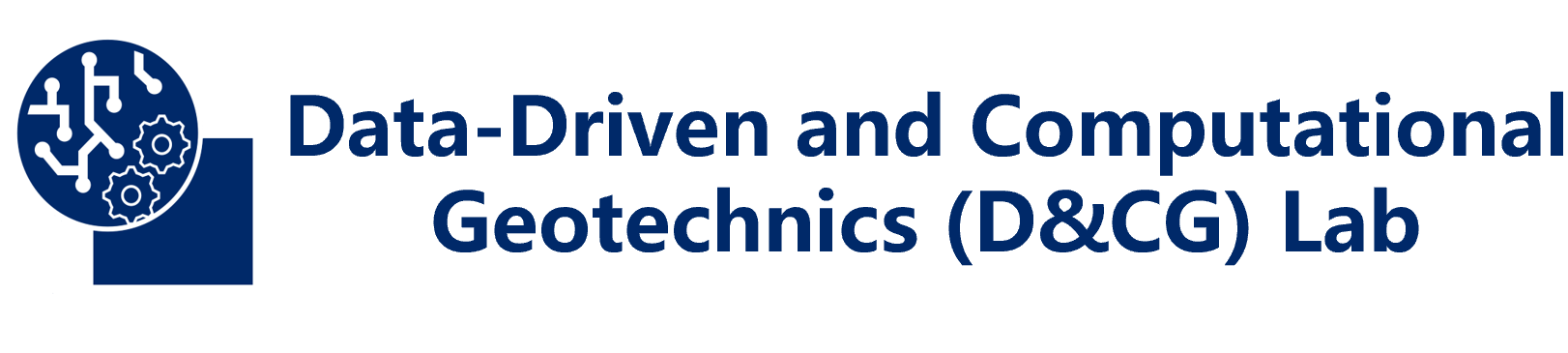}%
    \ifthenelse{\boolean{confidential}}{\\ \footerfont \internalonly}{}%
  }%
  \fancyhead[C]{}%
  \fancyfoot[L]{\ifthenelse{\boolean{copyright}}{\copyrightext}{}}%
  \fancyfoot[C]{%
    \ifthenelse{\boolean{confidential}}{%
      \ifdefined\reportnumber
        \if\relax\the\reportnumber\relax\else
          {\footerfont\itshape D\&CG Technical Report \the\reportnumber}%
        \fi
      \fi
    }{}%
  }%
  \fancyfoot[R]{\footerfont\thepage}%
}

\newcommand{\gawidth}{\linewidth}
\let\preprintabscontent\abscontent
\renewcommand{\abscontent}{%
  \noindent\includegraphics[width=\gawidth]{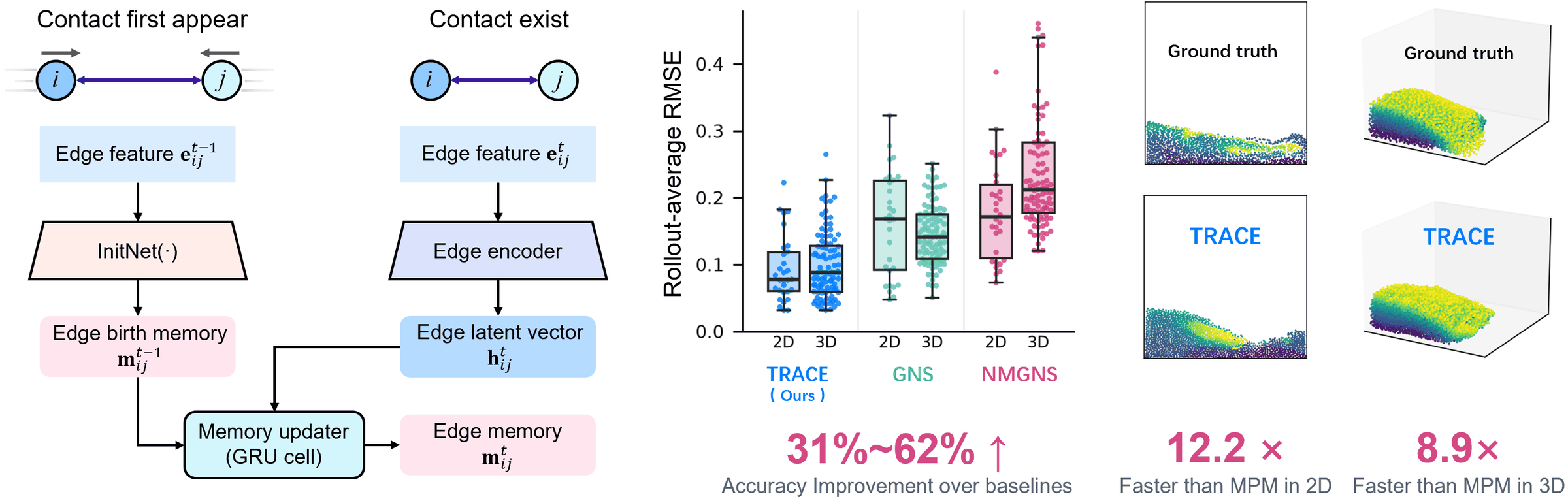}%
  \vskip9pt
  \preprintabscontent
}

\title{TRACE: A spatiotemporal contact memory graph network simulator for granular dynamics}
\setheadertitle{TRACE: A spatiotemporal contact memory graph network simulator for granular dynamics}

\author{%
  {\Authfont Changjian Zhou$^{a,\ddagger}$, Negin Yousefpour$^{a,\ddagger}$, Jie Qi$^{a}$, Junfeng Fang$^{b}$, Guillermo A. Narsilio$^{a}$, Hans Petter Jostad$^{c}$}\\
  $^a$ Faculty of Engineering and Information Technology, The University of Melbourne, Victoria 3010, Australia\\
  $^b$ School of Computing, National University of Singapore, Singapore 117417, Singapore\\
  $^c$ Norwegian Geotechnical Institute, Oslo 0855, Norway
}

\begin{document}

\begin{abstract}
Learned graph simulators provide an efficient alternative to high-fidelity solvers for granular dynamics. However, granular motion depends strongly on inter-granular contact history, which is difficult to preserve when particle contacts form, break, and rearrange. Existing simulators mainly store temporal information in node features or node-level memory. Here we introduce TRACE (spatio\textbf{T}emporal memo\textbf{R}y \textbf{A}cross \textbf{C}ontact \textbf{E}dge), a graph-network simulator that stores interaction history directly on contact edges. Each edge maintains a persistent memory updated by attention-based message passing and a gated recurrent unit, while an edge-identity dictionary preserves this memory as the contact graph changes. A physics-structured decoder predicts inter-granular normal and tangential contact forces, enforces the Coulomb friction limit, and applies equal-and-opposite internal forces. The model is trained with single-step pretraining followed by autoregressive rollout fine-tuning. We evaluate TRACE on 2D and 3D granular column-collapse benchmarks. In both cases, TRACE produces stable, physically consistent long-horizon rollouts, closely reproducing the final deposit geometry and the kinetic energy released during collapse. Compared with graph network simulator (GNS) and node-memory graph neural simulator (NMGNS), TRACE reduces long-rollout position error by 31--62\% and final-deposit error by 58--89\% across the two benchmarks, while using fewer parameters and maintaining near-zero particle interpenetration. TRACE also achieves 12.2$\times$ and 8.9$\times$ speedups over the material point method (MPM) reference solver in 2D and 3D, respectively. Our code is available at \url{https://github.com/Data-Driven-Computational-Geotechnics/TRACE}.


\end{abstract}

\maketitle


\clearpage

\section*{List of notations}\label{list-of-notations}

\begin{longtable}{>{\small}p{0.22\linewidth} >{\small}p{0.72\linewidth}}
\toprule
Symbol & Description \\
\midrule
\endhead
\bottomrule
\endlastfoot
$N$, $d$ & Number of particles and spatial dimension \\
$i$, $j$ & Particle indices of a contact pair, ordered so that $i < j$ \\
$p$ & Particle index in the attention pooling \\
$t$ & Time step index \\
$\mathbf{x}_i^t$, $\mathbf{v}_i^t$ & Position and velocity of particle $i$ at step $t$ \\
$r_i$, $c_i$, $m_i$ & Radius, type label, and mass of particle $i$ \\
$\mathbf{X}^t$ & System configuration at step $t$ collecting the states of all particles \\
$s_\theta$ & Learned single-step transition function \\
$\mathcal{G}^t = (\mathcal{V}, \mathcal{E}^t)$ & Contact graph at step $t$ \\
$\mathcal{V}$ & Node set (all $N$ particles, fixed throughout) \\
$\mathcal{E}^t$ & Edge set (rebuilt each step $t$) \\
$|\mathcal{E}^t|$ & Number of active contacts at step $t$ \\
$\alpha$ & Skin (pre-contact) factor in the contact threshold \\
$\mathbf{n}_i^t$, $\mathbf{e}_{ij}^t$ & Node and edge input feature vectors \\
$\operatorname{Enc}^{\mathcal{V}}$, $\operatorname{Enc}^{\mathcal{E}}$ & Node and edge encoder MLPs \\
$\| \cdot \|_2$, $[\,\cdot\,;\,\cdot\,]$ & Euclidean norm and vector concatenation \\
$\mathbf{h}_i^{(0)}$ & Initial node latent before message passing \\
$\mathbf{h}_{ij}^t$ & Edge latent encoding the instantaneous contact geometry \\
$D$ & Latent dimension of node and edge representations \\
$\mathbf{m}_{ij}^t$ & Persistent memory of contact ($i$, $j$) at step $t$. \\
$\tilde{\mathbf{m}}_{ij}^t$ & Pre-update memory state, inherited from step $t-1$ or generated for a new contact \\
InitNet & Trainable network generating the starting memory of a new contact \\
GRU & Gated recurrent unit updating the contact memory \\
$\mathbf{u}_{ij}^t$ & GRU update input, concatenation of the edge latent and the spatial context \\
$\mathbf{s}_{ij}^t$ & Spatial context of contact ($i$, $j$), mean of its two endpoint contexts \\
$d_m$ & Dimension of the contact memory vectors \\
$\mathbf{M}^t$ & Memory matrix stacking all contact memories at step $t$ \\
$k$ & Row index in the memory matrix, also the message-passing round index \\
$\operatorname{eid}(i, j)$ & Time-invariant contact identifier $\operatorname{eid}(i, j) = i \cdot N + j$ \\
$\mathcal{D}^t$ & Identity dictionary mapping $\operatorname{eid}(i, j)$ to a row of $\mathbf{M}^t$ \\
$\mathcal{E}(p)$ & Set of contacts incident to particle $p$ \\
$\mathbf{k}_{ij}^{t}$, $\mathbf{v}_{ij}^{t}$ & Attention key and value of contact ($i$, $j$) \\
$\mathbf{W}_{\mathbf{k}}$, $\mathbf{W}_{\mathbf{v}}$ & Learnable key and value projections \\
$\mathbf{q}$ & Learned query vector shared by all contacts \\
$\alpha_{p,ij}^{t}$ & Attention weight of contact ($i$, $j$) in the pool of particle $p$ \\
$\mathbf{c}_{p}^{t}$ & Pooled contact context of particle $p$ \\
$K$ & Number of message-passing rounds \\
$\mathbf{h}_{i}^{(k)}$, $\mathbf{h}_{i}^{(K)}$ & Node latent of particle $i$ after $k$ rounds and the final $K$ rounds \\
$\psi^{(k)}$, $\mathcal{X}^{(k)}$ & Message and node-update MLPs of round $k$ \\
$\mathbf{g}_{ij}^{(k)}$, $\mathbf{g}_{i}^{(k)}$ & Message from particle $j$ to $i$ at round $k$ and its sum over the neighborhood \\
$\mathcal{N}(i)$ & Contact neighborhood of particle $i$ \\
$\text{Dec}^{\mathcal{V}}$, $\text{Dec}^{\mathcal{E}}$ & Node and edge decoder networks \\
$\mathbf{a}_{i}^{t}$, $\mathbf{a}_{i}^{\text{ext}}$, $\mathbf{a}_{i}^{\text{int}}$ & Predicted, external and internal (contact) acceleration of particle $i$ \\
$F_{n}^{ij}$ & Normal force magnitude of contact ($i$, $j$), constrained non-negative \\
$\tilde{\mathbf{F}}_{t}^{ij}$, $\mathbf{F}_{t}^{ij}$ & Raw tangential force and its projection onto the Coulomb cone \\
$\mu_{ij}$ & Learned per-contact friction coefficient \\
$\mathbf{f}_{ij}$ & Total contact force on particle $i$ from particle $j$ \\
$\hat{\mathbf{n}}_{ij}$ & Unit contact normal from $j$ to $i$ \\
$\epsilon$ & Small constant for numerical stability \\
$\Delta t$ & Simulation time step \\
$\delta_{ij}$ & Overlap between particles $i$ and $j$ in the non-penetration projection \\
$\mathbf{a}_i^{\star,t}$ & Target acceleration from finite differences of reference velocities \\
$\ell^t$ & Per-step pretraining loss on normalized accelerations \\
$\operatorname{norm}(\cdot)$ & Per-channel normalizer fitted on the training set \\
$\mathcal{L}$, $T_{\text{W}}$ & Pretraining window loss and window length \\
$t_0$ & Start step of a training window or of the supervised segment \\
$\mathcal{L}_{\text{ft}}$, $T_{\text{ft}}$ & Fine-tuning rollout loss and supervised segment length \\
$\operatorname{RMSE}(t)$ & Position root mean square error at rollout step $t$ \\
$\hat{\mathbf{x}}_i^t$ & Reference position of particle $i$ in the evaluation metrics \\
$\epsilon_{\text{dep}}$ & Normalized deposit error combining runout and height discrepancies \\
$L$, $H$, $\hat{L}$, $\hat{H}$ & Predicted and reference runout length and deposit height \\
$L_{\text{box}}$ & Domain size normalizing the deposit error \\
$E_k(t)$ & Kinetic energy at step $t$ with unit particle mass \\
\end{longtable}
\setcounter{table}{0}

\section{Introduction}\label{introduction}

Granular materials are ubiquitous in nature and engineering, and their dynamics govern processes ranging from landslides and debris flows to industrial powder handling. Particle-based numerical methods, including the discrete element method (DEM) \citep{cundall1979discrete}, smoothed particle hydrodynamics (SPH) \citep{gingold1977smoothed,lucy1977numerical}, and the material point method (MPM) \citep{sulsky1994particle,sulsky1995application}, have become indispensable for studying complex phenomena involving large deformation, fragmentation, impact, and granular flow \citep{soga2016trends}. However, high-fidelity simulations require resolving numerous particle interactions with sufficiently small time steps to maintain numerical stability, resulting in substantial computational cost \citep{guo2015discrete}. In engineering applications, this burden is further increased by the need to recalibrate contact parameters for each new material \citep{coetzee2017review}. To reduce these costs, growing research efforts have focused on machine learning-based physical simulators, which approximate the underlying dynamics from data while seeking to preserve acceptable accuracy and substantially improve computational efficiency \citep{lino2023current,lu2021machine}.

\begin{figure}[htbp]
\centering
\includegraphics[width=\linewidth]{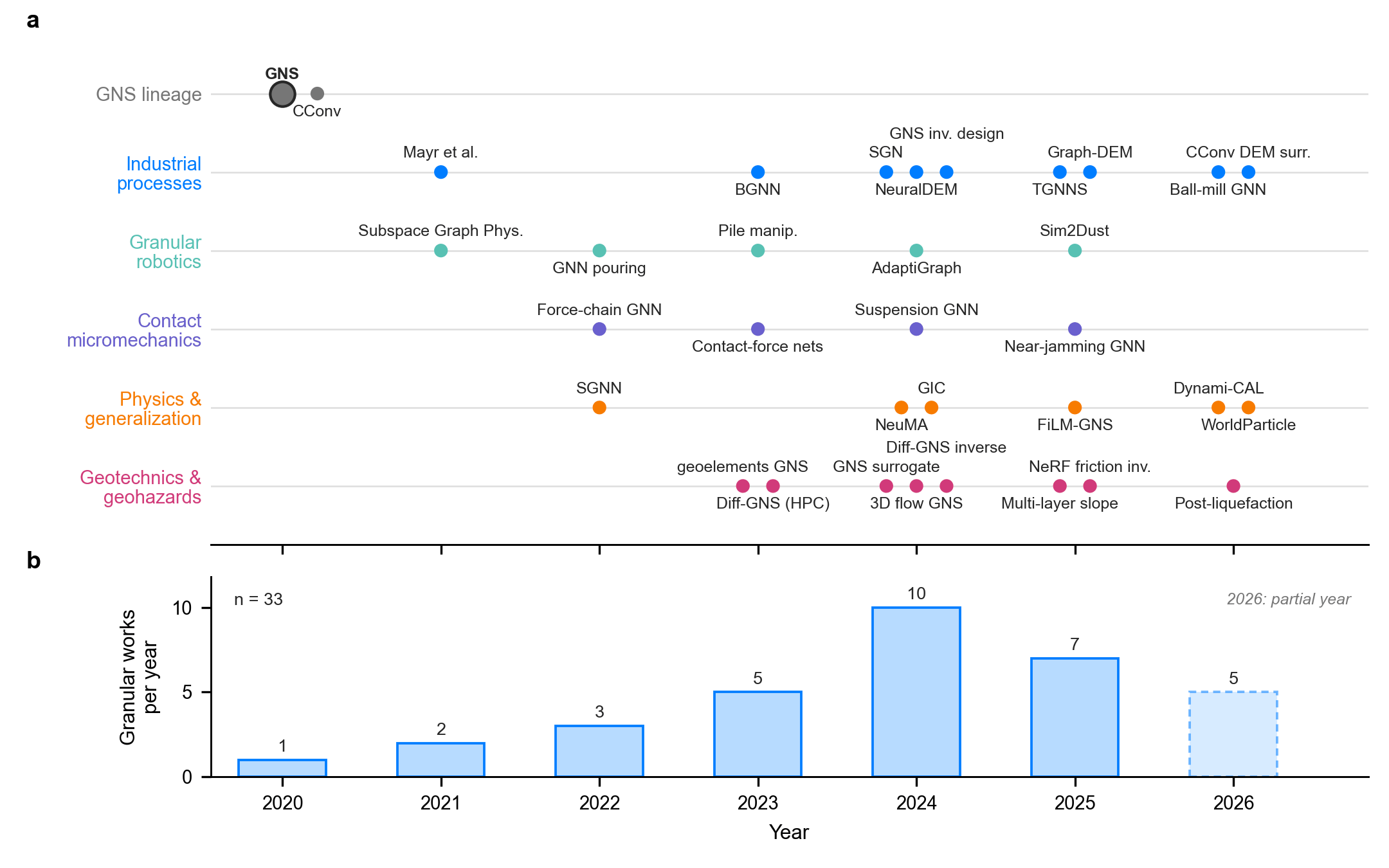}
\caption{The rise of learned granular simulators: representative works across five research threads and related works per year (2020--2026). Works are grouped by research thread: GNS lineage \citep{sanchezgonzalez2020learning,ummenhofer2020lagrangian}; Industrial processes \citep{mayr2021learning,mayr2023bgnn,li2024sgn,alkin2024neuraldem,jiang2024inversedesign,zhao2025physicalinformationflowconstrained,li2025graphdem,garridonunez2026ballmill,le2026cconvdem}; Granular robotics \citep{haeri2024subspace,tuomainen2022manipulation,wang2023pile,zhang2024adaptigraph,orsula2025sim2dust}; Contact micromechanics \citep{mandal2022forcechain,li2023contactforce,aminimajd2024suspension,aminimajd2025nearjamming}; Physics \& generalization \citep{han2022learning,sharma2026physicsinformed,cao2024neuma,cai2024gic,manoharan2025filmgns,wang2026worldparticle}; Geotechnics \& geohazards \citep{kumar2023gnsjoss,kumar2023differentiable,choi2024surrogate,choi2024threedimensional,choi2024inverse,choi2026multilayer,hsiao2025nerf,choi2026postliquefaction}.}
\label{fig:overview}
\end{figure}

Among existing machine-learning architectures, graph neural networks (GNNs) have attracted particular attention because they naturally represent the relational structure of physical systems \citep{battaglia2016interaction,battaglia2018relational}. By representing particles as nodes and their contacts as edges, GNNs propagate information and update system states directly on interaction graphs \citep{sanchezgonzalez2018graph}. The graph network simulator (GNS) introduced by Sanchez-Gonzalez et al. \citep{sanchezgonzalez2020learning} established the widely adopted Encoder--Processor--Decoder paradigm and demonstrated strong performance across a broad range of particle-based systems. Building on this framework, subsequent studies have developed simulators for unstructured meshes \citep{pfaff2021learning}, particle-based models for multi-material systems and fluids \citep{li2019learning,ummenhofer2020lagrangian}, hierarchical architectures that scale to large systems \citep{wu2022learning}, equivariant message-passing networks that preserve rotational and translational symmetries \citep{satorras2021en,brandstetter2022geometric,huang2022equivariant,han2022learning}, and physics-informed models that incorporate conservation laws and geometric constraints \citep{greydanus2019hamiltonian,cranmer2020lagrangian,rubanova2022constraintbased,allen2023learning,prantl2022guaranteed,sharma2026physicsinformed}. Together, these advances have substantially improved the accuracy, generalization, and physical consistency of learned simulators. Figure~\ref{fig:overview} summarizes some representative works from recent years.

Despite this progress in spatial representation learning and physics-based inductive biases, effectively modeling history-dependent behavior remains an open challenge. In many physical systems, future evolution depends not only on the current state but also on the sequence of past interactions. Existing GNN simulators typically represent this history at the node level. For example, GNS includes multiple past velocity frames as node features to provide short-term dynamical context \citep{sanchezgonzalez2020learning}. More recently, the temporal graph neural network simulator proposed by Zhao et al. \citep{zhao2025physicalinformationflowconstrained} maintains a recurrent hidden state for each node, allowing interaction histories to be encoded over longer temporal horizons.

This node-centric design is also prevalent in temporal graph learning beyond physical simulation. Methods such as temporal graph networks \citep{rossi2020temporal}, JODIE \citep{kumar2019predicting}, DyRep \citep{trivedi2019dyrep}, and TGAT \citep{xu2020inductive} associate each node in a dynamic graph with a memory vector that is updated at every interaction event, as reviewed in \citep{kazemi2020representation}. Earlier studies also applied recurrent units to spatiotemporal graphs with fixed topology \citep{jain2016structuralrnn}, including the recurrent graph network introduced by Sanchez-Gonzalez et al. \citep{sanchezgonzalez2018graph}. These approaches have shown that explicit temporal states can improve the representation of long-term dependencies and enhance prediction stability. A complementary line of research addresses error accumulation during long autoregressive rollouts, where small single-step errors compound and gradually drive predictions away from the training distribution. Representative strategies include training-time noise injection \citep{sanchezgonzalez2020learning}, scheduled sampling \citep{bengio2015scheduled}, differentiable multi-step unrolling \citep{um2020solverintheloop,brandstetter2022message}, and related rollout-stabilization techniques \citep{stachenfeld2022learned}.

For granular materials, however, many of the most important history-dependent mechanisms arise from the evolution of inter-particle contacts. Consider frictional contact. While the normal contact force is primarily determined by the current overlap, the tangential force depends on the tangential displacement accumulated since contact initiation and on the evolving sliding direction \citep{mindlin1953elastic}. Classical contact models, prevalent in DEM simulations, therefore integrate a tangential spring over the lifetime of each contact and constrain the resulting force using the Coulomb friction cone \citep{cundall1979discrete,luding2008cohesive}. The corresponding history variables are properties of the contact between a pair of particles.

More broadly, many constitutive responses and dissipation mechanisms in granular systems are governed by the evolution of pairwise interactions. Stress is transmitted through strongly heterogeneous force-chain networks \citep{radjai1996force,majmudar2005contact}, while the rheology of dense granular flows is closely linked to local contact conditions \citep{anon2004dense,jop2006constitutive}. At the continuum scale, the mechanical state of a granular material also depends on its loading history, as formalized in critical state soil mechanics \citep{schofield1968critical}. Data-driven constitutive models therefore often employ recurrent architectures such as long short-term memory (LSTM) networks \citep{hochreiter1997long} to capture stress--strain responses along different loading paths \citep{wang2018multiscale,zhang2021application,qu2021towards,karapiperis2021datadriven}. However, these models operate at the material-point level and do not resolve the evolution of individual inter-grain contacts, represented as edges that join nodes that represent grains. Representing history directly at the edge-level may therefore provide a more welcome physics-inspired inductive bias description than conventional node-level memory.

Motivated by this contact-scale origin of history dependence. we propose \textbf{TRACE} (spatio\textbf{T}emporal memo\textbf{R}y \textbf{A}cross \textbf{C}ontact \textbf{E}dge), a graph neural simulator that explicitly models the spatiotemporal memory of pairwise interaction edges. Unlike existing approaches that primarily store historical information in node states, TRACE assigns a persistent memory state to each active edge. This state is initialized when the edge is formed, updated throughout its lifetime, and removed when the edge disappears. A gated recurrent unit (GRU) \citep{cho2014learning} updates the memory over time, while an attention mechanism \citep{vaswani2017attention,velikovi2018graph} aggregates information from neighboring edges that act on the same nodes. Each memory update therefore reflects both the history of the edge itself and its current local multi-edge environment. To support this process on dynamically changing graphs, we introduce a lightweight memory-management mechanism that tracks persistent edge identities and reliably maintains their associated edge states.

TRACE further integrates the edge memory into a physics-structured decoder. For each edge, the decoder predicts normal and tangential force components and applies them to the two interacting particles with equal magnitude and opposite direction, thereby conserving linear momentum by construction. The predicted tangential force is then projected onto the Coulomb friction cone \citep{cundall1979discrete,mindlin1953elastic}. Through this design, the edge memory is given a concrete mechanical role: it serves as a learned representation of the history-dependent tangential state variables used in classical contact models.

This paper is organized as follows. Section~\ref{methodology} presents the architecture of TRACE, including the encoder, contact-edge memory module, message processor, physics-structured decoder, integrator, and two-stage training protocol. Section~\ref{experiments} reports experiments on 2D and 3D granular column collapse. Section~\ref{discussion} compares TRACE with baseline simulators. Section~\ref{conclusion} summarizes conclusions and future work.

\section{Methodology}\label{methodology}

The proposed model, TRACE, extends the graph network-based simulator by attaching a spatiotemporal memory state to each ``contact'' edge. Specifically, TRACE simulates the particle system through five stages: (i) encoding the current particle configuration into a latent contact graph (Section~\ref{encoder}); (ii) retrieving and updating the spatiotemporal memories associated with active contact edges (Section~\ref{edge-memory}); (iii) performing message passing over the latent graph to refine particle representations (Section~\ref{message-processor}); (iv) decoding the refined node representations and contact edge memories into particle accelerations (Section~\ref{physics-structured-decoder}); and (v) integrating the system dynamics while enforcing geometric constraints (Section~\ref{integration}).

\subsection{Problem formulation}\label{problem-formulation}

We consider a system composed of $N$ interacting spherical particles in $d$-dimensional space $d \in \{2, 3\}$. The state of particle $i$ at time step $t$ is described by its position $\mathbf{x}_i^t \in \mathbb{R}^d$ and velocity $\mathbf{v}_i^t \in \mathbb{R}^d$, along with static attributes including radius $r_i$ and type label $c_i$ (where $c_i = 0$ for mobile grains that evolve under the learned dynamics, and $c_i = 1$ for fixed boundary or obstacle particles whose kinematics are prescribed). The overall system configuration is denoted as $\mathbf{X}^t = \left\{ \left( \mathbf{x}_i^t, \mathbf{v}_i^t, r_i, c_i \right) \right\}_{i=1}^{N}$. The goal is to learn a parameterized single-step transition function $s_\theta$ such that, starting from the initial configuration $\mathbf{X}^0$, autoregressively and iteratively applying $s_\theta$ can generate the long-term trajectory of the entire particle system.

We adopt the \textbf{\emph{Encoder}--\emph{Processor--Decoder}} framework. At each simulation step, the system is represented as a graph $\mathcal{G}^t = (\mathcal{V}, \mathcal{E}^t)$ constructed, where the node set $\mathcal{V}$ represents all $N$ particles and remains fixed throughout the simulation, while the edge set $\mathcal{E}^t$ connects each pair of particles whose distance falls below the contact threshold and is reconstructed at every time step as particles move.

The encoder projects the raw physical features of nodes and edges into a latent representation space. The processor subsequently propagates local interactions over the contact graph through multiple rounds of message passing. The decoder then predicts the acceleration $\mathbf{a}_i^t$ of each particle from the final latent representations. The predicted accelerations are integrated using a semi-implicit Euler scheme to update the particle positions and velocities at the next time step. During training, the model parameters are optimized by minimizing the discrepancy between the predicted accelerations and the ground-truth accelerations generated by high-fidelity numerical simulators (e.g., DEM, SPH, and MPM).

This paper introduces a spatio-temporal edge memory module between the \emph{\textbf{Encoder}} and the \emph{\textbf{Processor}}. Each contact edge $(i, j) \in \mathcal{E}^t$ maintains a persistent memory vector $\mathbf{m}_{ij}^t$. The memory is refreshed at every time step through two coupled operations. First, each contact edge gathers context from the contacts incident to its two particles, so that the update reflects the local multi-contact configuration on the same grains. This spatial context is concatenated with the edge latent $\mathbf{h}_{ij}^t$ into the fused update input $\mathbf{u}_{ij}^t$. A GRU cell then performs a gated update, taking $\mathbf{u}_{ij}^t$ as its input and the pre-update state $\tilde{\mathbf{m}}_{ij}^t$ as its recurrent state. The pre-update state is the memory inherited from the previous step for a persistent contact edge, or a learned birth state for a newly formed one. The updated memory state $\mathbf{m}_{ij}^t$ is then incorporated into the message-passing process. It is shared across all processor layers when computing inter-particle interactions. By carrying memory states across simulation steps, the proposed mechanism accumulates edge-history information across time, while the spatial aggregation couples each contact to its neighbors. As a result, the network can represent history-dependent contact behavior.

\subsection{Encoder}\label{encoder}

\subsubsection{Graph construction}\label{graph-construction}

At each step $t$, a contact graph $\mathcal{G}^t = (\mathcal{V}, \mathcal{E}^t)$ is built from scratch from particle positions. Nodes correspond to all $N$ particles. An edge connects particles $i$ and $j$ when their separation falls below a threshold proportional to their radii:

\begin{equation}
\mathcal{E}^t = \left\{ (i, j) : \left\| \mathbf{x}_i^t - \mathbf{x}_j^t \right\|_2 < \alpha \left( r_i + r_j \right),\ i < j \right\}
\label{eq:edge-set}
\end{equation}

where $\| \cdot \|_2$ denotes the Euclidean distance and $r_i + r_j$ is the center-to-center distance at which the two particles are exactly in contact. The factor $\alpha$ is a skin factor, which causes edges to be established before particles actually collide, providing the network with an early warning of imminent contact. The graph is rebuilt at every time step based on the current particle positions.

\begin{figure}[htbp]
\centering
\includegraphics[width=\linewidth]{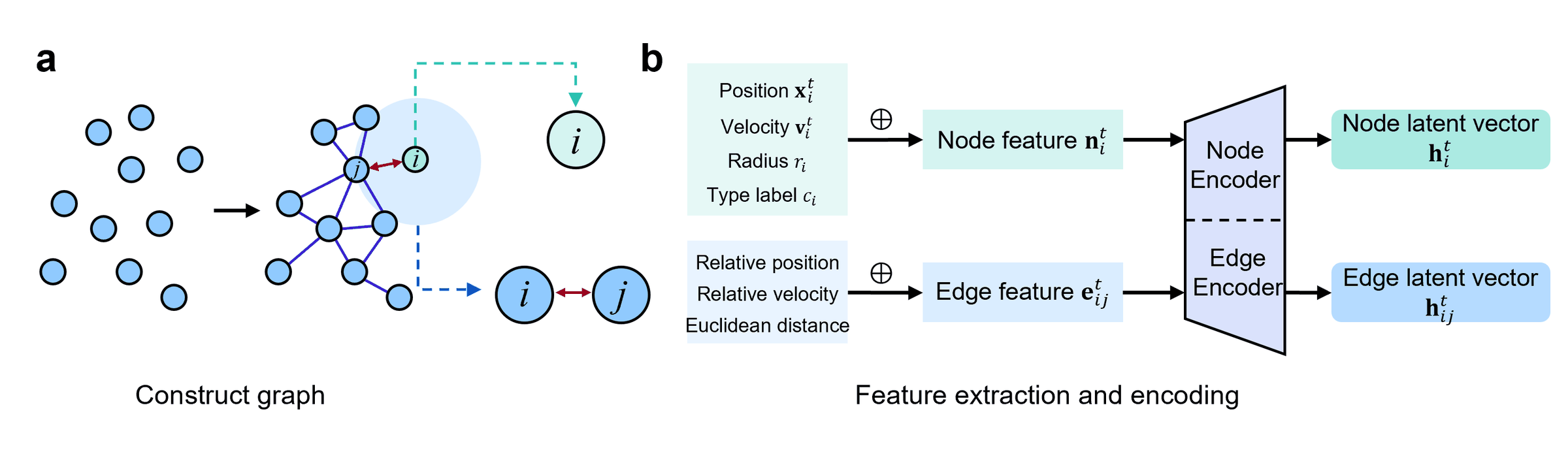}
\caption{Graph construction and feature encoding pipeline. (a) Constructing a graph from particles via a neighborhood radius; (b) Encoding node and edge features into their respective latent vectors.}
\label{fig:graph-encoding}
\end{figure}

Figure~\ref{fig:graph-encoding} illustrates the pipeline from graph construction to node and edge encoding, with the encoding of nodes and edges introduced immediately afterward.

\subsubsection{Node encoding}\label{node-encoding}

Each particle $i$ is described by its kinematic state and static properties. The input feature vector $\mathbf{n}_i^t$ concatenates position $\mathbf{x}_i^t$, velocity $\mathbf{v}_i^t$, radius $r_i$, and type label $c_i$:

\begin{equation}
\mathbf{n}_i^t = \left[ \mathbf{x}_i^t ; \mathbf{v}_i^t ; r_i ; c_i \right] \in \mathbb{R}^{2d+2}
\label{eq:node-feature}
\end{equation}

where $[\,\cdot\,;\,\cdot\,]$ denotes vector concatenation. This input is mapped to a $D$-dimensional latent vector by a node encoder $\operatorname{Enc}^{\mathcal{V}}$:

\begin{equation}
\mathbf{h}_i^{(0)} = \operatorname{Enc}^{\mathcal{V}}\left( \mathbf{n}_{i}^t \right) \in \mathbb{R}^D
\label{eq:node-encode}
\end{equation}

The superscript (0) indicates that no message passing has yet been performed. At this stage, each particle's representation encodes only its own physical state and does not yet incorporate information from its neighbors. After $K$ rounds of message passing, this latent state will evolve successively into $\mathbf{h}_i^{(1)}$, $\mathbf{h}_i^{(2)}$, \ldots, $\mathbf{h}_i^{(K)}$.

In the Equation~(\ref{eq:node-encode}), the node encoder $\operatorname{Enc}^{\mathcal{V}} : \mathbb{R}^{2d+2} \to \mathbb{R}^{D}$ maps the raw particle features to a latent representation. Specifically, the node encoder consists of a two-layer multilayer perceptron (MLP) with sigmoid linear unit (SiLU) activations and an intermediate layer normalization (LayerNorm), as illustrated by the following flow diagram:

\begin{equation}
\mathbb{R}^{2d+2} \xrightarrow{\text{Linear}} \mathbb{R}^{D} \xrightarrow[\text{LayerNorm}]{\text{SiLU}} \mathbb{R}^{D} \xrightarrow{\text{Linear}} \mathbb{R}^{D} \xrightarrow{\text{SiLU}} \mathbb{R}^{D}
\label{eq:encoder-mlp}
\end{equation}

where the first linear layer maps the $(2d+2)$-dimensional input into a $D$-dimensional latent space to provide sufficient representational capacity. Nonlinearity is introduced through the SiLU activation, SiLU is chosen given that it is smooth everywhere and retains a non-zero gradient for negative inputs, which suits a simulator whose outputs are fed back autoregressively and whose gradients propagate through many message-passing rounds and time steps. LayerNorm is employed to stabilize feature distributions and improve training robustness. A second linear transformation further refines the latent representation, followed by a final SiLU activation to enhance the expressive power of the encoder.

\subsubsection{Edge encoding}\label{edge-encoding}

For each edge $(i, j) \in \mathcal{E}^{t}$, a relative kinematic feature vector is computed using only pairwise differences, ensuring translation invariance:

\begin{equation}
\mathbf{e}_{ij}^{t} = \left[ \mathbf{x}_{j}^{t} - \mathbf{x}_{i}^{t} ;\, \mathbf{v}_{j}^{t} - \mathbf{v}_{i}^{t} ;\, \left\| \mathbf{x}_{j}^{t} - \mathbf{x}_{i}^{t} \right\|_{2} \right] \in \mathbb{R}^{2d+1}
\label{eq:edge-feature}
\end{equation}

The edge features comprise the relative position $\mathbf{x}_{j}^{t} - \mathbf{x}_{i}^{t}$, relative velocity $\mathbf{v}_{j}^{t} - \mathbf{v}_{i}^{t}$, and Euclidean separation distance $\left\| \mathbf{x}_{j}^{t} - \mathbf{x}_{i}^{t} \right\|_{2}$, where $\| \cdot \|_{2}$ denotes the $L_2$ (Euclidean) norm. These descriptors capture both the geometric relationship and the local kinematics of the contacting particles, forming the basis for interaction modeling.

Each edge feature is independently encoded through an identically structured MLP:

\begin{equation}
\mathbf{h}_{ij}^{t} = \operatorname{Enc}^{\mathcal{E}}\left( \mathbf{e}_{ij}^{t} \right) \in \mathbb{R}^{D}
\label{eq:edge-encode}
\end{equation}

$\operatorname{Enc}^{\mathcal{V}}$ and $\operatorname{Enc}^{\mathcal{E}}$ share identical depth and width, differing only in their input dimensionality ($2d+2$ and $2d+1$, respectively).

Crucially, $\mathbf{h}_{ij}^t$ carries no message-passing layer superscript. Unlike the node latents $\mathbf{h}_i^{(k)}$, which are updated at every layer, the edge latent is computed once per timestep and held static throughout all $K$ rounds of message passing. It serves as a time-invariant geometric context that nodes consult during message exchange, while the temporal dynamics are captured separately by the edge memory $\mathbf{m}_{ij}^{t}$.

The output of the encoder is a latent graph consisting of node representations $\left\{ \mathbf{h}_i^{(0)} \right\}_{i=1}^{N}$ and edge representations $\left\{ \mathbf{h}_{ij}^{t} \right\}_{(i,j) \in \mathcal{E}^{t}}$, which together form the input to the edge memory update (Section~\ref{edge-memory}) and the subsequent message-passing processor (Section~\ref{message-processor}).

\subsection{Edge memory}\label{edge-memory}

\subsubsection{Memory matrix and identity dictionary}\label{memory-matrix-and-identity-dictionary}

Let $\mathcal{E}^{t-1}$ denote the edge set at the previous simulation step $t-1$. Each edge $(i, j)$ is associated with a memory vector $\mathbf{m}_{ij}^{t-1}$ of dimension $d_{m}$. Stacking these vectors according to the ordering of edges in $\mathcal{E}^{t-1}$ yields the edge memory matrix:

\begin{equation}
\mathbf{M}^{t-1} = \left[ \left( \mathbf{m}_{ij}^{t-1} \right)^{\top} \right]_{(i,j) \in \mathcal{E}^{t-1}} \in \mathbb{R}^{\left| \mathcal{E}^{t-1} \right| \times d_{m}}
\label{eq:memory-matrix}
\end{equation}

where $\left| \mathcal{E}^{t-1} \right|$ denotes the number of edges in the interaction graph at step $t - 1$, and the $k$-th row stores the memory state of the $k$-th edge in $\mathcal{E}^{t-1}$.

However, the interaction graph is reconstructed independently at every simulation step, and the ordering of edges is therefore not guaranteed to remain consistent over time. As a result, the row index of $\mathbf{M}^{t-1}$ alone cannot serve as a persistent reference to the same physical contact across consecutive graph constructions. To establish a stable correspondence between contact edges and their memory states, each contact is assigned a unique, time-invariant identifier.

Specifically, a contact between particles $i$ and $j$ represents an intrinsically symmetric pairwise interaction. Therefore, $(i, j)$ and $(j, i)$ correspond to the same physical contact. Throughout this work, we adopt the canonical convention $i < j$, ensuring that every contact is represented uniquely by a single edge. Under this convention, the edge identifier is defined as:

\begin{equation}
\operatorname{eid}(i, j) = i \cdot N + j, \quad i < j
\label{eq:eid}
\end{equation}

Since particle indices remain fixed throughout the rollout, the same particle pair is assigned the same identifier at every time step. Moreover, the encoding is injective, guaranteeing that different particle pairs correspond to distinct identifiers. Consequently, each physical contact is associated with a unique and temporally invariant identity.

The edge identifier establishes a stable representation of a physical contact, but an additional mechanism is required to locate the corresponding memory state within $\mathbf{M}^{t-1}$. To this end, we maintain an identity dictionary:

\begin{equation}
\mathcal{D}^{t-1} = \left\{ \operatorname{eid}(i, j) \mapsto k \;\middle|\; (i, j) \text{ is the } k\text{-th edge in } \mathcal{E}^{t-1} \right\}
\label{eq:identity-dict}
\end{equation}

Querying $\operatorname{eid}(i, j)$ returns the row index $k$, from which the corresponding memory state is retrieved as the $k$-th row of $\mathbf{M}^{t-1}$. If the identifier is absent from $\mathcal{D}^{t-1}$, the edge is treated as a newly formed contact.

\begin{figure}[htbp]
\centering
\includegraphics[width=\linewidth]{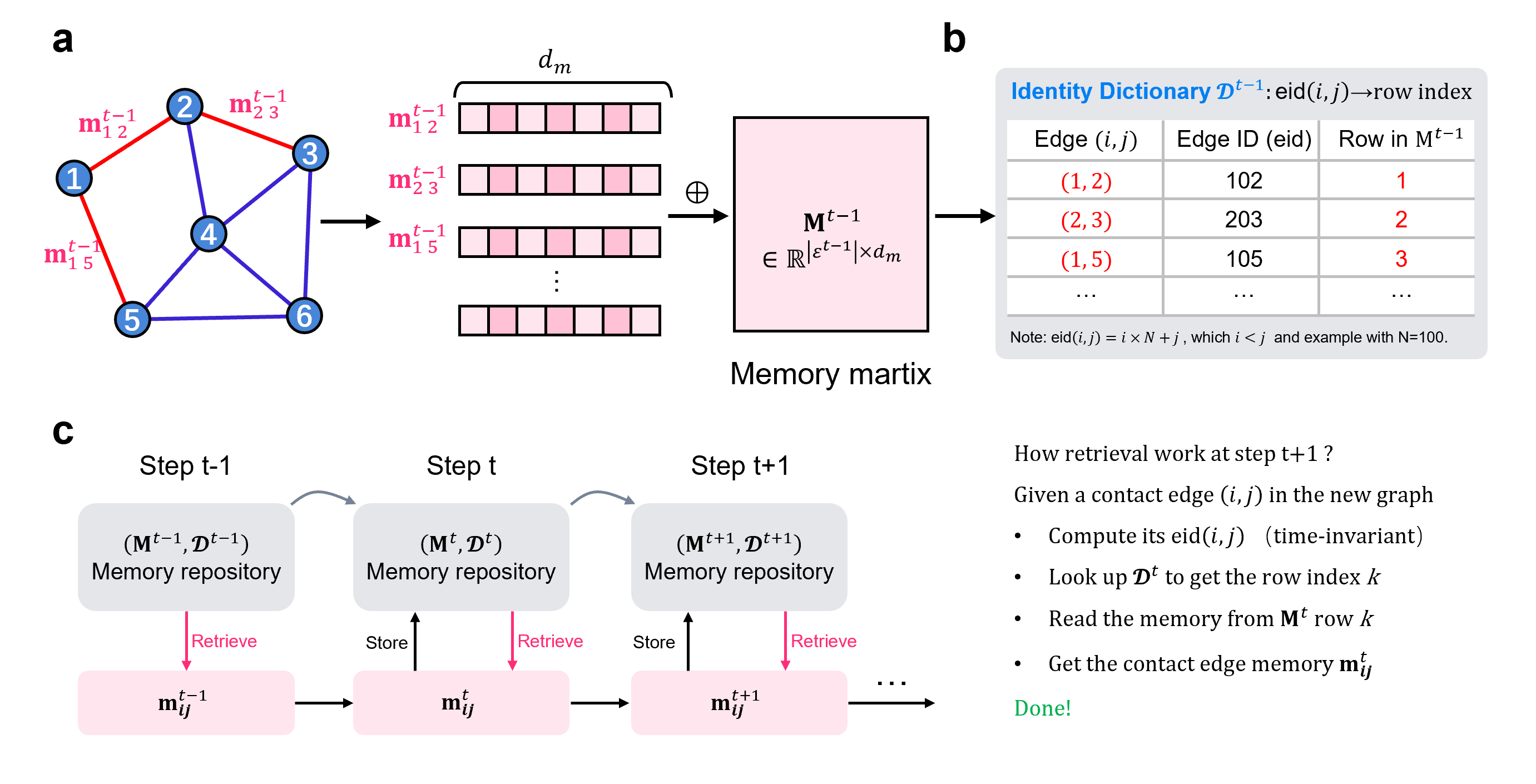}
\caption{Edge memory access mechanism with a memory matrix and an identity dictionary. (a) Memory matrix: edge memory vectors are stacked into matrix $\mathbf{M}^{t-1}$; (b) Identity dictionary: a time-invariant eid $\to$ row-index mapping; (c) Cross-step access: the repository evolves over time, retrieving and storing memories step by step.}
\label{fig:memory-access}
\end{figure}

\subsubsection{Spatiotemporal Memory update}\label{spatiotemporal-memory-update}

The pre-update state depends on whether the contact existed at the previous step. For each edge $(i, j) \in \mathcal{E}^{t}$, its identifier $\operatorname{eid}(i, j)$ is queried in the identity dictionary $\mathcal{D}^{t-1}$:

\begin{equation}
\tilde{\mathbf{m}}_{ij}^{t} = \begin{cases} \operatorname{InitNet}\left( \mathbf{e}_{ij}^{t} \right) & \operatorname{eid}\left( i, j \right) \notin \mathcal{D}^{t-1} \quad \text{(new contact)} \\ \mathbf{m}_{ij}^{t-1} & \operatorname{eid}\left( i, j \right) \in \mathcal{D}^{t-1} \quad \text{(persistent contact)} \end{cases}
\label{eq:pre-update}
\end{equation}

In the Equation~(\ref{eq:pre-update}), $\tilde{\mathbf{m}}_{ij}^t$ is the pre-update memory state of contact $(i, j)$, $\mathbf{m}_{ij}^{t-1}$ is the memory stored for this contact at the previous step, and $\operatorname{InitNet}(\cdot)$ is a trainable initialization network acting on the raw edge features $\mathbf{e}_{ij}^{t}$ of Equation~(\ref{eq:edge-feature}). The identifier $\operatorname{eid}(i, j)$ and the dictionary $\mathcal{D}^{t-1}$ are defined in Equations~(\ref{eq:eid}) and (\ref{eq:identity-dict}). A hit means the pair was already in contact, and the stored memory is inherited. A miss means the contact is new, and $\operatorname{InitNet}(\cdot)$ generates the starting state instead. Every contact, new or persistent, then undergoes the same gated update:

\begin{equation}
\mathbf{m}_{ij}^{t} = \operatorname{GRU}\left( \left[ \mathbf{h}_{ij}^{t}; \mathbf{s}_{ij}^{t} \right], \tilde{\mathbf{m}}_{ij}^{t} \right) = \operatorname{GRU}\left( \mathbf{u}_{ij}^{t}, \tilde{\mathbf{m}}_{ij}^{t} \right)
\label{eq:gru-update}
\end{equation}

\begin{figure}[htbp]
\centering
\includegraphics[width=0.64\linewidth]{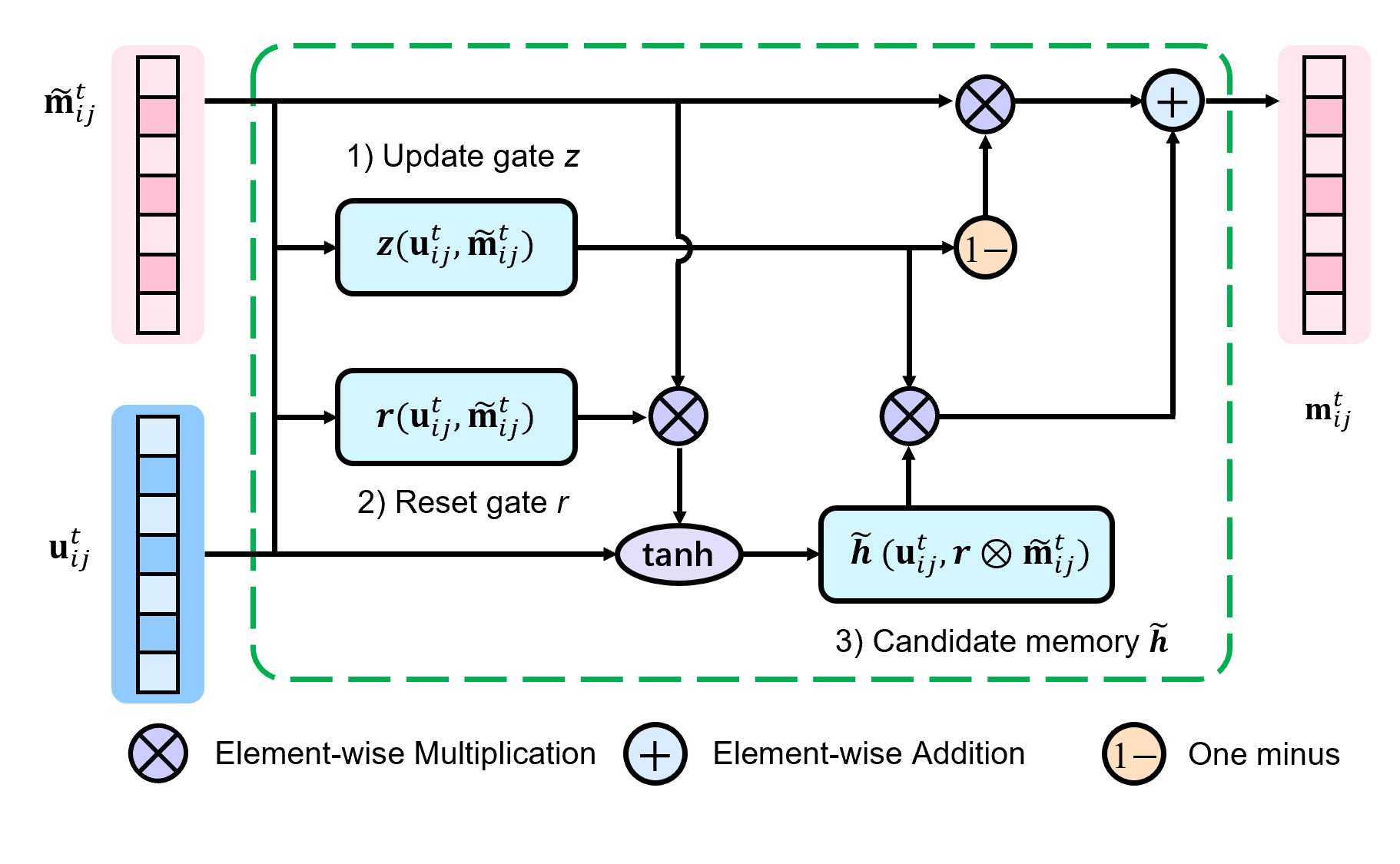}
\caption{The GRU cell used as the memory updater. Given the fused update input $\mathbf{u}_{ij}^{t}$ and the pre-update state $\tilde{\mathbf{m}}_{ij}^t$, the cell computes the update gate, the reset gate, and the candidate memory, and outputs the updated edge memory $\mathbf{m}_{ij}^{t}$.}
\label{fig:gru}
\end{figure}

In the Equation~(\ref{eq:gru-update}), $\mathbf{m}_{ij}^{t}$ is the updated memory of the current step, and $\operatorname{GRU}(\cdot, \cdot)$ is a gated recurrent unit that takes $\mathbf{u}_{ij}^t$ as its input and $\tilde{\mathbf{m}}_{ij}^t$ as its recurrent state. The fused update input $\mathbf{u}_{ij}^t$ concatenates the edge latent $\mathbf{h}_{ij}^{t}$ with the spatial context $\mathbf{s}_{ij}^{t}$. The two inputs play complementary roles. The pre-update state $\tilde{\mathbf{m}}_{ij}^t$ carries the past of the contact, and the fused update input $\mathbf{u}_{ij}^t$ carries the present of the contact and its neighborhood. Figure~\ref{fig:gru} illustrates the GRU module dedicated to memory updating.

\begin{figure}[htbp]
\centering
\includegraphics[width=\linewidth]{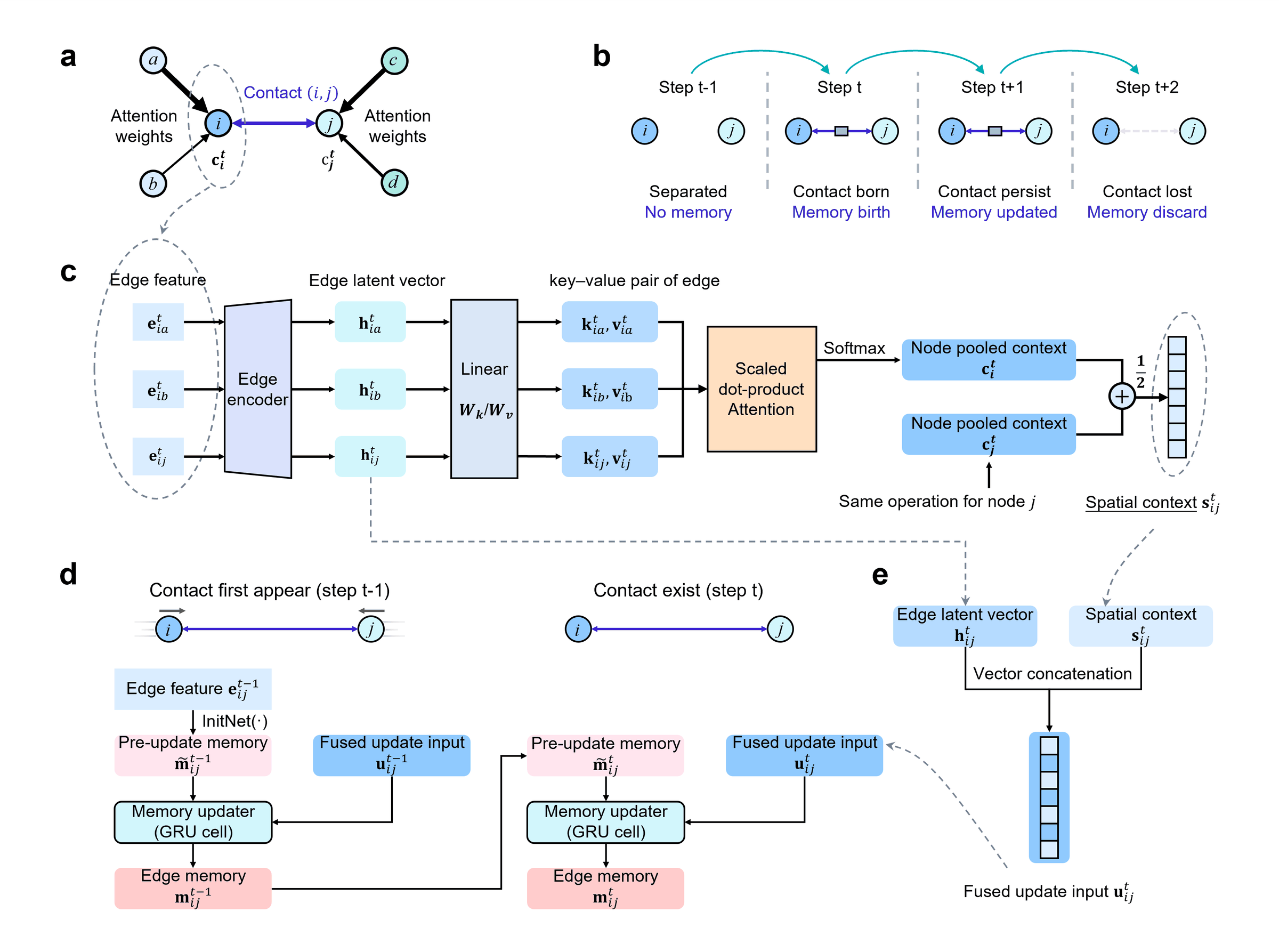}
\caption{Edge memory update mechanism. (a) An illustrative particle system and its contact graph. (b) The complete lifecycle of a contact edge, including absence, formation, persistence, and removal. (c) Attention-based aggregation of neighboring contact information to obtain the spatial context of the target edge. (d) Temporal propagation and update of the edge memory using a GRU. (e) Fusion of the current edge embedding with the aggregated spatial context to form the input for memory updating.}
\label{fig:memory-update}
\end{figure}

Figure~\ref{fig:memory-update} illustrates the complete update procedures, and the three sections below follow its panels, detailing the pre-update state (panels b and d), the spatial context (panel c), and the gated fusion (panels d and e).

\textbf{(1) Pre-update state}

The two branches of Equation~(\ref{eq:pre-update}) differ only in where the state comes from. For a persistent contact, the dictionary returns the row index $k$, and the memory $\mathbf{m}_{ij}^{t-1}$ is read from the $k$-th row of $\mathbf{M}^{t-1}$. For a new contact, no prior memory exists. This covers both a first encounter and a re-contact after separation. $\operatorname{InitNet}(\cdot)$ therefore maps the instantaneous geometric features $\mathbf{e}_{ij}^{t}$ to a starting state. As Figure~\ref{fig:memory-update} (d) shows, the birth state passes through the gated update of Equation~(\ref{eq:gru-update}) like every other contact, so even a newly formed contact ends the step informed of its own state and of its neighborhood.

\textbf{(2). Spatial context}

Before the gated update, each contact gathers information from the contacts that share one of its two particles. These co-incident contacts are its neighbors in the line graph of the interaction network. An attention mechanism performs the aggregation, so each co-incident contact contributes according to its learned relevance. Let $\mathcal{E}(p)$ denote the set of contacts incident to particle $p$. The computation follows the information in Figure~\ref{fig:memory-update} (c).

First, From the edge latent produced by the edge encoder (Section~\ref{edge-encoding}), a key and a value are obtained by linear projection:

\begin{equation}
\mathbf{k}_{ij}^{t} = \mathbf{W}_{\mathbf{k}} \mathbf{h}_{ij}^{t}, \quad \mathbf{v}_{ij}^{t} = \mathbf{W}_{\mathbf{v}} \mathbf{h}_{ij}^{t}
\label{eq:key-value}
\end{equation}

In Equation~(\ref{eq:key-value}), $\mathbf{k}_{ij}^{t}$ and $\mathbf{v}_{ij}^{t}$ are the key and the value of contact $(i, j)$, and $\mathbf{W}_{\mathbf{k}}$, $\mathbf{W}_{\mathbf{v}}$ are learnable linear projections shared by all contacts. The key controls how the contact is scored. The value is the content handed over once the weights are set.

Second, every particle scores and pools the contacts acting on it. A single learned query $\mathbf{q}$, shared by all contacts, rates each key. The scores are normalized among the contacts incident to the same particle, and the values are summed with the resulting weights:

\begin{equation}
\alpha_{p,ij}^{t} = \operatorname{softmax}\left( \mathbf{q}^{\top} \mathbf{k}_{ij}^{t} / \sqrt{D} \right)
\label{eq:attention}
\end{equation}

\begin{equation}
\mathbf{c}_{p}^{t} = \sum_{(ij) \in \mathcal{E}(p)} \alpha_{p,ij}^{t} \mathbf{v}_{ij}^{t}
\label{eq:pooled-context}
\end{equation}

In Equations~(\ref{eq:attention}) and (\ref{eq:pooled-context}), $\alpha_{p,ij}^{t}$ is the attention weight of contact $(i, j)$ within the pool of particle $p$, $\mathbf{q}$ is the shared learned query, the factor $\sqrt{D}$ keeps the scores in a numerically stable range, and $\mathbf{c}_{p}^{t}$ is the pooled context of particle $p$, a summary of the instantaneous state of all contacts acting on it.

Third, every contact reads the summaries back. Its spatial context is the average over its two endpoints:

\begin{equation}
\mathbf{s}_{ij}^{t} = \frac{1}{2}\left( \mathbf{c}_{i}^{t} + \mathbf{c}_{j}^{t} \right)
\label{eq:spatial-context}
\end{equation}

In Equation~(\ref{eq:spatial-context}), $\mathbf{s}_{ij}^{t}$ is the spatial context of contact $(i, j)$, and $\mathbf{c}_{i}^{t}$ and $\mathbf{c}_{j}^{t}$ are the pooled contexts of its two endpoint particles. The information has now completed one round trip between the contacts and the particles. The spatial context therefore describes what is happening on the two grains that the contact connects.

The GRU of Equation~(\ref{eq:gru-update}) merges the three information sources. The edge latent supplies the current state of the contact itself. The spatial context supplies the current state of its neighborhood. The pre-update state supplies the accumulated history. Through its gates, the GRU decides how much history to keep and how much of the present to write in.

\subsubsection{Dictionary Propagation}\label{dictionary-propagation}

After the memory states of all edges in $\mathcal{E}^{t}$ have been updated, the identity dictionary for the current step is constructed as:

\begin{equation}
\mathcal{D}^{t} = \left\{ \operatorname{eid}(i,j) \mapsto k \,\middle|\, (i,j) \text{ is the } k\text{-th edge in } \mathcal{E}^{t} \right\}
\label{eq:dict-propagation}
\end{equation}

Together with the updated memory matrix $\mathbf{M}^{t}$, the dictionary $\mathcal{D}^{t}$ forms the memory repository for the current step and is propagated to the next simulation step. The overall memory lifecycle can be summarized as:

\begin{equation}
\left( \mathbf{M}^{t-1}, \mathcal{D}^{t-1} \right) \to m_{ij}^{t-1} \to m_{ij}^{t} \to \left( \mathbf{M}^{t}, \mathcal{D}^{t} \right)
\label{eq:memory-lifecycle}
\end{equation}

where the repository from the previous step is first queried to retrieve edge memories, the retrieved states are updated using the current edge features, and the resulting memories are stored in a new repository for future retrieval.

At step $t+1$, the dictionary $\mathcal{D}^{t}$ enables the model to determine whether a contact is persistent and, if so, to retrieve the corresponding memory state from $\mathbf{M}^{t}$ regardless of any reordering of the contact graph.

\subsection{Message processor}\label{message-processor}

The processor propagates information through the particle interaction network to enrich the node representations before decoding. Since one round communicates information across a single contact, stacking $K$ rounds allows each particle to gather information from particles that are up to $K$ contacts away.

To propagate contact-history information through the interaction network, the contact memory is injected into every message-passing round. For the message sent from a contacting neighbor $j$ to particle $i$ at round $k$, a multilayer perceptron $\psi^{(k)}$ combines the edge latent, the contact memory, and the current node representations:

\begin{equation}
\mathbf{g}_{ij}^{(k)} = \psi^{(k)}\left(\left[\mathbf{h}_{ij}^{t}; \mathbf{m}_{ij}^{t}; \mathbf{h}_{i}^{(k)}; \mathbf{h}_{j}^{(k)}\right]\right)
\end{equation}

Here, $\mathbf{g}_{ij}^{(k)}$ denotes the message passed from a neighboring particle $j$ to particle $i$ at the $k$-th propagation round, and $\psi^{(k)}$ is a round-specific MLP that adopts the same architecture as the encoder but maintains independent parameters across rounds. The four concatenated inputs contribute complementary information: the edge latent $\mathbf{h}_{ij}^{t}$ provides the instantaneous contact geometry at step $t$; the contact memory $\mathbf{m}_{ij}^{t}$ supplies the accumulated interaction history of this contact; and $\mathbf{h}_{i}^{(k)}$ and $\mathbf{h}_{j}^{(k)}$ are the current latent states of the receiving particle $i$ and the sending particle $j$, respectively. As noted above, $\mathbf{h}_{ij}^{t}$ and $\mathbf{m}_{ij}^{t}$ are held fixed throughout the $K$ rounds, so that only the node states $\mathbf{h}_{i}^{(k)}$ and $\mathbf{h}_{j}^{(k)}$ vary with the round index $k$.

Injecting ($\mathbf{m}_{ij}^{t}$) at every propagation round allows accumulated contact history to continuously modulate information flow, enabling path-dependent effects associated with individual contacts to influence neighboring particles through repeated message passing. Messages arriving at particle $i$ are aggregated by summation:

\begin{equation}
\mathbf{g}_{i}^{(k)} = \sum_{j \in \mathcal{N}(i)} \mathbf{g}_{ij}^{(k)}
\end{equation}

where $\mathcal{N}(i)$ denotes the contact neighborhood of particle $i$. The aggregated message is then combined with the current node representation and applied as a residual update:

\begin{equation}
\mathbf{h}_{i}^{(k+1)} = \mathbf{h}_{i}^{(k)} + \operatorname{RMSNorm}\left(\mathcal{X}^{(k)}\left(\left[\mathbf{h}_{i}^{(k)}; \mathbf{g}_{i}^{(k)}\right]\right)\right)
\end{equation}

Both $\mathcal{X}^{(k)}$ and $\psi^{(k)}$ adopt the same architecture as the encoder, with independent parameters across propagation rounds. Residual connections and normalization are employed to stabilize deep message passing.

After $K$ rounds, the processor outputs the final node representations $\left\{\mathbf{h}_{i}^{(K)}\right\}_{i=1}^{N}$, which encode both local contact geometry and contact-history information propagated through the interaction graph. These representations are subsequently mapped to particle accelerations by the decoder.

\subsection{Physics-structured decoder}\label{physics-structured-decoder}

Reflecting the structure of granular dynamics, the predicted acceleration of particle $i$ is decomposed into an external contribution and an internal contact contribution:

\begin{equation}
\mathbf{a}_{i}^{t} = \mathbf{a}_{i}^{\text{ext}} + \mathbf{a}_{i}^{\text{int}}
\end{equation}

where the external term $\mathbf{a}_{i}^{\text{ext}}$ collects the body forces and the smooth response to the domain boundaries and is read from the node representation, whereas the internal term $\mathbf{a}_{i}^{\text{int}}$ collects the inter-particle contact forces and is read from the contact memory. The two terms are produced by two separate read-out heads, described below.

\subsubsection{External forces}\label{external-forces}

Body forces, together with any smoothly varying response to the simulation boundaries, act on individual particles. They are decoded from the final node representation $\mathbf{h}_{i}^{(K)}$ returned by the processor through a node decoder:

\begin{equation}
\mathbf{a}_{i}^{\text{ext}} = \operatorname{Dec}^{\mathcal{V}}\left(\mathbf{h}_{i}^{(K)}\right) \in \mathbb{R}^{d}
\end{equation}

Here, $\text{Dec}^{\mathcal{V}}$ is a two-layer MLP whose output layer is linear (no activation), so that $\mathbf{a}_i^{\text{ext}}$ is unconstrained in sign and can represent, for example, the downward gravitational acceleration. This is the only stage at which the spatially propagated node representation enters the prediction: through the $K$ rounds of message passing, $\mathbf{h}_i^{(K)}$ summarizes the local packing context of particle $i$, which the external read-out requires.

\subsubsection{Internal contact forces}\label{internal-contact-forces}

The inter-particle contact forces are decoded directly from each contact. For every contact $(i, j) \in \mathcal{E}^t$ in the canonical ordering $i < j$ (Section~\ref{graph-construction}), the contact memory $\mathbf{m}_{ij}^t$ and the edge latent $\mathbf{h}_{ij}^t$ are concatenated and passed through an edge decoder $\text{Dec}^{\mathcal{E}}$ (a shared MLP followed by three linear heads) that outputs a normal-force magnitude, a tangential-force vector, and a friction coefficient:

\begin{equation}
\left( F_n^{ij}, \tilde{\mathbf{F}}_t^{ij}, \mu_{ij} \right) = \operatorname{Dec}^{\mathcal{E}}\left( \left[ \mathbf{m}_{ij}^t ; \mathbf{h}_{ij}^t \right] \right)
\end{equation}

\begin{figure}[htbp]
\centering
\includegraphics[width=\linewidth]{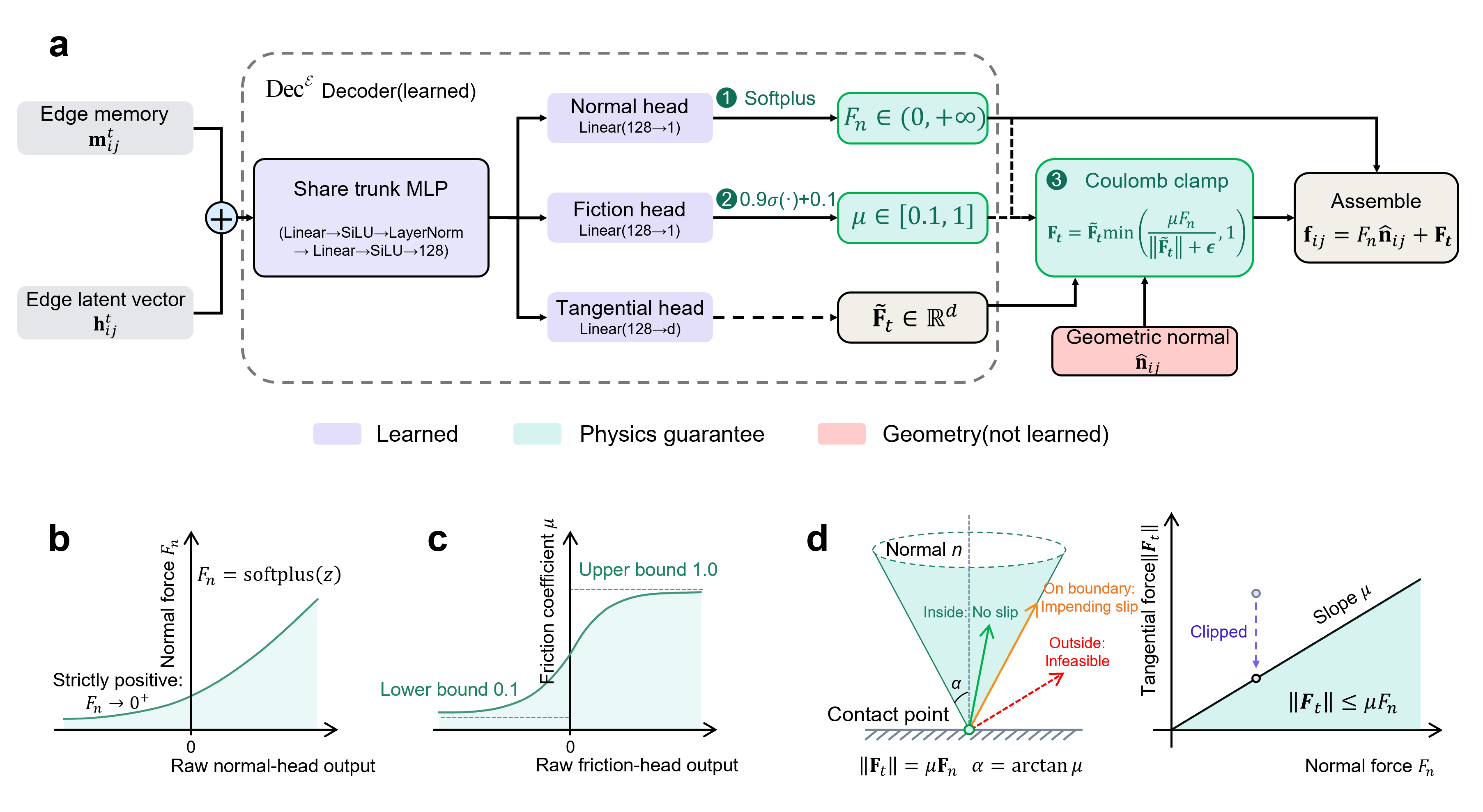}
\caption{Physics-constrained contact-force decoder. (a) The concatenated edge memory and edge latent pass through a shared trunk MLP and three parallel heads, decoding the normal force $F_n$, the friction coefficient $\mu$, and an unconstrained tangential force $\tilde{\mathbf{F}}_t$. (b) A softplus output guarantees strictly positive, purely repulsive normal forces. (c) A scaled sigmoid $0.9\sigma(\cdot)+0.1$ bounds the friction coefficient within [0.1, 1.0]. (d) Coulomb clamping rescales the tangential force into the friction cone $\|\mathbf{F}_t\| \le \mu F_n$; combined with $\mu \le 1$, the total contact force remains repulsive.}
\label{fig:decoder}
\end{figure}

Here, $F_{n}^{ij}$ is the scalar normal-force magnitude, $\tilde{\mathbf{F}}_{t}^{ij}$ the unconstrained tangential-force vector, and $\mu_{ij}$ the friction coefficient of contact edge $(i, j)$, while $[\,\cdot\,;\,\cdot\,]$ denotes the vector concatenation introduced in Section~\ref{node-encoding}. Figure~\ref{fig:decoder} shows the contact force decoding process under physical constraints.

Each of these outputs is then constrained to a physically admissible range. The normal magnitude is constrained to be non-negative, $F_n^{ij} \geq 0$, so that contacts are purely repulsive, while $\mu_{ij}$ is a learned (per-contact) friction coefficient. The raw tangential vector $\tilde{\mathbf{F}}_t^{ij}$ is then projected onto the Coulomb friction cone, so that the tangential force never exceeds the frictional limit:

\begin{equation}
\mathbf{F}_t^{ij} = \tilde{\mathbf{F}}_t^{ij} \min\left( \frac{\mu_{ij} F_n^{ij}}{\left\| \tilde{\mathbf{F}}_t^{ij} \right\| + \epsilon}, 1 \right), \quad \text{so that} \quad \left\| \mathbf{F}_t^{ij} \right\| \leq \mu_{ij} F_n^{ij}
\end{equation}

Here, $\epsilon$ is a small constant for numerical stability. The total force exerted on particle $i$ by particle $j$ combines the normal and tangential components as follow:

\begin{equation}
\mathbf{f}_{ij} = F_n^{ij} \hat{\mathbf{n}}_{ij} + \mathbf{F}_t^{ij}, \quad \hat{\mathbf{n}}_{ij} = \frac{\mathbf{x}_i^t - \mathbf{x}_j^t}{\left\| \mathbf{x}_i^t - \mathbf{x}_j^t \right\|}
\end{equation}

where $\hat{\mathbf{n}}_{ij}$ is the unit normal directed from $j$ to $i$, along which the repulsive normal force acts.

\subsubsection{Momentum conservation}\label{momentum-conservation}

Because each contact is represented by a single undirected edge, the decoded force is applied to its two particles with equal magnitude and opposite sign, in accordance with Newton's third law:

\begin{equation}
\mathbf{a}_i^{\text{int}} = \frac{1}{m_i} \sum_{j \in \mathcal{N}(i)} \mathbf{f}_{ij}, \quad \mathbf{f}_{ji} = -\mathbf{f}_{ij}
\end{equation}

where $\mathcal{N}(i)$ denotes the contact neighborhood of particle $i$ and $m_i$ its mass. Summed over all particles, the internal forces cancel pairwise, so that $\sum_i m_i \mathbf{a}_i^{\text{int}} = \mathbf{0}$ at every step, independently of the network parameters. The external and internal contributions are finally summed to yield the physical acceleration $\mathbf{a}_i^t = \mathbf{a}_i^{\text{ext}} + \mathbf{a}_i^{\text{int}}$, which is advanced in time by the integrator described in Section~\ref{integration}.

\subsection{Integration}\label{integration}

Given the physical acceleration $\mathbf{a}_i^t$ produced by the decoder, the integrator advances the particle state to the next step. The state is advanced with a semi-implicit Euler scheme:

\begin{equation}
\mathbf{v}_i^{t+1} = \mathbf{v}_i^t + \mathbf{a}_i^t \Delta t, \quad \mathbf{x}_i^{t+1} = \mathbf{x}_i^t + \mathbf{v}_i^{t+1} \Delta t
\end{equation}

The learned dynamics do not strictly guarantee two geometric properties of the physical system: particles must not interpenetrate, and they must remain inside the simulation domain. Two hard constraints are therefore applied after each update.

(1) \emph{Non-penetration projection.} The learned update can leave pairs of particles overlapping. A position projection removes this residual overlap. The assigned radii define the minimum admissible distance between particle centers. For every pair $(i, j)$ with positive overlap $\delta_{ij} = \left( r_i + r_j \right) - \left\| \mathbf{x}_i - \mathbf{x}_j \right\| > 0$, the two particles are displaced apart along their contact normal by half of the overlap each:

\begin{equation}
\mathbf{x}_i \leftarrow \mathbf{x}_i + \frac{1}{2}\delta_{ij}\hat{\mathbf{n}}_{ij}, \quad \mathbf{x}_j \leftarrow \mathbf{x}_j - \frac{1}{2}\delta_{ij}\hat{\mathbf{n}}_{ij}
\end{equation}

where $\hat{\mathbf{n}}_{ij}$ is the unit contact normal. This position-based (Jacobi-style) projection is applied to all overlapping pairs in parallel and iterated a fixed number of times per step.

(2) \emph{Boundary projection.} Particles are confined to the simulation domain. Any particle whose center penetrates the floor or a lateral wall is projected back onto the boundary, and the velocity component pointing into the wall is set to zero, so that particles neither cross nor stick to the boundary.

The overlap projection adjusts only positions; the wall projection also zeros the inward velocity component. Both act only during rollout. Together with the force-level guarantees built into the decoder, they keep the simulated trajectory free of interpenetration and out-of-domain states without constraining the learned force model.

\subsection{Training strategy}\label{training-strategy}

The first stage adopts teacher forcing. At each time step, the model takes the reference state as input and independently predicts the single-step acceleration, while only the contact memory is propagated across the temporal window.

The second stage uses autoregressive rollout: starting from a single reference state, all subsequent states are generated from the model's own predictions. A gradient-free drift segment first produces off-trajectory states that reflect the model's accumulated errors, after which multi-step position supervision is applied to reduce further error growth.

The first stage is intended to learn accurate local dynamics, whereas the second improves long-horizon stability under the model's own rollout distribution.

\subsubsection{Single-step pretraining}\label{single-step-pretraining}

The model is first trained to predict the single-step acceleration of each particle. Meanwhile, the recurrent contact memory is propagated across consecutive frames so that it can learn interaction histories relevant to future predictions. This stage combines normalized acceleration regression, input noise, and truncated backpropagation through time.

At time step $t$, the target acceleration is obtained from the reference trajectory by finite differencing the particle velocities:

\begin{equation}
\mathbf{a}_i^{\star,t} = \frac{\mathbf{v}_i^{t+1} - \mathbf{v}_i^{t}}{\Delta t}
\label{eq:accel-target}
\end{equation}

where $\mathbf{a}_i^{\star,t}$ is the target acceleration of particle $i$, $\mathbf{v}_i^{t}$ and $\mathbf{v}_i^{t+1}$ are its reference velocities at two consecutive steps, and $\Delta t$ is the simulation time step.

The predicted and target accelerations are compared in a normalized space. The per-step loss is:

\begin{equation}
\ell^t = \frac{1}{N} \sum_{i=1}^{N} \operatorname{Huber}\left( \operatorname{norm}(\mathbf{a}_i^{t}), \operatorname{norm}(\mathbf{a}_i^{\star,t}) \right)
\end{equation}

where $\ell^t$ is the per-step loss, and $N$ is the number of particles, $\mathbf{a}_i^{t}$ is the acceleration predicted by the decoder (Section~\ref{physics-structured-decoder}), and $\operatorname{norm}(\cdot)$ is a per-channel normalizer fitted to the training set. The Huber loss is quadratic for small residuals and linear for large ones, making training less sensitive to occasional acceleration outliers. Small Gaussian perturbations are also added to the input positions and velocities to improve robustness to imperfect states encountered during rollout.

Because the contact memory evolves recurrently (Section~\ref{edge-memory}), training uses truncated backpropagation through time over windows of $T_{\text{W}}$ consecutive frames. Within each window, the contact memories and identity dictionary are propagated exactly as during inference, while each step remains supervised by the acceleration target in Equation~(\ref{eq:accel-target}). The window loss is:

\begin{equation}
\mathcal{L} = \frac{1}{T_{\text{W}}} \sum_{k=0}^{T_{\text{W}}-1} \ell^{t_0+k}
\end{equation}

where $t_0$ denotes the beginning of the window. Gradients are propagated through the memory recurrence within the window, allowing the contact states to retain history that improves subsequent predictions.

\subsubsection{Rollout fine-tuning}\label{rollout-fine-tuning}

Teacher-forced pretraining exposes the model primarily to reference states. During autoregressive inference, however, each predicted state becomes the input to the next step, causing errors to accumulate and gradually shifting the rollout away from the training distribution. Input noise alleviates this mismatch only partially because it introduces small independent perturbations rather than the spatially and temporally correlated errors produced by the model itself.

\begin{figure}[htbp]
\centering
\includegraphics[width=\linewidth]{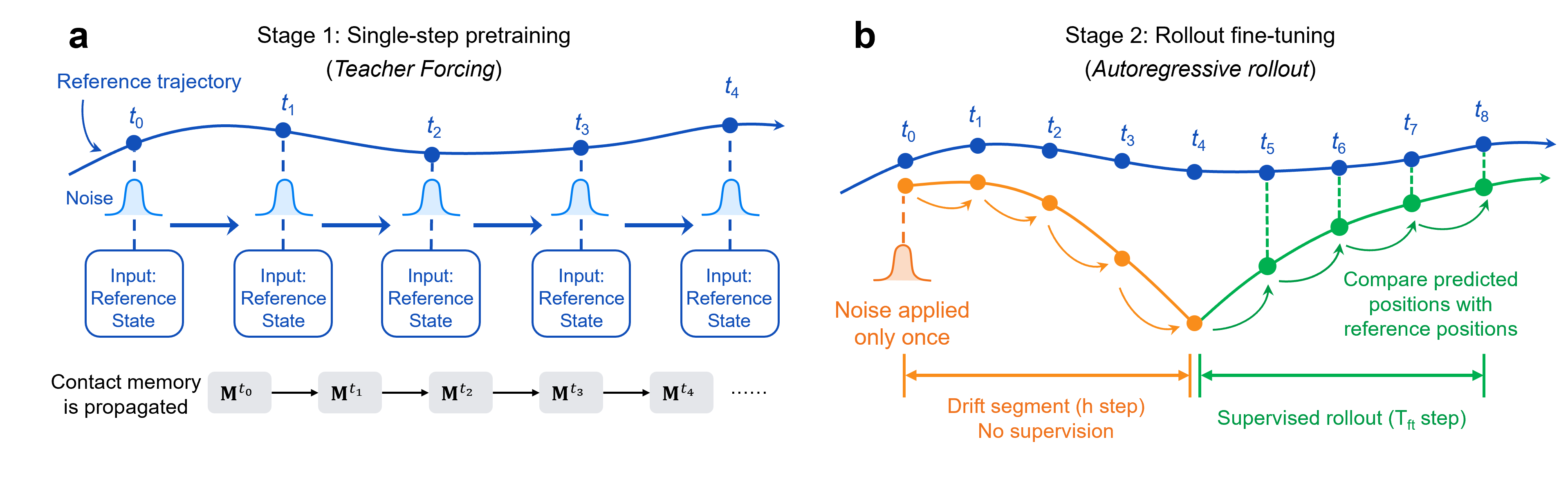}
\caption{Two-stage training strategy for the simulator. (a) Single-step pretraining with teacher forcing. (b) Autoregressive rollout fine-tuning.}
\label{fig:training}
\end{figure}

To reduce this train--test mismatch, we follow the pushforward principle of Brandstetter et al.~\citep{brandstetter2022message} and fine-tune the model through autoregressive rollouts. Each training sample consists of a gradient-free drift segment followed by a supervised rollout segment.

The drift segment starts from a reference state and advances the model autoregressively for a randomly sampled number of steps, up to a maximum drift depth $H_{\max}$. The geometric constraints described in Section~\ref{integration} are applied exactly as during inference. This segment is executed without supervision or gradient tracking, its purpose is to generate a realistic off-trajectory particle state together with the corresponding contact memory.

The resulting state initializes a supervised rollout of $T_{\text{ft}}$ steps. Because the model state has already deviated from the reference trajectory, the single-step acceleration target in Equation~(\ref{eq:accel-target}), which is defined from reference states, is no longer consistent with the current model state. Supervision is therefore applied directly to the particle positions:

\begin{equation}
\mathcal{L}_{\text{ft}} = \frac{1}{T_{\text{ft}}} \sum_{k=1}^{T_{\text{ft}}} \frac{1}{N} \sum_{i=1}^{N} \left\| \mathbf{x}_i^{t_0+k} - \hat{\mathbf{x}}_i^{t_0+k} \right\|_2^2
\end{equation}

where $t_0$ denotes the start of the supervised segment, $\mathbf{x}_i^{t_0+k}$ is the position of particle $i$ generated by the constrained rollout, and $\hat{\mathbf{x}}_i^{t_0+k}$ is the corresponding reference position. The loss averages the squared position error over all particles and supervised rollout steps. By optimizing predictions from states generated by the model itself, this objective improves the model's ability to limit error accumulation during long-horizon simulation.

\section{Numerical Experiments}\label{experiments}

\subsection{Datasets}\label{datasets}

We evaluate the method on the Sand benchmark introduced by Sanchez-Gonzalez et al.~\citep{sanchezgonzalez2020learning}, considering both its two-dimensional (2D-Sand) and three-dimensional (3D-Sand) settings. In each scenario, one or more compact granular bodies are initialized at random locations inside a closed rectangular container and released under gravity. As the material impacts the boundaries, spreads, and interacts with neighboring bodies, contacts continuously emerge, slide, separate, and reform throughout the simulation. Detail benchmark specifications are provided in Appendix A.1.

Both datasets were generated using the MPM, a high-fidelity continuum solver for large-deformation granular dynamics that has become a standard benchmark for learned particle simulators. Their main characteristics are summarized in Table~\ref{tab:datasets}.

\begin{table}[htbp]
\centering
\caption{Statistics of the Granular Sand benchmark datasets used in this work.}
\label{tab:datasets}
\small
\begin{tabular}{@{}lcc@{}}
\toprule
\textbf{Property} & \textbf{2D-Sand} & \textbf{3D-Sand} \\
\midrule
Spatial dimension & 2 & 3 \\
Ground-truth solver & \multicolumn{2}{c}{MPM} \\
Training / validation / test trajectories & 1000 / 30 / 30 & 1000 / 100 / 100 \\
Particles per scene (min--max) & 107--1976 & 4586--19762 \\
Mean particles per scene & $\approx$1159 & $\approx$10000 \\
Steps per trajectory & 320 & 350 \\
Time step $\Delta t$ & \multicolumn{2}{c}{$2.5\times 10^{-3}$} \\
Simulation domain & $[0.1,0.9]^{2}$ & $[0.2,0.8]^{3}$ \\
Boundary condition & \multicolumn{2}{c}{Rigid confining walls} \\
\bottomrule
\end{tabular}
\end{table}

Note. Both datasets are generated using the MPM. One or more compact granular bodies are randomly initialized inside a closed container and released under gravity, producing free fall, impact, spreading, and wall/inter-body collisions before reaching a stable deposit. Particle velocities and accelerations are reconstructed from positions by finite differences. Ground-truth contact forces are not available. During graph construction, particles are connected within a fixed neighborhood radius (0.015 in 2D and 0.025 in 3D). Since the MPM data contain only point positions, each particle is assigned a uniform radius for contact reasoning, chosen as approximately half of the median nearest-neighbour spacing.

Throughout, a trajectory denotes one simulated sequence produced by the reference solver. Each test trajectory is defined by its initial particle configuration, and the model is evaluated by rolling out from that initial state. Each trajectory stores only particle positions. Particle velocities and accelerations are reconstructed by finite differences, and the acceleration targets are normalized independently for each channel using statistics computed from the training set. Importantly, neither dataset provides ground-truth contact forces. Consequently, the per-contact normal and tangential forces predicted by TRACE are learned entirely without force supervision, requiring the model to infer path-dependent frictional interactions solely from particle kinematics.

\subsection{Computing environment}\label{computing-environment}

All experiments reported in this work, including training, rollout evaluation, and timing measurements, were carried out on a single workstation (Table~\ref{tab:hardware}). The machine is equipped with an AMD Ryzen Threadripper PRO 5995WX processor (64 cores, 128 threads), 512 GB of system memory, and four NVIDIA GeForce RTX 4090 GPUs with 24 GB of memory each. The software stack is Ubuntu 22.04 LTS, PyTorch 2.5.1 with CUDA 12.1, and PyTorch Geometric 2.8.0. All models are trained and evaluated in single precision. 2D-Sand training uses one GPU. 3D-Sand training uses four GPUs through distributed data parallelism, while all 3D-Sand evaluations run on a single GPU. All models and baselines were evaluated on the same machine to ensure a fair comparison of accuracy and runtime across the two case studies. All reported timings are measured after a warm-up phase and with explicit device synchronization. The complete hardware and software configuration is listed below.

\begin{table}[htbp]
\centering
\caption{Hardware and software configurations}
\label{tab:hardware}
\small
\begin{tabular}{@{}lp{0.62\linewidth}@{}}
\toprule
Component & Specification \\
\midrule
CPU & AMD Ryzen Threadripper PRO 5995WX, 64 cores/128 threads \\
System memory & 512 GB \\
GPU & 4$\times$ NVIDIA GeForce RTX 4090, 24 GB GDDR6X each \\
Operating system & Ubuntu 22.04.3 LTS. Linux kernel 6.8 \\
Deep-learning framework & PyTorch 2.5.1 (CUDA 12.1, cuDNN 9.1) \\
Graph library & PyTorch Geometric 2.8.0 \\
Python & 3.13 \\
Numerical precision & float32 \\
Training parallelism & 1 GPU (2D-Sand); 4-GPU DistributedDataParallel (3D-Sand) \\
Inference & 1 GPU for all reported rollouts and timings \\
\bottomrule
\end{tabular}
\end{table}

\subsection{Evaluation protocol}\label{evaluation-protocol}

At test time, the simulator is applied exactly as at deployment. Given the state at one step, it predicts the particle accelerations, the integrator advances the state, and the result becomes the input of the next step. Section~\ref{short-and-long-rollout-protocols} defines the two rollout protocols under which this loop is executed. Performance is quantified by position RMSE for kinematic accuracy (Section~\ref{kinematic-accuracy}), by a normalized deposit error for deposit morphology (Section~\ref{deposit-morphology}), and by a kinetic-energy diagnostic for physical consistency (Section~\ref{physical-consistency-diagnostics}).

\subsubsection{Short and long rollout protocols}\label{short-and-long-rollout-protocols}

The two protocols differ only in how long the autoregressive loop runs before the state is corrected.

\textbf{Short rollout.} The simulator runs autoregressively in windows of 20 steps. At the start of each window, the state is re-initialized from the ground truth, and any recurrent state, such as the contact memory of TRACE, is reset. Errors can therefore accumulate for at most 20 steps. This protocol isolates local prediction accuracy, and a window length of one reduces to single-step evaluation.

\textbf{Long rollout.} The simulator starts from the initial state and velocity and predicts the entire trajectory without further access to the ground truth. This is the deployment condition. Errors accumulate freely over the full horizon, so this protocol measures long-term stability in addition to accuracy.

The two protocols are complementary. A model that scores well under the short protocol but poorly under the long one has learned the local dynamics but not their stable long-horizon composition. Unless otherwise stated, all quantitative results are reported for the long rollout.

\subsubsection{Kinematic accuracy}\label{kinematic-accuracy}

The primary metric is the rollout position RMSE:

\begin{equation}
\operatorname{RMSE}(t) = \sqrt{ \frac{1}{N} \sum_{i=1}^{N} \left\| \mathbf{x}_i^{t} - \hat{\mathbf{x}}_i^{t} \right\|_2^2 }
\end{equation}

where $\mathbf{x}_i^{t}$ denotes the predicted position and $\hat{\mathbf{x}}_i^{t}$ the reference position of particle $i$ at rollout step $t$. We report both the time-resolved curve and its mean over the rollout horizon, averaged over the test trajectories.

\subsubsection{Deposit morphology}\label{deposit-morphology}

For granular collapse problems, the engineering-relevant outcome is the geometry of the final deposit. We therefore measure a normalized deposit error combining runout and height:

\begin{equation}
\epsilon_{\text{dep}} = \frac{\left| L - \hat{L} \right| + \left| H - \hat{H} \right|}{L_{\text{box}}}
\end{equation}

where $L$ and $H$ denote the runout (horizontal extent) and height of the predicted deposit at the final frame, $\hat{L}$ and $\hat{H}$ their reference counterparts, and $L_{\text{box}}$ the domain size.

\subsubsection{Physical-consistency diagnostics}\label{physical-consistency-diagnostics}

Low positional error does not necessarily imply physically consistent behavior, so we evaluate three additional diagnostics. The first is the kinetic-energy evolution compared with the MPM reference. The kinetic energy $E_k$ at step $t$ is:

\begin{equation}
E_k(t) = \frac{1}{2} \sum_i \left| \mathbf{v}_i^{t} \right|^2
\end{equation}

where the velocities $\mathbf{v}_i^{t}$ are reconstructed from consecutive positions by finite differences and all particles carry unit mass. Because the test scenes differ in size and drop height, the energy of each trajectory is normalized by the peak ground-truth value of that trajectory, placing all curves on a common scale.

\subsection{Case study 1: 2D Sand column collapse}\label{case-studey-1-2d-sand-column-collapse}

The first case study is the classical column-collapse configuration on the 2D-Sand dataset: compact granular bodies are released under gravity, undergo free fall and impact, spread along the floor, and settle into a final deposit. TRACE is trained on the 1000 training trajectories with the two-stage protocol of Section~\ref{training-strategy}. The first stage trains single-step prediction under teacher forcing. Within each training window, the contact memory is created and updated in the same way as at inference. Small Gaussian noise perturbs the input positions and velocities. The loss is the normalized Huber error of the predicted accelerations. The second stage fine-tunes the model through autoregressive rollouts. The model is first unrolled from a reference state without supervision and is then supervised on particle positions over the following steps. This stage trains the model to correct its own accumulated errors.

At test time, the model predicts accelerations, the semi-implicit Euler integrator advances the state under the constraints of Section~\ref{integration}, and the contact graph and edge memories evolve step by step. We evaluate TRACE on all 30 test trajectories, unrolling 300 steps of each 320-step reference trajectory, and report results under the two protocols of Section~\ref{evaluation-protocol}.

\subsubsection{Rollout accuracy}\label{rollout-accuracy}

\begin{figure}[htbp]
\centering
\includegraphics[width=\linewidth]{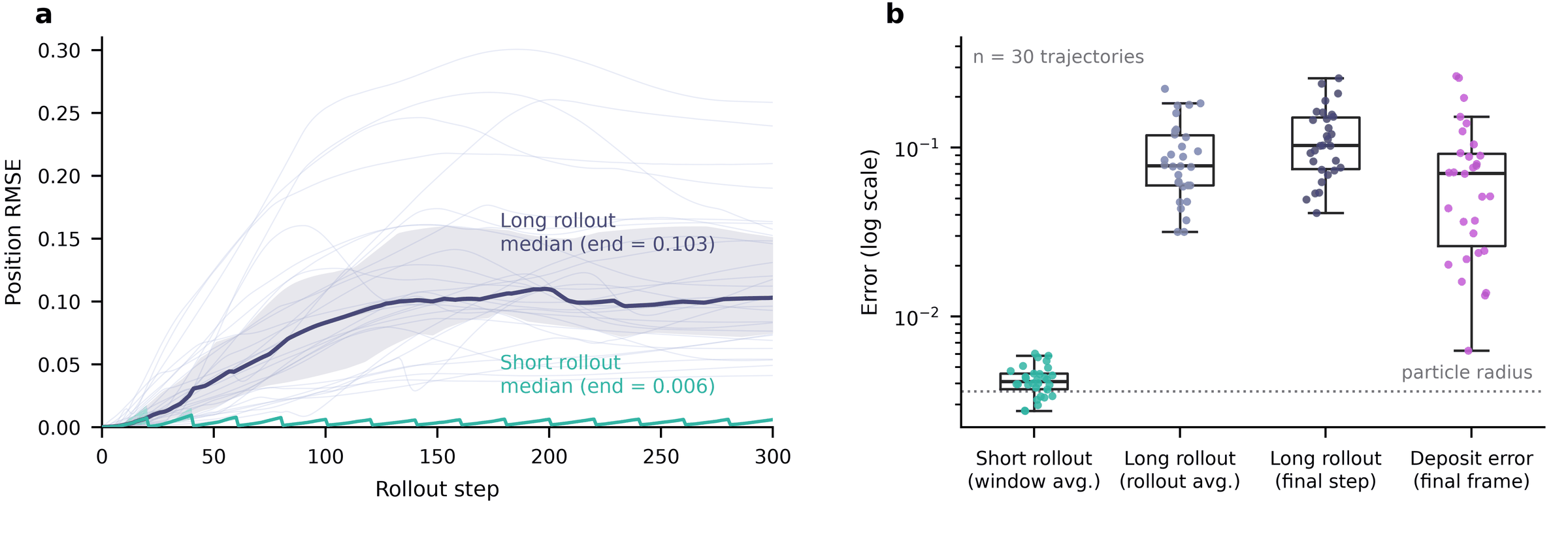}
\caption{Accuracy on the 2D-Sand test set under the two rollout protocols. (a) Time-resolved position RMSE(t). Curves show the median, shaded bands the interquartile range, and thin lines individual long rollouts. (b) Window-averaged short-rollout RMSE, rollout-averaged long-rollout RMSE, final-step RMSE, and deposit error.}
\label{fig:2d-accuracy}
\end{figure}

We first evaluate the position accuracy of TRACE under the two protocols on all 30 test trajectories. Figure~\ref{fig:2d-accuracy} summarizes the results. Panel (a) shows the position RMSE at every rollout step. Panel (b) shows the per-trajectory distribution of the four summary metrics.

In Figure~\ref{fig:2d-accuracy} (a), the two thick curves show the median position RMSE over the 30 test trajectories, one for each protocol. The shaded bands show the interquartile range. The thin light curves show the individual long rollouts. The short-rollout median stays near the bottom of the axis over the full horizon. Its window-averaged RMSE is 0.0042 on average (maximum 0.0060), comparable to the assigned particle radius. The long-rollout median grows during the transient flow, peaks at about 0.110 around step 200, and decreases slightly as the material settles. The individual curves spread widely around the median. The reduction after the peak indicates that the model partially corrects the accumulated drift as the system approaches its final equilibrium.

Figure~\ref{fig:2d-accuracy} (b) each dot is one test trajectory and each column is one accuracy metric. The boxes show the median and the interquartile range, and the dotted line marks the assigned particle radius. Short-rollout errors cluster tightly around the particle radius. Long-rollout errors span a continuous range from 0.032 to 0.224 without a distinct failure cluster. The rollout-averaged RMSE is 0.093 on average (median 0.078). The final-step RMSE is 0.118 on average (median 0.103). The correlation with particle count is weak (Pearson $r$ = 0.23). Prediction difficulty therefore depends mainly on highly energetic ejection events, not on scene size. The deposit error stays low (mean 0.078, median 0.071), so the predicted final deposits closely match the ground-truth geometry. The gap between the two protocols shows that long-horizon errors come mainly from autoregressive error accumulation.

\subsubsection{Physical consistency}\label{physical-consistency}

\begin{figure}[htbp]
\centering
\includegraphics[width=\linewidth]{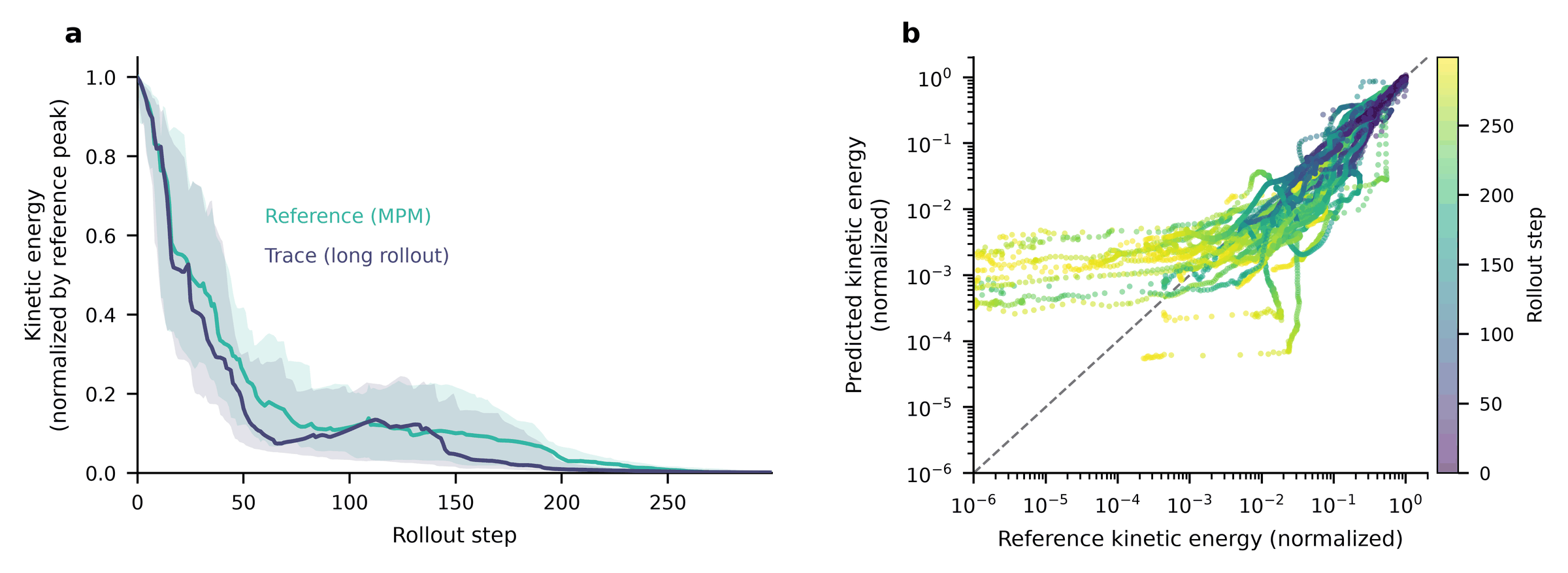}
\caption{Kinetic-energy consistency on the 2D-Sand test set. (a) Kinetic-energy evolution normalized by the per-trajectory reference peak. Lines show medians over the 30 test trajectories and bands the interquartile range. (b) Step-wise parity of predicted against reference kinetic energy, colored by rollout step. The dashed line marks parity.}
\label{fig:2d-energy}
\end{figure}

We next verify the kinetic-energy evolution of the long rollout against the MPM ground truth. The kinetic energy of the prediction and of the ground truth is computed and normalized as defined in Section~\ref{physical-consistency-diagnostics}. Figure~\ref{fig:2d-energy} shows the result.

In Figure~\ref{fig:2d-energy} (a), compares the kinetic-energy decay of the prediction with the ``ground truth'' reference MPM. Each curve is the median over the 30 test trajectories, the bands are the interquartile range, and each trajectory is normalized by its reference peak. The two curves overlap during free fall and impact. During spreading, the predicted energy drops slightly faster. To quantify this, we record the step at which the kinetic energy of each trajectory falls below 2 percent of its peak. The prediction crosses this level earlier than the reference in 27 of the 30 trajectories, and about 30 steps earlier on average.

Figure~\ref{fig:2d-energy} (b) plots the predicted against the ground-truth kinetic energy, one point per trajectory per step, colored by rollout step. The dashed line is the one-to-one line. At energies above one tenth of the peak, the points lie on the line. At intermediate energies the points sit slightly below it. Downward streaks below the line show steps where the prediction has settled while the reference still moves. The horizontal bands in the lower left show the opposite state. There the reference is essentially at rest, while the predicted energy stays near one thousandth of the peak. The prediction keeps a small residual motion after the reference has stopped.

In short, the prediction loses kinetic energy at almost the same rate as the ``ground truth''. The main deviation is the slightly faster decay during spreading.

\subsubsection{Representative rollout visualizations}\label{representative-rollout-visualizations}

\rev{Renders of all 30 test trajectories, together with animated visualizations of every rollout, are available in the accompanying repository} (\url{https://github.com/Data-Driven-Computational-Geotechnics/TRACE}). To provide a representative view of the model's performance across the accuracy spectrum, we present two trajectories with the highest rollout-averaged RMSE and two with the lowest RMSE in the main text.

\begin{figure}[htbp]
\centering
\includegraphics[width=\linewidth]{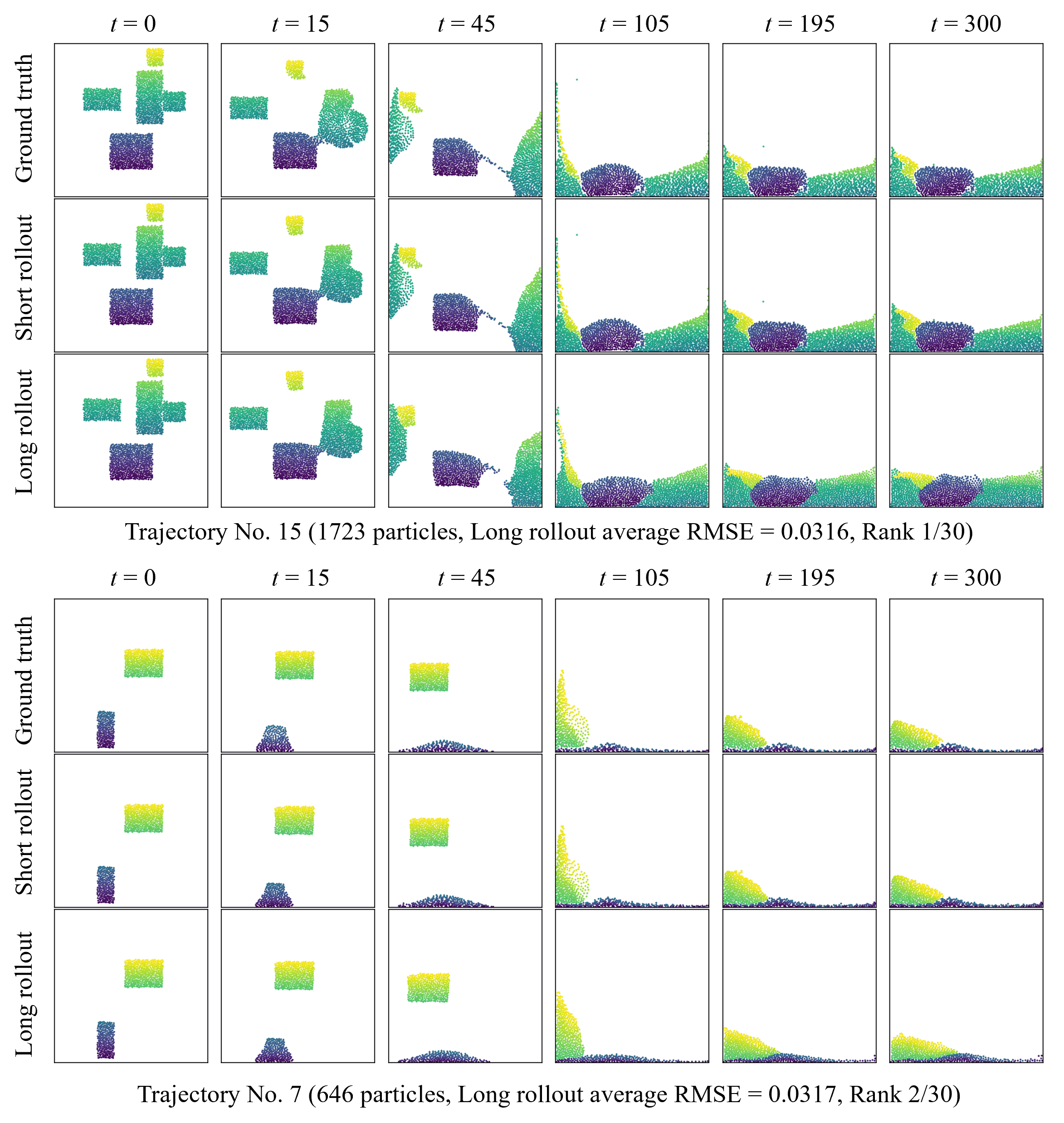}
\caption{The two most accurate test trajectories in 2D-Sand dataset. Within each example, rows show the MPM ``ground truth'', and TRACE's short rollout, and long rollout at six time instants. Particles retain their initial-height color throughout, providing a consistent material tracer of internal transport and layering. The same coloring is used across all rows, with the color bar indicating initial height.}
\label{fig:2d-best}
\end{figure}

Figure~\ref{fig:2d-best} presents the two best-performing trajectories for TRACE's short and long rollout compared against the MPM ``ground truth'' at six representative time instants, with particles colored according to their initial height. Trajectories No. 15 and No. 7 achieve rollout-averaged RMSE values of 0.0316 and 0.0317, ranking first and second among all 30 test trajectories, respectively.

Trajectory No. 15 consists of five granular bodies, with four smaller blocks arranged around a larger central block. After release, the surrounding blocks collapse and flow around the central body, which remains largely intact while the displaced material merges into a single connected deposit. Trajectory No. 7 contains two granular bodies positioned at different heights. The lower column collapses first to form a small mound, after which the elevated block impacts it and spreads into a left-leaning deposit.

In both trajectories, the long rollout closely follows the ground-truth dynamics throughout the entire evolution. The predicted final deposits agree well with the MPM reference in terms of runout distance, deposit height, and free-surface profile, demonstrating the model's ability to accurately capture both transient motion and final equilibrium configurations.

\begin{figure}[htbp]
\centering
\includegraphics[width=\linewidth]{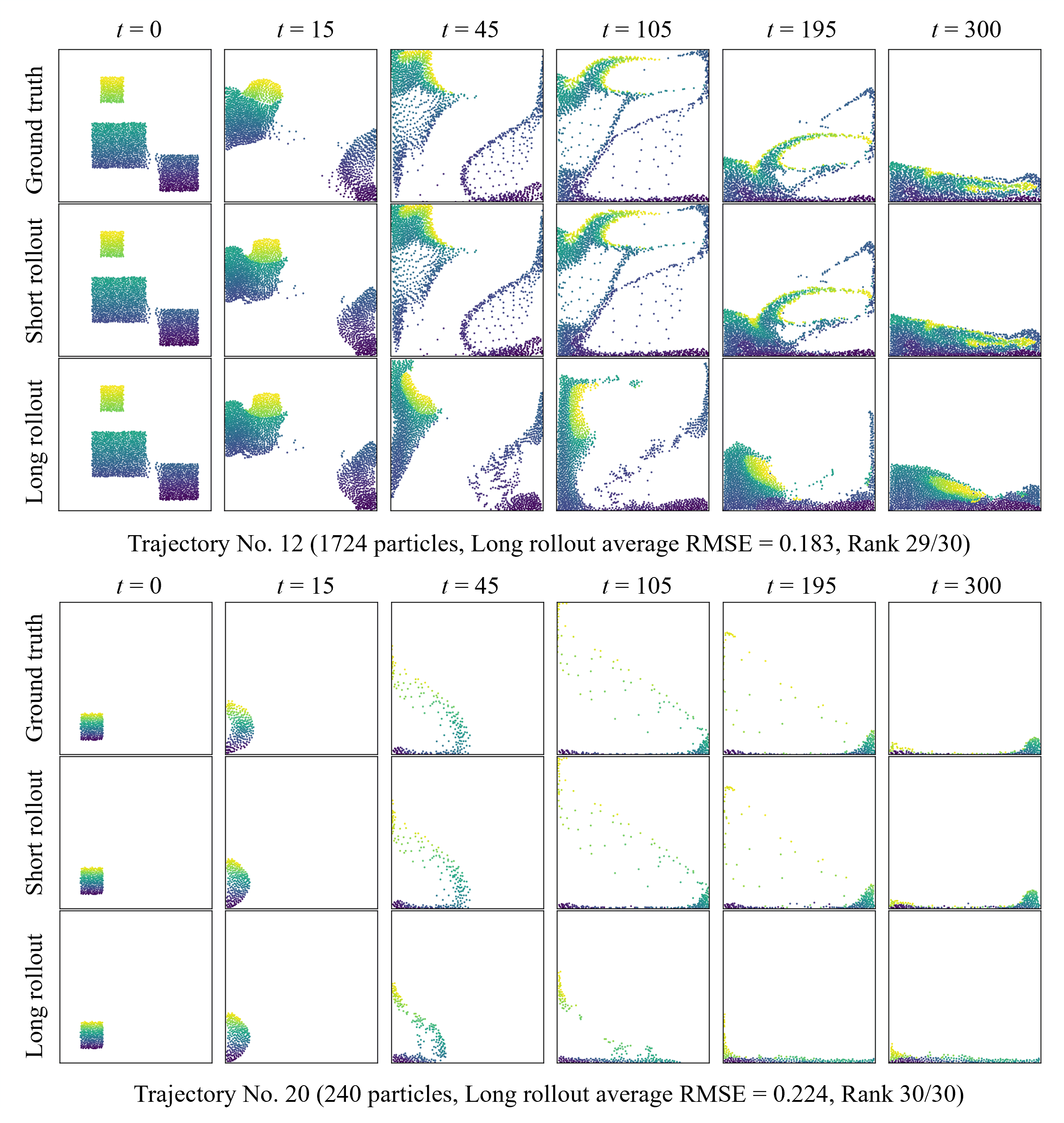}
\caption{The two least accurate test trajectories in 2D-Sand dataset. Layout and coloring as in Figure~\ref{fig:2d-best}}
\label{fig:2d-worst}
\end{figure}

Figure~\ref{fig:2d-worst} presents the two lowest-ranked trajectories, No. 12 and No. 20, with rollout-averaged RMSE values of 0.183 and 0.224, ranking 29th and 30th among the 30 test trajectories, respectively.

Trajectory No. 12 consists of three granular bodies whose collapse generates a sustained arching wave that curls over before impact. The short rollout captures this complex vortex-like motion well, whereas the long rollout underestimates both the strength of the ejection and the subsequent material dispersion. As a result, the arch collapses prematurely, more material accumulates near the left wall, and the predicted final deposit reproduces the overall slope of the ground-truth profile.

Trajectory No. 20 contains a single small body that is launched across nearly the entire container, making it the most energetic scenario in the test set relative to its size. Here, the long rollout again predicts a weaker ejection than observed in the ground truth. The material follows a lower trajectory, settles earlier along the floor, and fails to form the large deposit accumulated against the right wall in the reference simulation.

Despite their different initial configurations, both least accurate cases exhibit the same underlying error mechanism. When a large fraction of the material becomes airborne simultaneously, the model tends to underestimate the intensity of the ejection, and the resulting error propagates through the remainder of the rollout. Nevertheless, both simulations remain numerically stable throughout the long rollout and converge to physically plausible final deposits without divergence.

\subsection{Case study 2: 3D Sand column collapse}\label{case-studey-2-3d-sand-column-collapse}

The second case study evaluates TRACE on 3D-Sand. Each scene contains 4975 to 19762 particles, making the system roughly one order of magnitude larger on average than its two-dimensional counterpart. The contact geometry is also fully three-dimensional. This setting therefore evaluates both the predictive accuracy of TRACE in three dimensions and the scalability of the memory-management mechanism in Section~\ref{edge-memory} to a substantially larger number of active contact states.

The network architecture, loss functions, and two-stage training protocol remain unchanged from the 2D study. Three implementation adaptations are introduced to make training at this scale computationally feasible. Neighbor searches are performed blockwise outside the differentiation graph, gradient checkpointing is used to recompute message-passing activations during backpropagation, and training is distributed across multiple GPUs using data parallelism. The training configurations are detailed in Appendix A.1.

Evaluation follows the long-rollout protocol defined in Section~\ref{evaluation-protocol} and covers all 100 test trajectories over a horizon of 300 steps. Because long autoregressive rollouts represent the intended inference setting and provide the more demanding evaluation at this scale, the 3D-Sand study focuses exclusively on this protocol.

\subsubsection{Accuracy and physical consistency}\label{accuracy-and-physical-consistency}

\begin{figure}[htbp]
\centering
\includegraphics[width=\linewidth]{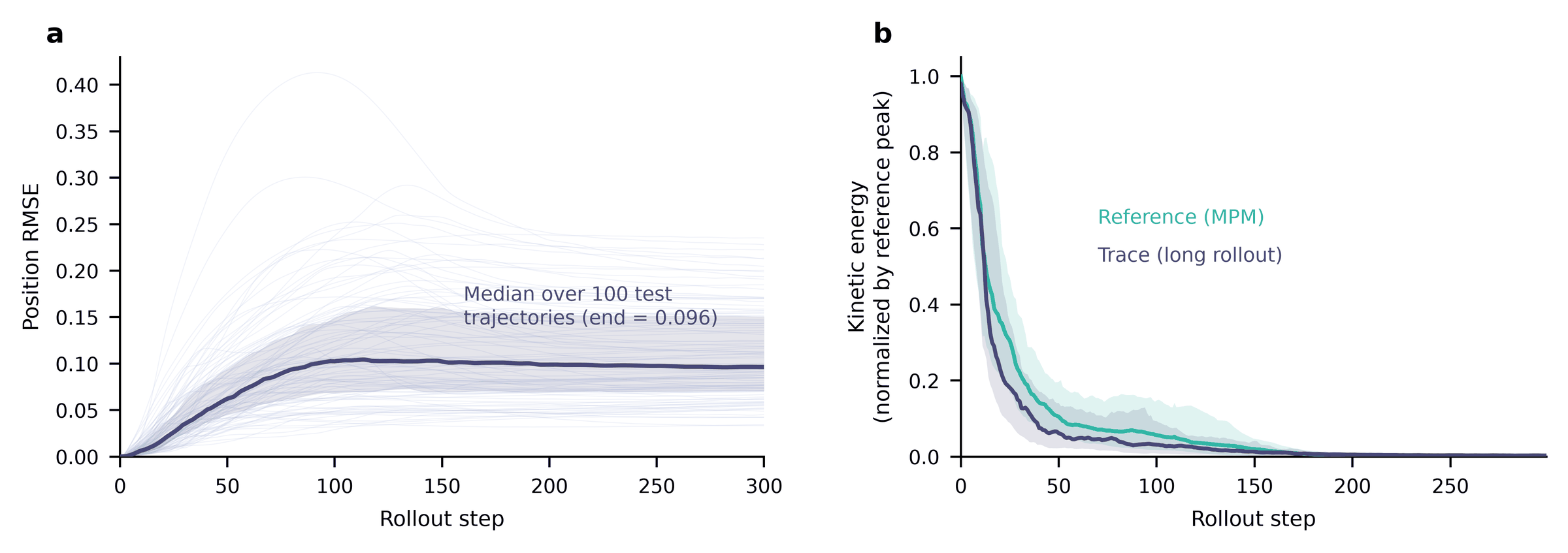}
\caption{Accuracy and kinetic-energy consistency on the 3D-Sand test set. (a) Per-step position RMSE. (b) Kinetic energy normalized by the per-trajectory reference peak.}
\label{fig:3d-accuracy}
\end{figure}

This subsection examines the accuracy and the physical consistency of the long rollouts on the 100 test trajectories. Figure~\ref{fig:3d-accuracy} shows the results: position error in panel (a) and kinetic energy in panel (b).

In Figure~\ref{fig:3d-accuracy} (a), the thin curves are the 100 individual test trajectories, the thick curve is their median, and the band is the interquartile range. The median error grows during the collapse, peaks at 0.104 near step 113, and declines to 0.096 as the material settles. All 100 rollouts stay bounded over the full horizon. The final-step RMSE has a mean of 0.111 and a median of 0.096. These values are close to their 2D counterparts, with ten times more particles per scene.

In Figure~\ref{fig:3d-accuracy} (b), the curves compare the kinetic-energy decay of the prediction with the reference. For each trajectory, we divide the peak of the predicted kinetic-energy curve by the peak of the reference curve. This gives one ratio per trajectory, and the median of the 100 ratios is 0.992. The decay through impact and spreading follows the reference, with the same slightly faster dissipation during spreading as in 2D. At late times the two curves separate. After the reference has come to rest, the predicted kinetic energy settles near $3\times10^{-3}$ of the peak.

To provide a more straightforward statistical characterization, Table~\ref{tab:3d-error-dist} summarizes the distribution of position errors across the 100 test trajectories. The mean and standard deviation (s.d.) characterize the average error and its variability across trajectories. The median and interquartile range (IQR) describe the central tendency and spread of the distribution, with the IQR reported as the 25th--75th percentile interval. The 95\% confidence interval (CI) quantifies the uncertainty in the estimated mean.

\begin{table}[htbp]
\centering
\setlength{\tabcolsep}{4pt}
\caption{Distribution of the position error across the 100 3D-Sand test trajectories under the long-rollout protocol. Errors are reported in normalized domain units.}
\label{tab:3d-error-dist}
\small
\begin{tabular}{lccccc}
\toprule
\textbf{Metric} & \textbf{Mean $\pm$ s.d.} & \textbf{95\% CI} & \textbf{Median} & \textbf{IQR} & \textbf{Range} \\
\midrule
Rollout-averaged RMSE & 0.099\textsubscript{$\pm$0.050} & [0.089, 0.109] & 0.088 & [0.059, 0.129] & [0.032, 0.266] \\
Final-step RMSE & 0.111\textsubscript{$\pm$0.051} & [0.101, 0.121] & 0.097 & [0.071, 0.151] & [0.033, 0.235] \\
\bottomrule
\end{tabular}
\end{table}

For both metrics in Table~\ref{tab:3d-error-dist}, the mean exceeds the median, indicating a right-skewed error distribution. Most trajectories are predicted accurately, while a few strongly ejecting collapses form the upper tail, reaching 0.266. Rapid contact rearrangements during collapse amplify one-step errors, especially during violent impacts, while the dissipative settling stage limits further error growth. The narrow 95\% confidence intervals (about $\pm$0.010) indicate that the reported means are estimated precisely, while the standard deviations capture the variation across scenes.

\subsubsection{Scaling with scene size}\label{scaling-with-scene-size}

\begin{figure}[htbp]
\centering
\includegraphics[width=\linewidth]{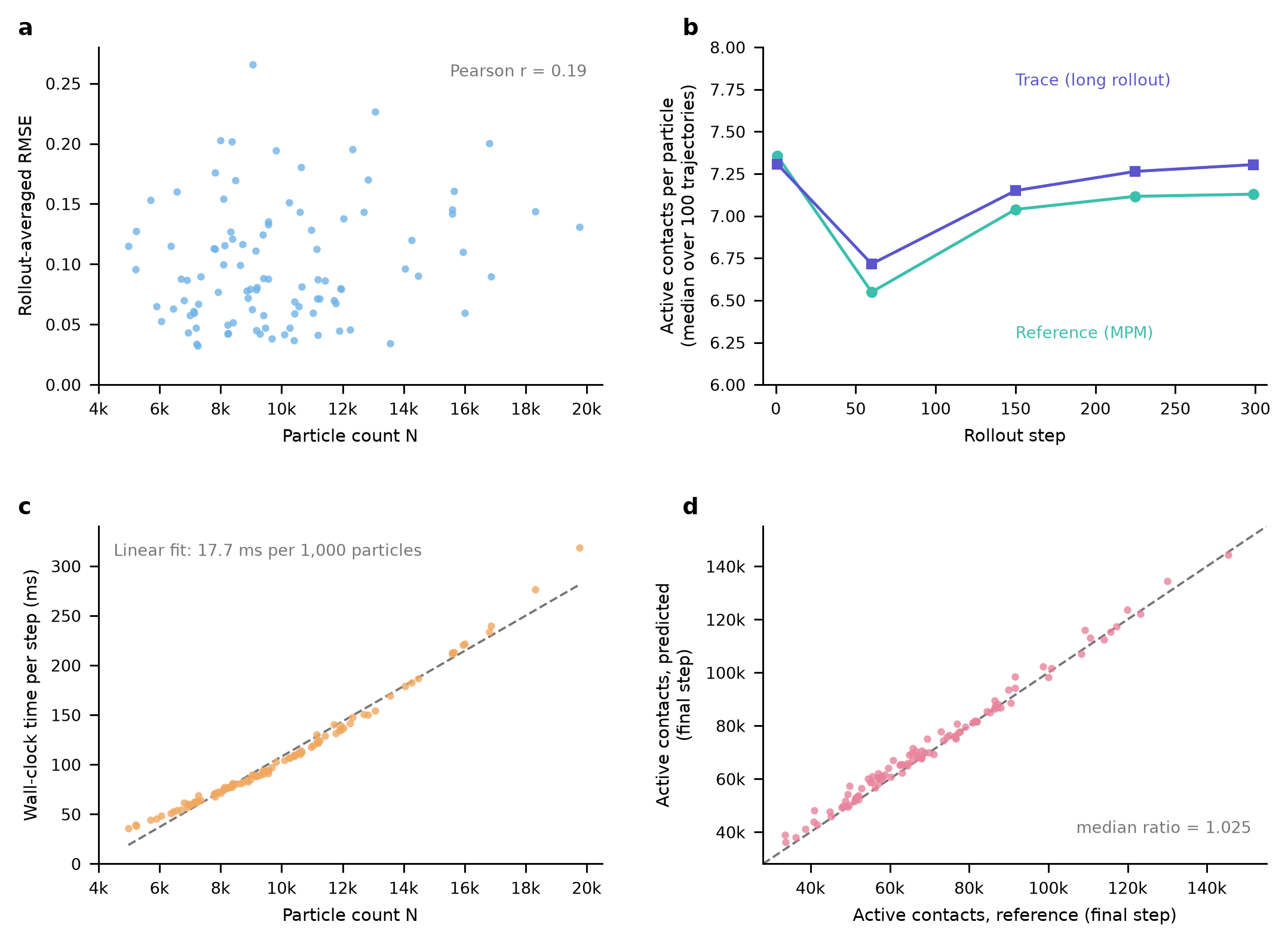}
\caption{Scaling with scene size on the 3D-Sand test set. (a) Rollout-averaged RMSE against particle count. (b) Median active contacts per particle at five sampled rollout steps. (c) Wall-clock time per step against particle count. (d) Predicted against reference number of active contacts at the final step.}
\label{fig:3d-scaling}
\end{figure}

The 100 test scenes span 4975 to 19762 particles, a factor of four. Figure~\ref{fig:3d-scaling} shows the accuracy, the number of active contacts, and the runtime for these scenes.

In Figure~\ref{fig:3d-scaling} (a), each dot is one test scene. The rollout-averaged RMSE correlates only weakly with the particle count (Pearson $r$ = 0.19). In Figure~\ref{fig:3d-scaling} (b), the two lines show the median number of active contacts per particle at five sampled steps, for the reference and for the prediction. Both lines start near 7.4, dip by about eleven percent during the collapse, and recover to 7.1 as the pile settles. The two lines stay close at all five steps.

In Figure~\ref{fig:3d-scaling} (c), the runtime grows linearly with the particle count. One rollout step takes 108 ms on average, and the per-step cost follows a linear trend of 17.7 ms per thousand particles (Pearson $r$ = 0.99). The largest test scene completes its 300-step rollout in 96 seconds on a single RTX 4090 GPU. In Figure~\ref{fig:3d-scaling} (d), each dot compares the predicted and the reference number of active contacts at the final step, and the dashed line marks equality. The dots lie close to the line. For each scene, we divide the predicted contact count by the reference count. This gives one ratio per scene, and the median of the 100 ratios is 1.025.

\subsubsection{Long rollout prediction results}\label{long-rollout-prediction-results}

Per-trajectory results for all 100 test trajectories are reported in Appendix A.3 and are also available in the code repository. Due to space limitations, this subsection presents only the two best-ranked and two least-ranked trajectories to illustrate the range of model performance.

\begin{figure}[htbp]
\centering
\includegraphics[width=\linewidth]{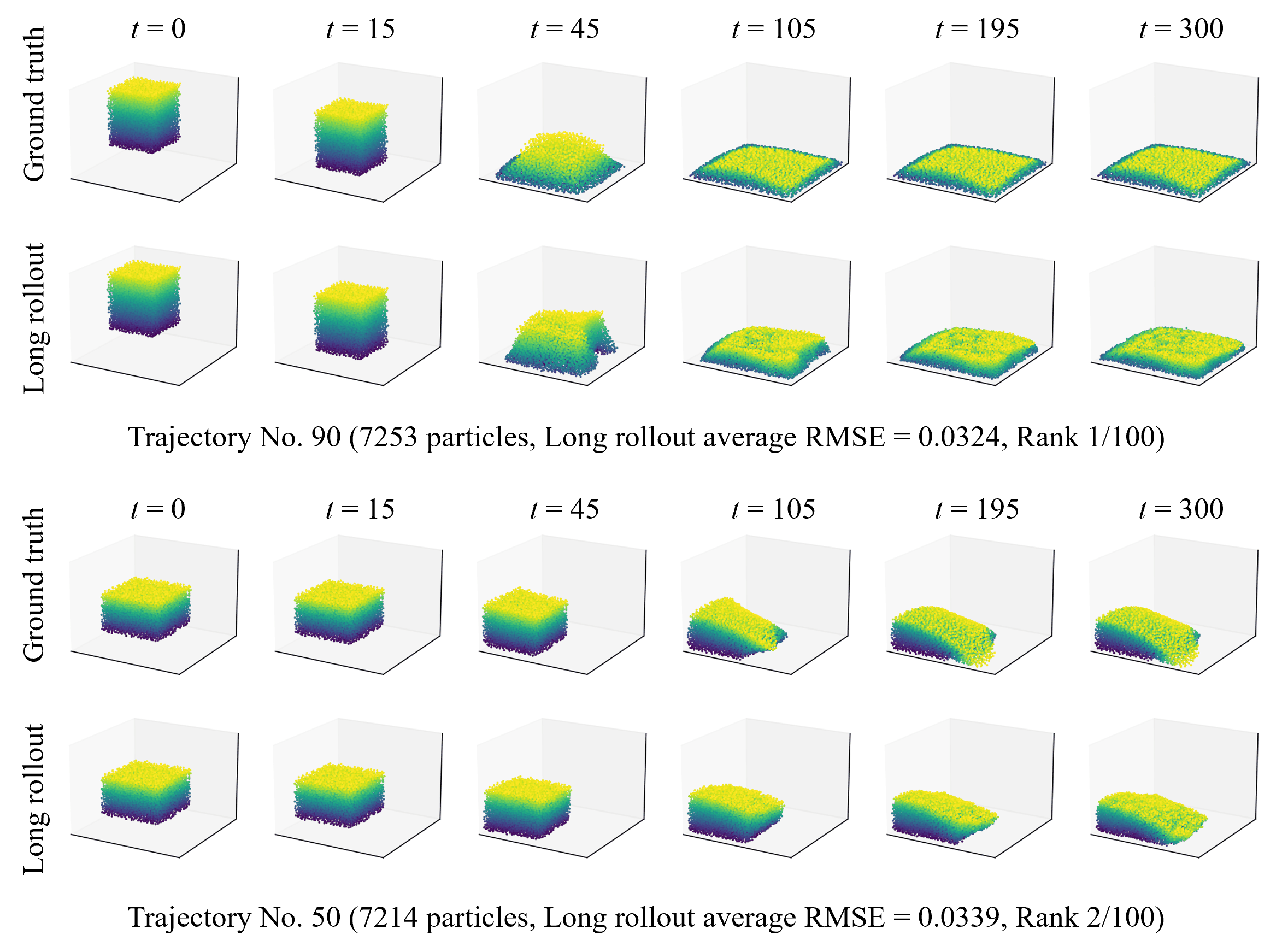}
\caption{The two most accurate test trajectories in 3D-Sand dataset}
\label{fig:3d-best}
\end{figure}

Figure~\ref{fig:3d-best} shows the two most accurate test trajectories. Trajectory No. 90 (7,253 particles, rollout-averaged RMSE 0.0324) is a single cubic block that collapses symmetrically into a thin, nearly uniform sheet. Trajectory No. 50 (7,214 particles, RMSE 0.0339) is a wide slab that stays coherent and spreads gradually toward one side of the domain. In both cases the prediction is close to indistinguishable from the reference at every instant, and the final deposits match in extent, height, and surface profile.

\begin{figure}[htbp]
\centering
\includegraphics[width=\linewidth]{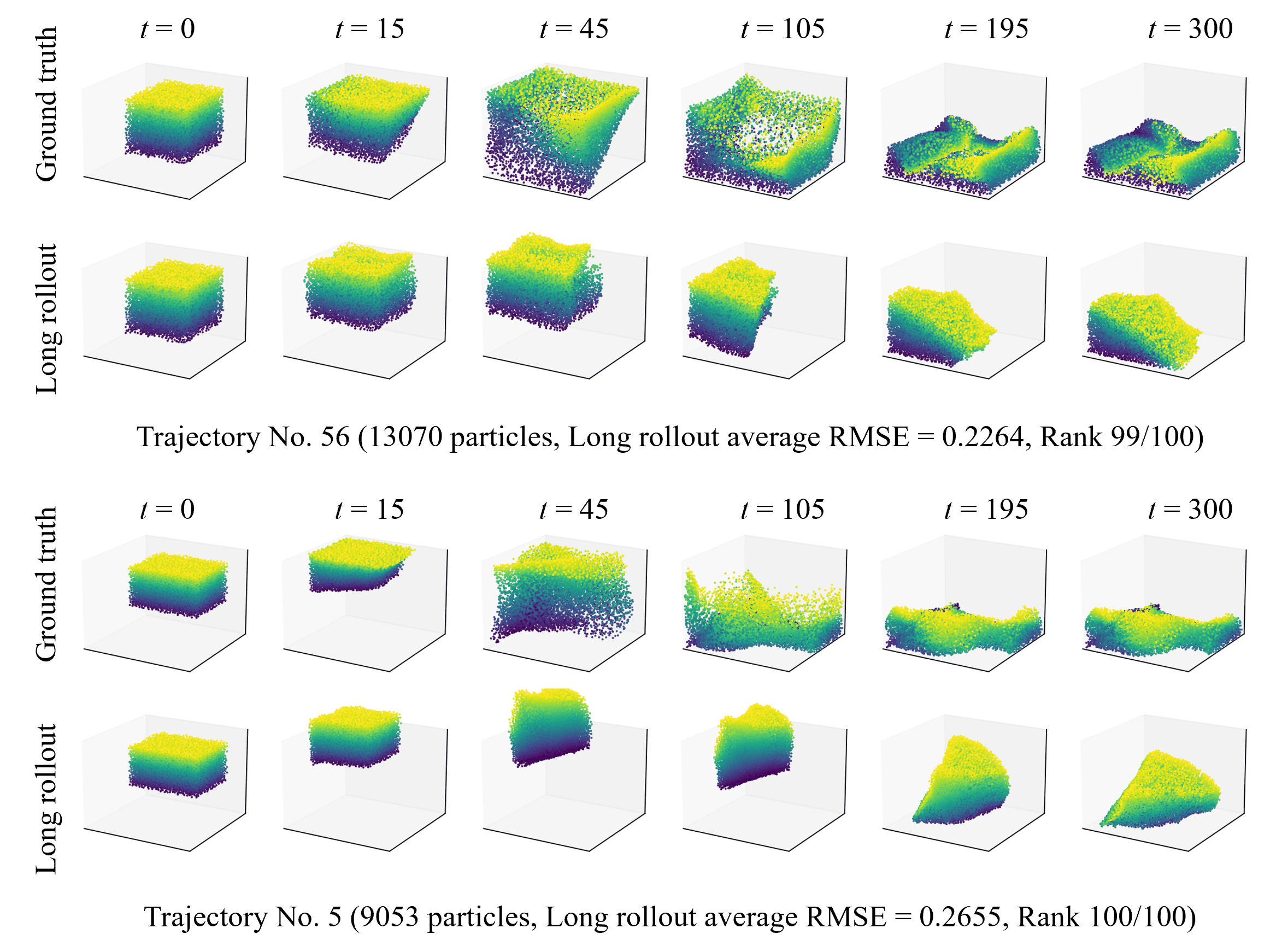}
\caption{The two least accurate test trajectories in 3D-Sand dataset}
\label{fig:3d-worst}
\end{figure}

Figure~\ref{fig:3d-worst} shows the two least accurate trajectories, and both fail through the mechanism already identified in 2D. In trajectory No. 5 (9,053 particles, RMSE 0.2655) the reference collapse launches a large fraction of the material into a dispersed airborne splash that settles into a broad, low deposit. The prediction keeps the material too coherent. It slides into a compact heap instead of dispersing, and the final deposit is too tall and too narrow. Trajectory No. 56 (13,070 particles, RMSE 0.2264) behaves the same way at larger scale. The reference disperses strongly at impact and settles into a broad deposit with a central ridge, while the prediction under-ejects and remains too compact. In both cases the deposits still settle into stable piles and neither rollout diverges. The under-estimated splash is therefore the dominant failure mode in both dimensions, and we return to its likely origin in the Discussion.

\section{Discussion}\label{discussion}

We compare TRACE with two representative graph-based learned simulators that use different approaches to incorporating temporal information:

\textbf{GNS} (Graph Network Simulator), which incorporates recent motion history through the node input features but does not maintain a recurrent state across simulation steps. Our re-implementation follows the original architecture, with ten message-passing blocks, 128-dimensional latent representations, and a node-level acceleration decoder.

\textbf{NMGNS} (Node-Memory Graph Neural Simulator), where node-memory simulators represent temporal history through recurrent hidden states attached to particles. These node states are propagated across consecutive simulation steps, while the interaction edges and their features are reconstructed from the current particle configuration at each step. Inspired by this node-memory paradigm \citep{zhao2025physicalinformationflowconstrained}, we develop a node-memory baseline under the same experimental framework, with each node maintaining a recurrent hidden state over time and the edge representations being recomputed at every step. We evaluate TRACE against two baselines using the long-rollout protocol described in Section~\ref{evaluation-protocol}. For a controlled comparison, both baselines are re-implemented within our framework. Table~\ref{tab:baseline-comparison} reports the results averaged over multiple random seeds for each simulator to assess sensitivity to initialization.

\begin{table}[htbp]
\centering
\caption{Comparison of predictive accuracy with baseline graph-network simulators}
\label{tab:baseline-comparison}
\small
\begin{tabular}{llcccc}
\toprule
Dataset & Model & Parameters & Average RMSE$\downarrow$ & Final RMSE$\downarrow$ & Deposit error$\downarrow$ \\
\midrule
2D-Sand & GNS & 1.29M & 0.159\textsubscript{$\pm$0.004} & 0.198\textsubscript{$\pm$0.015} & 0.333\textsubscript{$\pm$0.064} \\
 & NMGNS & 1.43M & 0.191\textsubscript{$\pm$0.011} & 0.267\textsubscript{$\pm$0.039} & 0.497\textsubscript{$\pm$0.183} \\
 & TRACE(Ours) & 1.08M & \textbf{0.110\textsubscript{$\pm$0.018}} & \textbf{0.135\textsubscript{$\pm$0.020}} & \textbf{0.094\textsubscript{$\pm$0.020}} \\
\midrule
3D-Sand & GNS & 1.30M & 0.151\textsubscript{$\pm$0.005} & 0.212\textsubscript{$\pm$0.013} & 0.367\textsubscript{$\pm$0.065} \\
 & NMGNS & 1.43M & 0.252\textsubscript{$\pm$0.004} & 0.334\textsubscript{$\pm$0.007} & 1.361\textsubscript{$\pm$0.366} \\
 & TRACE(Ours) & 1.08M & \textbf{0.097\textsubscript{$\pm$0.002}} & \textbf{0.109\textsubscript{$\pm$0.003}} & \textbf{0.154\textsubscript{$\pm$0.001}} \\
\bottomrule
\end{tabular}
\end{table}

TRACE achieves consistently competitive performance than both baseline graph-network simulators across all datasets and evaluation metrics. On the 2D-Sand dataset, TRACE reduces the average RMSE by 30.8\% and 42.4\% compared with GNS and NMGNS, respectively. The final RMSE is reduced by 31.8\% and 49.4\% relative to the two baselines, indicating that TRACE better controls error accumulation during rollout. Notably, TRACE achieves a substantial reduction in deposit error, improving the final deposit morphology prediction by 71.8\% and 81.1\% compared with GNS and NMGNS.

Similar trends are observed in the 3D-Sand dataset. TRACE reduces the average RMSE by 36.0\% and 61.5\%, and the final RMSE by 48.6\% and 67.4\% compared with GNS and NMGNS, respectively. For deposit error, TRACE achieves reductions of 58.0\% and 88.7\%. These results suggest that the proposed spatiotemporal contact memory mechanism effectively preserves interaction information during dynamic graph reconstruction, reducing error accumulation and improving the stability of long-term granular flow simulation.

\begin{figure}[htbp]
\centering
\includegraphics[width=\linewidth]{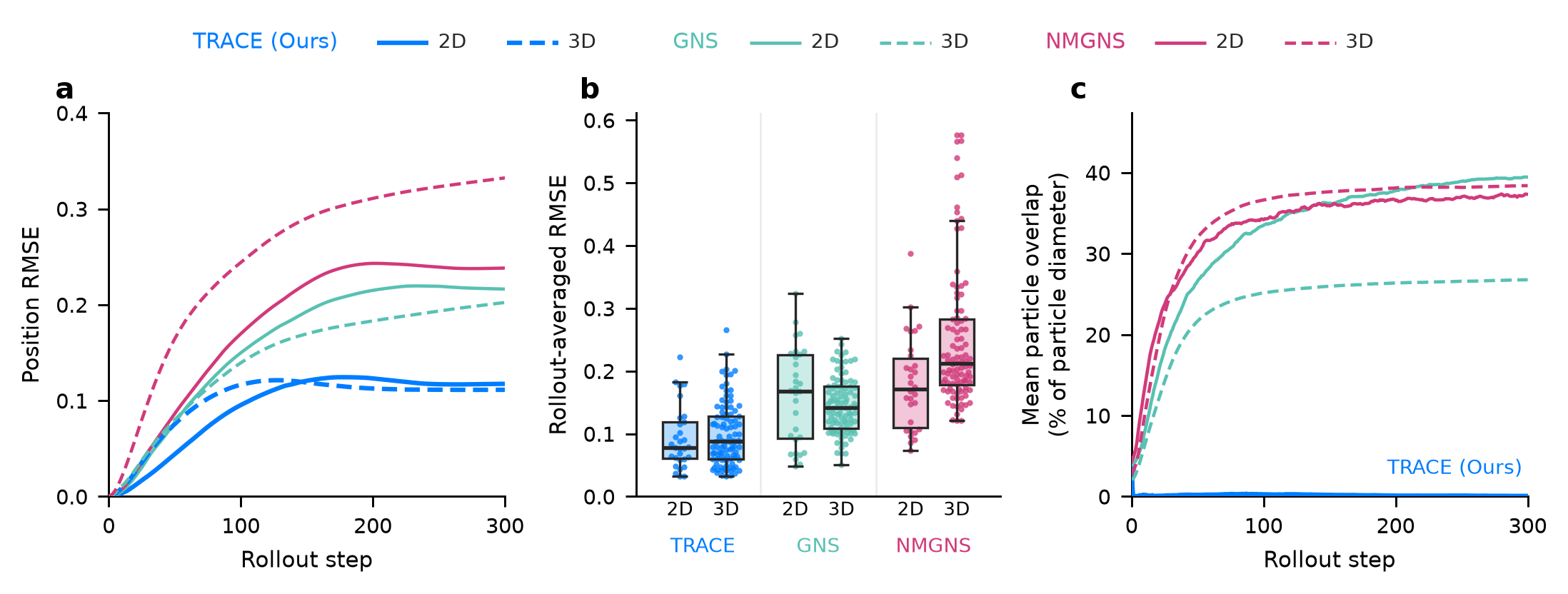}
\caption{Behavior of the three simulators over the full rollout on the dataset. (a) Position RMSE against rollout step. (b) Distribution of the rollout-averaged RMSE across the scenes. (c) Mean particle overlap, as a percentage of a particle diameter, against rollout step.}
\label{fig:baseline-rollout}
\end{figure}

The aggregate metrics provide only a summary of the rollout performance. Figure~\ref{fig:baseline-rollout} provides a more detailed comparison of the three models across the 2D and 3D datasets, focusing on rollout accuracy, error distribution, and physical validity.

Figure~\ref{fig:baseline-rollout}(a) shows the position RMSE over the rollout step, with solid lines representing the 2D-Sand results and dashed lines representing the 3D-Sand results. In the 2D scenes, both baselines quickly reach error plateaus of 0.22--0.24, whereas TRACE converges to a substantially lower level of about 0.12, corresponding to an approximately 50\% reduction in error. In the 3D scenes, this advantage is even more evident. The errors of both baselines continue to increase throughout the rollout, reaching 0.20 for GNS and 0.33 for NMGNS at the final step, with neither curve reaching a stable error level within the evaluated rollout. However, TRACE maintains an RMSE of approximately 0.12 throughout the latter part of the rollout. Despite an approximately tenfold increase in particle number from 2D to 3D, the TRACE curves remain nearly identical, indicating minimal performance degradation as scene complexity increases.

Figure~\ref{fig:baseline-rollout} (b) further confirms this advantage at the trajectory level. TRACE produces a lower and tighter error distribution, with median RMSEs of 0.08 and 0.09 in 2D and 3D, respectively. The corresponding medians are 0.17 for both baselines in 2D, and 0.14 and 0.21 for GNS and NMGNS in 3D. Moreover, every quartile of the TRACE distribution lies below the corresponding quartile of both baselines in both dimensions, indicating that the improvement is consistent across trajectories.

Figure~\ref{fig:baseline-rollout} (c) measures particle interpenetration and shows the effect of the hard geometric constraints of Section~\ref{integration}. The metric is computed in three steps. At each rollout step, we collect every pair of particles whose center distance is smaller than the collision diameter, which equals twice the assigned particle radius. For each such pair, the overlap depth is the shortfall of the center distance below the collision diameter, expressed as a fraction of that diameter. The curve plots the mean overlap depth over these pairs at each step, averaged over all the test scenes. Because sand grains are treated as rigid particles, large overlaps indicate physically unrealistic predictions.

TRACE applies the non-penetration projection of Section~\ref{integration} at every step. The projection separates every overlapping pair, so the Trace curve stays near zero over the entire rollout. The two baselines follow their original formulations, which contain no such projection. Their overlap climbs steadily and settles between 27\% and 39\% of a diameter, so on average an overlapping pair is driven more than a third of a diameter into each other and the baseline packing is compressed well beyond the collision radius.

Beyond prediction accuracy, computational efficiency is another important consideration for practical granular flow simulation. Compared with conventional numerical solvers such as MPM, graph neural network-based simulators avoid iterative solution procedures by directly predicting particle dynamics through learned surrogate models, offering substantially lower inference costs.

\begin{table}[htbp]
\centering
\caption{Computational efficiency compared with MPM (All methods run on a single RTX 4090).}
\label{tab:cost}
\setlength{\tabcolsep}{5pt}
\small
\begin{tabular}{lllll}
\toprule
Dataset & Method & Time/frame (ms) $\downarrow$ & Time/trajectory (s) $\downarrow$ & Speedup over MPM ($\times$) $\uparrow$ \\
\midrule
2D-Sand & MPM & 146.22\textsubscript{$\pm$2.91} & 43.87\textsubscript{$\pm$0.87} & 1.0 \\
 & TRACE(Ours) & \textbf{12.01\textsubscript{$\pm$0.61}} & \textbf{3.60\textsubscript{$\pm$0.18}} & \textbf{12.19\textsubscript{$\pm$0.71}} \\
\midrule
3D-Sand & MPM & 957.31\textsubscript{$\pm$30.21} & 287.19\textsubscript{$\pm$9.06} & 1.0 \\
 & TRACE(Ours) & \textbf{107.63\textsubscript{$\pm$53.75}} & \textbf{32.29\textsubscript{$\pm$16.13}} & \textbf{8.90\textsubscript{$\pm$4.47}} \\
\bottomrule
\end{tabular}
\end{table}

Table~\ref{tab:cost} summarizes the computational cost of TRACE and MPM on the test scenes using a single RTX 4090 GPU. In 2D, TRACE requires only 12.01 ms per frame, compared with 146.22 ms for MPM, corresponding to a 12.2-fold speedup and an approximately 91.8\% reduction in per-frame computational cost. Over a 300-frame trajectory, TRACE reduces the computational time from 43.87 s to 3.60 s, again corresponding to a 91.8\% reduction.

The computational advantage persists in 3D, where TRACE requires 107.63 ms per frame compared with 957.31 ms for MPM. This corresponds to an 8.9-fold speedup and an approximately 88.8\% reduction in per-frame computational cost. At the trajectory level, TRACE reduces the computational time from 287.19 s to 32.29 s, saving approximately 4.2 min per trajectory. The relatively large standard deviation of the TRACE timing in 3D (53.75 ms per frame and 16.13 s per trajectory) reflects greater variation in computational cost across the more complex scenes. Despite this variability, TRACE retains nearly an order-of-magnitude speedup over MPM.

Overall, TRACE reduces the computational cost by approximately 89--92\% and achieves 12.2-fold and 8.9-fold speedups in 2D and 3D, respectively. Combined with the rollout accuracy reported above, these results demonstrate that TRACE substantially reduces the computational cost of long-horizon granular flow simulation while maintaining accurate predictions.

\section{Conclusion}\label{conclusion}

We presented TRACE (spatio\textbf{T}emporal memo\textbf{R}y \textbf{A}cross \textbf{C}ontact \textbf{E}dge), a graph network simulator for granular flow. Each contact edge maintains a persistent memory state, with attention pooling aggregating neighborhood information and a GRU updating the state over time. To address the changing identities caused by dynamic contact-graph reconstruction, this study introduces an identity dictionary that preserves contact memory as the graph evolves. A physics-structured decoder then predicts contact forces and particle accelerations, while enforcing Coulomb friction and equal-and-opposite forces to preserve momentum. During state integration, non-penetration and boundary projections prevent particle interpenetration and keep particles within the computational domain. Together, these designs ensure stable and physically consistent long-horizon rollouts.

We systematically evaluate TRACE on both the 2D and 3D granular column-collapse benchmarks. Both datasets use MPM simulations as the reference solutions, with particular emphasis on the long-rollout protocol, in which only the initial state is provided and the entire trajectory is generated autoregressively. On the 2D benchmark, all test trajectories complete the long rollout without divergence. TRACE performs well in terms of kinematic accuracy, final deposit morphology, and physical consistency. The rollout-averaged position RMSE is 0.093, while the combined deviation in runout and deposit height of the final deposit remains below 8\% of the domain size. The overall kinetic-energy evolution agrees well with the MPM reference, particularly during the free-fall and impact stages, with only slightly faster energy dissipation during the later spreading stage. On the 3D benchmark, the average number of particles per scene is approximately 10 times that of the 2D cases, yet TRACE maintains a rollout-averaged RMSE of 0.099, comparable to that in 2D. All test trajectories again remain stable throughout the long rollout. The median ratios of the predicted peak kinetic energy and final active contact count to the corresponding reference values are 0.992 and 1.025, respectively, indicating that both the energy-release process and the contact microstructure of the final deposit are well reproduced. In the largest test scene, TRACE simultaneously maintains memory for more than $1.4 \times 10^{5}$ contact edges, while the per-step computational cost scales approximately linearly with particle count, at about 17.7 ms per 1000 particles. The computational cost therefore remains well controlled as the problem size increases.

We further compare TRACE with the standard GNS and the node-memory simulator NMGNS within the same implementation framework. Based on the mean values, TRACE reduces the rollout-averaged position RMSE on the 2D benchmark by 30.8\% and 42.4\% relative to GNS and NMGNS, respectively, while reducing the deposit error by 71.8\% and 81.1\%. On the 3D benchmark, the corresponding reductions are 36.0\% and 61.5\% for the average RMSE, and 58.0\% and 88.7\% for the deposit error. Meanwhile, enabled by the physics-structured decoder and non-penetration projection, TRACE maintains particle overlap close to zero throughout the rollout, whereas the two baselines exhibit overlaps of approximately 27\%--39\% of the particle diameter. These results suggest that explicitly storing interaction history on contact edges preserves key local interaction information as the contact graph is repeatedly reconstructed. This helps suppress error accumulation during long rollouts while maintaining accurate predictions and physically consistent behavior across both 2D and 3D settings.

Furthermore, the ability to explicitly model interaction history provides TRACE with broader potential beyond granular flow simulation. The proposed method offers a general strategy for learning physical interactions in particle-based systems. The edge-based contact memory mechanism could be extended to other complex dynamical processes, including fluid--particle interactions, debris flows, landslides, fracture evolution, and multiphase material transport, where the history of local interactions plays an important role. Furthermore, efficient physics simulators such as TRACE could provide valuable physical feedback for embodied intelligence. By integrating learned simulators into robotic training environments, agents could interact with physically realistic worlds rather than relying solely on visual observations. For example, quadruped robots learning locomotion on deformable terrains such as sand or soil could benefit from particle-based simulations that capture terrain response, reducing the sim-to-real gap and enabling more physically grounded reinforcement learning.

\section*{Acknowledgment}

The first author is grateful for the support from the University of Melbourne's Melbourne Research Scholarship (MRS). Partial support from the Australian Department of Education's AEA IV240100185 is also acknowledged.

\section*{Authors contribution}

Changjian Zhou: Conceptualization, Methodology, Investigation, Visualization, Software, Writing - original draft, Writing - review \& editing. Negin Yousefpour: Conceptualization, Resources, Writing - original draft, Writing - review \& editing. Jie Qi: Conceptualization, Writing - review \& editing. Junfeng Fang: Investigation, Writing - original draft, Writing - review \& editing. Guillermo A. Narsilio: Resources, Visualization, Writing - review \& editing. Hans Petter Jostad: Writing -- original draft, Writing - review \& editing.

\section*{Data availability}

All data and code are available at \url{https://github.com/Data-Driven-Computational-Geotechnics/TRACE}.

\section*{Declaration of Competing Interest}

The authors declare that they have no known competing financial interests or personal relationships that could have appeared to influence the work reported in this paper.

\appendix
\setcounter{table}{0}\renewcommand{\thetable}{A.\arabic{table}}
\setcounter{figure}{0}\renewcommand{\thefigure}{A.\arabic{figure}}
\setcounter{equation}{0}\renewcommand{\theequation}{A.\arabic{equation}}

\section{Datasets and qualitative results}\label{appendix-a}

\subsection{Benchmark detail and training setup}

We evaluate TRACE on the granular Sand benchmark of Sanchez-Gonzalez et al., which has become a standard testbed for learned particle simulators. We use two settings of this benchmark, a two-dimensional dataset (2D-Sand) and a three-dimensional dataset (3D-Sand). Their properties are summarized in Table~\ref{tab:benchmark-properties}.

\begin{table}[htbp]
\centering
\caption{Properties of the two Sand benchmark datasets.}
\label{tab:benchmark-properties}
\small
\begin{tabular}{@{}lll@{}}
\toprule
Property & 2D-Sand & 3D-Sand \\
\midrule
Spatial dimension & 2 & 3 \\
Ground-truth solver & MPM & MPM \\
Train/validation/test trajectories & 1000/30/30 & 1000/100/100 \\
Particles per scene (full dataset) & 107 $\sim$ 1976 & 4391 $\sim$ 19762 \\
Mean particles per scene & $\approx 1190$ & $\approx 10000$ \\
Particles per scene (test split) & 212 $\sim$ 1971 & 4975 $\sim$ 19762 \\
Steps per trajectory & 320 & 350 \\
Time step $\Delta t$ & $2.5\times 10^{-3}$ & $2.5\times 10^{-3}$ \\
Simulation domain & $[0.1, 0.9]^{2}$ & $[0.2,0.8]^{3}$ \\
Boundary condition & Non-periodic walls & Non-periodic walls \\
Connectivity radius & 0.015 & 0.025 \\
Assigned particle radius & $3.6\times 10^{-3}$ & $6.8\times 10^{-3}$ \\
Material types & Sand & Sand \\
Ground-truth contact forces & Not provided & Not provided \\
\bottomrule
\end{tabular}
\end{table}

\textbf{Physical configuration.} Both datasets describe dry sand in closed, non-periodic domains with a fixed time step of ($\Delta t = 2.5\times 10^{-3}$). 2D-Sand uses the domain ($[0.1,0.9]^{2}$) over 320 steps, whereas 3D-Sand uses ($[0.2,0.8]^{3}$) over 350 steps. All spatial quantities are expressed in normalized domain units.

\textbf{Dataset scale.} 2D-Sand contains 1000 training, 30 validation, and 30 test trajectories, with 107 to 1976 particles per scene. 3D-Sand contains 1000 training, 100 validation, and 100 test trajectories, with 4391 to 19762 particles per scene. Randomized initial configurations produce diverse scenarios, ranging from a single collapsing body to multiple bodies undergoing mid-air collisions.

\textbf{Preprocessing.} Each trajectory stores particle positions only. Velocities and accelerations are reconstructed by finite differences, and acceleration targets are normalized using training-set statistics. Because the MPM data provide neither contact forces nor particle radii, TRACE learns contact interactions without force supervision. Uniform particle radii of ($3.6\times 10^{-3}$) in 2D and ($6.8\times 10^{-3}$) in 3D are used for graph construction, with neighborhood radii of 0.015 and 0.025, respectively.

To ensure a consistent comparison between the 2D and 3D settings, both models are trained using the same architecture and optimization framework. The 2D and 3D models use a latent dimension of 128, an edge-memory dimension of 16, eight message-passing rounds, and 1.08 million trainable parameters.

Stage 1 minimizes the mean squared error between the predicted and ground-truth normalized accelerations using AdamW and trains the model for single-step prediction over windows of 15 consecutive frames. Stage 2 fine-tunes the model through constrained rollouts, consisting of a drift phase of up to 150 steps followed by 12 supervised steps. The two stages use 512 and 256 training trajectories for 2D-Sand, and 128 and 64 trajectories for 3D-Sand, respectively. Training is completed within approximately one day for each model, using a single NVIDIA RTX 4090 GPU for 2D-Sand and 4 parallel RTX 4090 GPUs for 3D-Sand. The remaining hyperparameters for both stages are reported in Table~\ref{tab:hyperparams}

\begin{table}[htbp]
\centering
\caption{Training hyperparameters for TRACE on 2D-Sand and 3D-Sand.}
\label{tab:hyperparams}
\small
\begin{tabular}{@{}lll@{}}
\toprule
Hyper-parameter & 2D-Sand & 3D-Sand \\
\midrule
Optimization & \multicolumn{2}{>{\raggedright\arraybackslash}p{0.55\linewidth}@{}}{AdamW (weight decay $10^{-6}$), gradient clipping at norm 1.0, linear warmup then cosine decay} \\
Batch size & 1 window per step & 4 windows per step (one per GPU) \\
Stage 1: input noise std & $4\times 10^{-4}$ & $2.5\times 10^{-4}$ \\
Stage 1: learning rate & $3\times 10^{-4} \to 3\times 10^{-6}$ & $3\times 10^{-4} \to 3\times 10^{-6}$ \\
Stage 2: input noise std & $2\times 10^{-4}$ & $1.2\times 10^{-4}$ \\
Stage 2: learning rate & $5\times 10^{-5} \to 1\times 10^{-6}$ & $5\times 10^{-5} \to 1\times 10^{-6}$ \\
\bottomrule
\end{tabular}
\end{table}

\subsection{Statistical results of 2D-Sand Dataset}

Table~\ref{tab:2d-per-traj} reports the per-trajectory results of TRACE on all 30 test trajectories of the 2D-Sand dataset, ordered by trajectory index.

\begin{longtable}{@{}
  >{\small\centering\arraybackslash}p{0.10\textwidth}
  >{\small\centering\arraybackslash}p{0.12\textwidth}
  >{\small\centering\arraybackslash}p{0.16\textwidth}
  >{\small\centering\arraybackslash}p{0.16\textwidth}
  >{\small\centering\arraybackslash}p{0.15\textwidth}
  >{\small\centering\arraybackslash}p{0.15\textwidth}@{}}
\caption{Per-trajectory results of TRACE on the 2D-Sand test set (30 trajectories)}\label{tab:2d-per-traj}\\
\toprule
Trajectory No. & Particles number & Short-rollout RMSE ($\times 10^{-2}$) & Long-rollout RMSE ($\times 10^{-2}$) & Final RMSE ($\times 10^{-2}$) & Deposit error ($\times 10^{-2}$) \\
\midrule
\endfirsthead
\toprule
Trajectory No. & Particles number & Short-rollout RMSE ($\times 10^{-2}$) & Long-rollout RMSE ($\times 10^{-2}$) & Final RMSE ($\times 10^{-2}$) & Deposit error ($\times 10^{-2}$) \\
\midrule
\endhead
\bottomrule
\endlastfoot
1 & 1730 & 0.43 & 8.86 & 8.37 & 2.38 \\
2 & 1471 & 0.45 & 10.16 & 14.81 & 5.13 \\
3 & 1314 & 0.47 & 17.94 & 15.74 & 15.26 \\
4 & 1517 & 0.40 & 6.91 & 7.64 & 1.38 \\
5 & 722 & 0.33 & 7.71 & 14.59 & 12.50 \\
6 & 1882 & 0.5 & 11.95 & 15.24 & 26.61 \\
7 & 646 & 0.33 & 3.17 & 4.92 & 1.61 \\
8 & 1910 & 0.39 & 7.72 & 10.29 & 3.11 \\
9 & 1852 & 0.42 & 11.59 & 11.22 & 0.63 \\
10 & 1650 & 0.46 & 7.78 & 10.27 & 7.81 \\
11 & 1241 & 0.38 & 9.10 & 9.62 & 8.02 \\
12 & 1724 & 0.51 & 18.31 & 19.01 & 3.70 \\
13 & 1551 & 0.44 & 16.07 & 23.95 & 8.86 \\
14 & 1036 & 0.30 & 6.27 & 11.76 & 5.15 \\
15 & 1723 & 0.39 & 3.17 & 4.11 & 7.10 \\
16 & 1456 & 0.42 & 9.53 & 12.03 & 7.14 \\
17 & 1112 & 0.37 & 6.21 & 5.35 & 2.18 \\
18 & 1148 & 0.34 & 4.79 & 10.31 & 25.99 \\
19 & 582 & 0.46 & 4.35 & 7.41 & 4.36 \\
20 & 240 & 0.58 & 22.37 & 25.83 & 3.65 \\
21 & 212 & 0.28 & 5.97 & 9.29 & 19.73 \\
22 & 881 & 0.43 & 5.97 & 6.26 & 2.03 \\
23 & 1711 & 0.60 & 17.77 & 20.97 & 8.96 \\
24 & 1321 & 0.32 & 4.77 & 5.40 & 1.33 \\
25 & 1084 & 0.40 & 8.44 & 8.31 & 2.45 \\
26 & 1817 & 0.40 & 7.89 & 13.10 & 13.96 \\
27 & 1275 & 0.57 & 12.81 & 16.19 & 7.64 \\
28 & 404 & 0.28 & 3.71 & 7.32 & 9.29 \\
29 & 1971 & 0.55 & 12.60 & 16.35 & 10.45 \\
30 & 921 & 0.40 & 5.90 & 6.90 & 7.01 \\
\midrule
mean & 1270 & 0.42 & 9.33 & 11.75 & 7.85 \\
median & 1318 & 0.41 & 7.83 & 10.30 & 7.06 \\
\end{longtable}

For each trajectory it lists the particle count and five accuracy measures: the window-averaged short-rollout RMSE, the rollout-averaged and final-step long-rollout RMSE, the deposit error, and the linear-momentum residual of the internal forces. The short-rollout RMSE is uniformly small, between $0.28\times 10^{-2}$ and $0.60\times 10^{-2}$, confirming that single-step dynamics are learned accurately on every scene regardless of its size or violence. The long-rollout errors span a much wider range, from $3.2\times 10^{-2}$ to $22.4\times 10^{-2}$, and this spread reflects the accumulation of error over the autoregressive horizon.

\subsection{Statistical results of 2D-Sand Dataset}

Table~\ref{tab:3d-per-traj} reports the per-trajectory results of TRACE on all 100 test trajectories of the 3D-Sand dataset, ordered by trajectory index.

\begin{longtable}{@{}
  >{\small\centering\arraybackslash}p{0.10\textwidth}
  >{\small\centering\arraybackslash}p{0.12\textwidth}
  >{\small\centering\arraybackslash}p{0.16\textwidth}
  >{\small\centering\arraybackslash}p{0.16\textwidth}
  >{\small\centering\arraybackslash}p{0.15\textwidth}
  >{\small\centering\arraybackslash}p{0.15\textwidth}@{}}
\caption{Per-trajectory results of TRACE on the 3D-Sand test set (100 trajectories)}\label{tab:3d-per-traj}\\
\toprule
Trajectory No. & Particles number & Long-rollout RMSE ($\times 10^{-2}$) & Final RMSE ($\times 10^{-2}$) & KE peak ratio & Active contacts \\
\midrule
\endfirsthead
\toprule
Trajectory No. & Particles number & Long-rollout RMSE ($\times 10^{-2}$) & Final RMSE ($\times 10^{-2}$) & KE peak ratio & Active contacts \\
\midrule
\endhead
\bottomrule
\endlastfoot
1 & 11145 & 11.27 & 11.88 & 0.929 & 80,601 \\
2 & 10404 & 3.65 & 5.17 & 0.998 & 75,133 \\
3 & 5220 & 9.59 & 11.84 & 1.031 & 38,040 \\
4 & 8090 & 15.42 & 19.09 & 1.001 & 59,949 \\
5 & 9053 & 26.55 & 20.46 & 0.993 & 65,472 \\
6 & 9194 & 8.06 & 7.32 & 0.988 & 67,865 \\
7 & 9170 & 7.89 & 7.63 & 0.967 & 65,304 \\
8 & 8229 & 4.94 & 5.24 & 0.983 & 60,706 \\
9 & 5696 & 15.31 & 17.00 & 0.993 & 41,174 \\
10 & 10426 & 5.93 & 7.72 & 0.973 & 76,104 \\
11 & 11172 & 7.14 & 7.93 & 1.343 & 81,674 \\
12 & 6573 & 15.99 & 17.12 & 0.958 & 48,206 \\
13 & 9564 & 13.28 & 15.66 & 1.012 & 70,273 \\
14 & 19762 & 13.08 & 16.20 & 0.971 & 144,379 \\
15 & 11180 & 8.75 & 10.83 & 1.051 & 81,807 \\
16 & 8484 & 16.97 & 20.02 & 1.016 & 61,995 \\
17 & 10272 & 4.73 & 6.29 & 0.976 & 75,698 \\
18 & 8226 & 4.21 & 4.23 & 1.004 & 60,128 \\
19 & 12316 & 19.53 & 22.24 & 1.000 & 89,556 \\
20 & 9469 & 4.70 & 5.47 & 1.046 & 70,552 \\
21 & 8982 & 7.92 & 9.06 & 1.001 & 65,820 \\
22 & 11925 & 7.98 & 10.02 & 1.022 & 86,832 \\
23 & 6063 & 5.27 & 6.18 & 0.864 & 43,851 \\
24 & 6928 & 4.31 & 4.88 & 1.103 & 50,200 \\
25 & 8096 & 9.96 & 12.49 & 0.992 & 60,539 \\
26 & 6808 & 7.02 & 8.49 & 0.965 & 49,195 \\
27 & 9411 & 5.75 & 7.65 & 1.033 & 69,847 \\
28 & 12252 & 4.58 & 5.91 & 1.035 & 88,554 \\
29 & 10090 & 4.19 & 4.52 & 0.972 & 74,444 \\
30 & 12030 & 13.77 & 12.16 & 0.957 & 88,018 \\
31 & 7344 & 8.98 & 10.64 & 1.092 & 53,633 \\
32 & 8370 & 20.17 & 22.8 & 0.985 & 60,844 \\
33 & 8134 & 11.53 & 15.05 & 1.012 & 58,787 \\
34 & 12696 & 14.34 & 14.59 & 0.986 & 93,423 \\
35 & 7827 & 17.60 & 17.13 & 0.969 & 57,244 \\
36 & 8648 & 9.90 & 9.92 & 0.978 & 62,225 \\
37 & 15946 & 11.02 & 14.11 & 0.995 & 115,264 \\
38 & 6705 & 8.78 & 7.96 & 0.996 & 49,667 \\
39 & 15585 & 14.54 & 17.00 & 1.021 & 113,108 \\
40 & 12823 & 17.01 & 18 & 1.009 & 94,117 \\
41 & 16007 & 5.93 & 6.68 & 0.992 & 117,277 \\
42 & 8397 & 5.17 & 5.91 & 0.865 & 60,993 \\
43 & 9807 & 19.45 & 23.52 & 0.985 & 71,535 \\
44 & 13562 & 3.42 & 4.24 & 1.025 & 98,197 \\
45 & 15650 & 16.09 & 17.04 & 0.984 & 115,901 \\
46 & 11028 & 5.93 & 7.75 & 1.166 & 81,062 \\
47 & 16859 & 8.99 & 9.34 & 0.977 & 123,602 \\
48 & 6443 & 6.28 & 7.07 & 0.971 & 47,711 \\
49 & 10657 & 8.12 & 11.06 & 1.031 & 77,330 \\
50 & 7214 & 3.39 & 3.33 & 0.827 & 53,227 \\
51 & 7273 & 6.72 & 7.37 & 0.971 & 52,912 \\
52 & 8328 & 12.70 & 15.20 & 0.997 & 60,581 \\
53 & 16816 & 20.01 & 19.36 & 0.956 & 122,115 \\
54 & 5225 & 12.72 & 13.56 & 1.012 & 38,872 \\
55 & 9179 & 4.53 & 5.81 & 1.187 & 68,190 \\
56 & 13070 & 22.64 & 21.26 & 0.99 & 98,469 \\
57 & 10242 & 15.12 & 16.7 & 1.004 & 75,000 \\
58 & 9407 & 8.84 & 11.67 & 1.026 & 68,696 \\
59 & 10591 & 14.34 & 18.48 & 1.007 & 76,538 \\
60 & 7786 & 11.32 & 13.89 & 0.976 & 56,382 \\
61 & 9680 & 3.81 & 5.57 & 0.982 & 69,242 \\
62 & 7919 & 7.71 & 9.14 & 0.978 & 58,170 \\
63 & 8904 & 7.21 & 8.73 & 0.981 & 64,951 \\
64 & 7129 & 5.93 & 7.16 & 0.909 & 52,410 \\
65 & 6370 & 11.52 & 15.41 & 0.98 & 45,767 \\
66 & 10963 & 12.84 & 15.51 & 1.002 & 79,618 \\
67 & 11893 & 4.49 & 5.79 & 0.786 & 86,835 \\
68 & 9552 & 13.55 & 15.73 & 0.953 & 69,363 \\
69 & 7200 & 4.69 & 6.42 & 1.011 & 52,218 \\
70 & 11253 & 7.13 & 8.54 & 1.086 & 81,560 \\
71 & 4975 & 11.52 & 12.38 & 0.992 & 36,233 \\
72 & 9294 & 4.20 & 5.58 & 1.062 & 67,573 \\
73 & 6993 & 5.73 & 7.04 & 0.939 & 51,800 \\
74 & 9564 & 8.76 & 9.38 & 1.003 & 69,988 \\
75 & 18307 & 14.39 & 14.72 & 1.016 & 134,529 \\
76 & 11719 & 6.97 & 8.46 & 1.108 & 84,922 \\
77 & 8727 & 11.64 & 13.37 & 0.992 & 64,080 \\
78 & 10424 & 6.90 & 8.92 & 0.824 & 75,565 \\
79 & 9157 & 11.13 & 11.22 & 0.904 & 67,101 \\
80 & 7819 & 11.24 & 14.35 & 1.005 & 56,698 \\
81 & 10643 & 18.05 & 19.87 & 1.011 & 77,864 \\
82 & 11188 & 4.14 & 5.72 & 0.976 & 81,685 \\
83 & 14053 & 9.61 & 11.28 & 0.976 & 102,196 \\
84 & 7999 & 20.28 & 21.34 & 0.99 & 58,988 \\
85 & 11775 & 6.74 & 8.75 & 1.004 & 86,273 \\
86 & 7111 & 6.09 & 7.00 & 0.93 & 51,390 \\
87 & 11951 & 7.95 & 8.16 & 0.991 & 87,521 \\
88 & 9384 & 12.45 & 13.14 & 1.004 & 68,906 \\
89 & 8378 & 12.11 & 11.00 & 1.002 & 61,644 \\
90 & 7253 & 3.24 & 3.43 & 0.951 & 54,168 \\
91 & 8857 & 7.78 & 9.18 & 1.002 & 65,457 \\
92 & 9034 & 6.23 & 6.12 & 0.992 & 66,790 \\
93 & 6906 & 8.66 & 9.36 & 0.952 & 49,432 \\
94 & 10561 & 6.51 & 7.43 & 0.993 & 77,672 \\
95 & 15596 & 14.16 & 15.88 & 0.9 & 112,365 \\
96 & 14260 & 11.98 & 14.33 & 0.931 & 101,671 \\
97 & 14479 & 9.02 & 7.39 & 0.981 & 107,040 \\
98 & 11414 & 8.64 & 11.16 & 0.993 & 85,510 \\
99 & 5903 & 6.49 & 7.09 & 0.976 & 42,683 \\
100 & 8254 & 4.25 & 5.10 & 1.000 & 59,955 \\
\midrule
mean & 9967 & 9.89 & 11.13 & 0.992 & 72,827 \\
median & 9396 & 8.75 & 9.65 & 0.992 & 68,801 \\
\end{longtable}

Here, we report the long-rollout results for the 3D test trajectories. For each trajectory, the table lists the particle count, the rollout-averaged and final-step position RMSE, the ratio of predicted to reference peak kinetic energy, the number of active contacts at the final step, and the wall-clock time for the full 300-step rollout.

The peak kinetic-energy ratio remains close to one across the test set, with a median of 0.992, indicating that the model accurately captures the energy released during collapse without excessive numerical gain or dissipation. In the largest scenes, the number of active contacts exceeds ($1.4\times 10^{5}$), corresponding to the same number of live edge-memory states. The memory-management mechanism described in Section~\ref{edge-memory} tracks these contacts reliably throughout the rollout. The computational time increases approximately linearly with particle count, ranging from about 11 seconds for the smallest scenes to 96 seconds for the largest.

\bibliography{references}

@article{cundall1979discrete,
  author  = {Cundall, P.A. and Strack, O.D.L.},
  title   = {A discrete numerical model for granular assemblies},
  journal = {G\'eotechnique},
  volume  = {29},
  pages   = {47--65},
  year    = {1979},
  doi     = {10.1680/geot.1979.29.1.47}
}

@article{gingold1977smoothed,
  author  = {Gingold, R.A. and Monaghan, J.J.},
  title   = {Smoothed particle hydrodynamics: theory and application to non-spherical stars},
  journal = {Monthly Notices of the Royal Astronomical Society},
  volume  = {181},
  pages   = {375--389},
  year    = {1977},
  doi     = {10.1093/mnras/181.3.375}
}

@article{lucy1977numerical,
  author  = {Lucy, L.B.},
  title   = {A numerical approach to the testing of the fission hypothesis},
  journal = {The Astronomical Journal},
  volume  = {82},
  pages   = {1013--1024},
  year    = {1977},
  doi     = {10.1086/112164}
}

@article{sulsky1994particle,
  author  = {Sulsky, D. and Chen, Z. and Schreyer, H.L.},
  title   = {A particle method for history-dependent materials},
  journal = {Computer Methods in Applied Mechanics and Engineering},
  volume  = {118},
  pages   = {179--196},
  year    = {1994},
  doi     = {10.1016/0045-7825(94)90112-0}
}

@article{sulsky1995application,
  author  = {Sulsky, D. and Zhou, S.-J. and Schreyer, H.L.},
  title   = {Application of a particle-in-cell method to solid mechanics},
  journal = {Computer Physics Communications},
  volume  = {87},
  pages   = {236--252},
  year    = {1995},
  doi     = {10.1016/0010-4655(94)00170-7}
}

@article{soga2016trends,
  author  = {Soga, K. and Alonso, E. and Yerro, A. and Kumar, K. and Bandara, S.},
  title   = {Trends in large-deformation analysis of landslide mass movements with particular emphasis on the material point method},
  journal = {G\'eotechnique},
  volume  = {66},
  pages   = {248--273},
  year    = {2016},
  doi     = {10.1680/jgeot.15.LM.005}
}

@article{guo2015discrete,
  author  = {Guo, Y. and Curtis, J.S.},
  title   = {Discrete element method simulations for complex granular flows},
  journal = {Annual Review of Fluid Mechanics},
  volume  = {47},
  pages   = {21--46},
  year    = {2015},
  doi     = {10.1146/annurev-fluid-010814-014644}
}

@article{coetzee2017review,
  author  = {Coetzee, C.J.},
  title   = {Review: calibration of the discrete element method},
  journal = {Powder Technology},
  volume  = {310},
  pages   = {104--142},
  year    = {2017},
  doi     = {10.1016/j.powtec.2017.01.015}
}

@article{lino2023current,
  author  = {Lino, M. and Fotiadis, S. and Bharath, A.A. and Cantwell, C.D.},
  title   = {Current and emerging deep-learning methods for the simulation of fluid dynamics},
  journal = {Proceedings of the Royal Society A: Mathematical, Physical and Engineering Sciences},
  volume  = {479},
  pages   = {20230058},
  year    = {2023},
  doi     = {10.1098/rspa.2023.0058}
}

@article{lu2021machine,
  author  = {Lu, L. and Gao, X. and Dietiker, J.F. and Shahnam, M. and Rogers, W.A.},
  title   = {Machine learning accelerated discrete element modeling of granular flows},
  journal = {Chemical Engineering Science},
  volume  = {245},
  pages   = {116832},
  year    = {2021},
  doi     = {10.1016/j.ces.2021.116832}
}

@inproceedings{battaglia2016interaction,
  author    = {Battaglia, P.W. and Pascanu, R. and Lai, M. and Rezende, D. and Kavukcuoglu, K.},
  title     = {Interaction networks for learning about objects, relations and physics},
  booktitle = {Advances in Neural Information Processing Systems (NeurIPS)},
  volume    = {29},
  pages     = {4502--4510},
  year      = {2016},
  url       = {https://proceedings.neurips.cc/paper_files/paper/2016/hash/3147da8ab4a0437c15ef51a5cc7f2dc4-Abstract.html}
}

@misc{battaglia2018relational,
  author       = {Battaglia, P.W. and Hamrick, J.B. and Bapst, V. and Sanchez-Gonzalez, A. and Zambaldi, V. and Malinowski, M. and Tacchetti, A. and Raposo, D. and Santoro, A. and Faulkner, R. and Gulcehre, C. and Song, F. and Ballard, A. and Gilmer, J. and Dahl, G. and Vaswani, A. and Allen, K. and Nash, C. and Langston, V. and Dyer, C. and Heess, N. and Wierstra, D. and Kohli, P. and Botvinick, M. and Vinyals, O. and Li, Y. and Pascanu, R.},
  title        = {Relational inductive biases, deep learning, and graph networks},
  howpublished = {arXiv:1806.01261},
  year         = {2018},
  doi          = {10.48550/arXiv.1806.01261}
}

@inproceedings{sanchezgonzalez2018graph,
  author    = {Sanchez-Gonzalez, A. and Heess, N. and Springenberg, J.T. and Merel, J. and Riedmiller, M. and Hadsell, R. and Battaglia, P.},
  title     = {Graph networks as learnable physics engines for inference and control},
  booktitle = {International Conference on Machine Learning (ICML)},
  series    = {Proceedings of Machine Learning Research},
  volume    = {80},
  pages     = {4470--4479},
  year      = {2018},
  url       = {https://proceedings.mlr.press/v80/sanchez-gonzalez18a.html}
}

@inproceedings{sanchezgonzalez2020learning,
  author    = {Sanchez-Gonzalez, A. and Godwin, J. and Pfaff, T. and Ying, R. and Leskovec, J. and Battaglia, P.},
  title     = {Learning to simulate complex physics with graph networks},
  booktitle = {International Conference on Machine Learning (ICML)},
  series    = {Proceedings of Machine Learning Research},
  volume    = {119},
  pages     = {8459--8468},
  year      = {2020},
  url       = {https://proceedings.mlr.press/v119/sanchez-gonzalez20a.html}
}

@inproceedings{pfaff2021learning,
  author    = {Pfaff, T. and Fortunato, M. and Sanchez-Gonzalez, A. and Battaglia, P.W.},
  title     = {Learning mesh-based simulation with graph networks},
  booktitle = {International Conference on Learning Representations (ICLR)},
  year      = {2021},
  url       = {https://openreview.net/forum?id=roNqYL0_XP}
}

@inproceedings{li2019learning,
  author    = {Li, Y. and Wu, J. and Tedrake, R. and Tenenbaum, J.B. and Torralba, A.},
  title     = {Learning particle dynamics for manipulating rigid bodies, deformable objects, and fluids},
  booktitle = {International Conference on Learning Representations (ICLR)},
  year      = {2019},
  url       = {https://openreview.net/forum?id=rJgbSn09Ym}
}

@inproceedings{ummenhofer2020lagrangian,
  author    = {Ummenhofer, B. and Prantl, L. and Thuerey, N. and Koltun, V.},
  title     = {Lagrangian fluid simulation with continuous convolutions},
  booktitle = {International Conference on Learning Representations (ICLR)},
  year      = {2020},
  url       = {https://openreview.net/forum?id=B1lDoJSYDH}
}

@inproceedings{wu2022learning,
  author    = {Wu, T. and Wang, Q. and Zhang, Y. and Ying, R. and Cao, K. and Sosi\v{c}, R. and Jalali, R. and Hamam, H. and Maucec, M. and Leskovec, J.},
  title     = {Learning large-scale subsurface simulations with a hybrid graph network simulator},
  booktitle = {ACM SIGKDD Conference on Knowledge Discovery and Data Mining (KDD)},
  pages     = {4184--4194},
  year      = {2022},
  doi       = {10.1145/3534678.3539045}
}

@inproceedings{satorras2021en,
  author    = {Satorras, V.G. and Hoogeboom, E. and Welling, M.},
  title     = {{E(n)} equivariant graph neural networks},
  booktitle = {International Conference on Machine Learning (ICML)},
  series    = {Proceedings of Machine Learning Research},
  volume    = {139},
  pages     = {9323--9332},
  year      = {2021},
  url       = {https://proceedings.mlr.press/v139/satorras21a.html}
}

@inproceedings{brandstetter2022geometric,
  author    = {Brandstetter, J. and Hesselink, R. and van der Pol, E. and Bekkers, E.J. and Welling, M.},
  title     = {Geometric and physical quantities improve {E(3)} equivariant message passing},
  booktitle = {International Conference on Learning Representations (ICLR)},
  year      = {2022},
  url       = {https://openreview.net/forum?id=_xwr8gOBeV1}
}

@inproceedings{huang2022equivariant,
  author    = {Huang, W. and Han, J. and Rong, Y. and Xu, T. and Sun, F. and Huang, J.},
  title     = {Equivariant graph mechanics networks with constraints},
  booktitle = {International Conference on Learning Representations (ICLR)},
  year      = {2022},
  url       = {https://openreview.net/forum?id=SHbhHHfePhP}
}

@inproceedings{han2022learning,
  author    = {Han, J. and Huang, W. and Ma, H. and Li, J. and Tenenbaum, J. and Gan, C.},
  title     = {Learning physical dynamics with subequivariant graph neural networks},
  booktitle = {Advances in Neural Information Processing Systems (NeurIPS)},
  volume    = {35},
  pages     = {26256--26268},
  year      = {2022},
  doi       = {10.52202/068431-1904}
}

@inproceedings{greydanus2019hamiltonian,
  author    = {Greydanus, S. and Dzamba, M. and Yosinski, J.},
  title     = {{Hamiltonian} neural networks},
  booktitle = {Advances in Neural Information Processing Systems (NeurIPS)},
  volume    = {32},
  year      = {2019},
  url       = {https://proceedings.neurips.cc/paper_files/paper/2019/hash/26cd8ecadce0d4efd6cc8a8725cbd1f8-Abstract.html}
}

@inproceedings{cranmer2020lagrangian,
  author    = {Cranmer, M. and Greydanus, S. and Hoyer, S. and Battaglia, P. and Spergel, D. and Ho, S.},
  title     = {{Lagrangian} neural networks},
  booktitle = {ICLR Workshop on Integration of Deep Neural Models and Differential Equations},
  year      = {2020},
  url       = {https://arxiv.org/abs/2003.04630}
}

@inproceedings{rubanova2022constraintbased,
  author    = {Rubanova, Y. and Sanchez-Gonzalez, A. and Pfaff, T. and Battaglia, P.},
  title     = {Constraint-based graph network simulator},
  booktitle = {International Conference on Machine Learning (ICML)},
  series    = {Proceedings of Machine Learning Research},
  volume    = {162},
  pages     = {18844--18870},
  year      = {2022},
  url       = {https://proceedings.mlr.press/v162/rubanova22a.html}
}

@inproceedings{allen2023learning,
  author    = {Allen, K.R. and Rubanova, Y. and Lopez-Guevara, T. and Whitney, W. and Sanchez-Gonzalez, A. and Battaglia, P. and Pfaff, T.},
  title     = {Learning rigid dynamics with face interaction graph networks},
  booktitle = {International Conference on Learning Representations (ICLR)},
  year      = {2023},
  url       = {https://openreview.net/forum?id=J7Uh781A05p}
}

@inproceedings{prantl2022guaranteed,
  author    = {Prantl, L. and Ummenhofer, B. and Koltun, V. and Thuerey, N.},
  title     = {Guaranteed conservation of momentum for learning particle-based fluid dynamics},
  booktitle = {Advances in Neural Information Processing Systems (NeurIPS)},
  volume    = {35},
  pages     = {6901--6913},
  year      = {2022},
  doi       = {10.52202/068431-0500}
}

@article{sharma2026physicsinformed,
  author  = {Sharma, V. and Fink, O.},
  title   = {A physics-informed graph neural network conserving linear and angular momentum for dynamical systems},
  journal = {Nature Communications},
  volume  = {17},
  pages   = {1045},
  year    = {2026},
  doi     = {10.1038/s41467-025-67802-5}
}

@article{zhao2025physicalinformationflowconstrained,
  author  = {Zhao, S. and Chen, H. and Zhao, J.},
  title   = {A physical-information-flow-constrained temporal graph neural network-based simulator for granular materials},
  journal = {Computer Methods in Applied Mechanics and Engineering},
  volume  = {433},
  pages   = {117536},
  year    = {2025},
  doi     = {10.1016/j.cma.2024.117536}
}

@misc{rossi2020temporal,
  author       = {Rossi, E. and Chamberlain, B. and Frasca, F. and Eynard, D. and Monti, F. and Bronstein, M.},
  title        = {Temporal graph networks for deep learning on dynamic graphs},
  howpublished = {arXiv:2006.10637},
  year         = {2020},
  doi          = {10.48550/arXiv.2006.10637}
}

@inproceedings{kumar2019predicting,
  author    = {Kumar, S. and Zhang, X. and Leskovec, J.},
  title     = {Predicting dynamic embedding trajectory in temporal interaction networks},
  booktitle = {ACM SIGKDD Conference on Knowledge Discovery and Data Mining (KDD)},
  pages     = {1269--1278},
  year      = {2019},
  doi       = {10.1145/3292500.3330895}
}

@inproceedings{trivedi2019dyrep,
  author    = {Trivedi, R. and Farajtabar, M. and Biswal, P. and Zha, H.},
  title     = {{DyRep}: learning representations over dynamic graphs},
  booktitle = {International Conference on Learning Representations (ICLR)},
  year      = {2019},
  url       = {https://openreview.net/forum?id=HyePrhR5KX}
}

@inproceedings{xu2020inductive,
  author    = {Xu, D. and Ruan, C. and Korpeoglu, E. and Kumar, S. and Achan, K.},
  title     = {Inductive representation learning on temporal graphs},
  booktitle = {International Conference on Learning Representations (ICLR)},
  year      = {2020},
  url       = {https://openreview.net/forum?id=rJeW1yHYwH}
}

@article{kazemi2020representation,
  author  = {Kazemi, S.M. and Goel, R. and Jain, K. and Kobyzev, I. and Sethi, A. and Forsyth, P. and Poupart, P.},
  title   = {Representation learning for dynamic graphs: a survey},
  journal = {Journal of Machine Learning Research},
  volume  = {21},
  pages   = {1--73},
  year    = {2020},
  url     = {https://jmlr.org/papers/v21/19-447.html}
}

@inproceedings{jain2016structuralrnn,
  author    = {Jain, A. and Zamir, A.R. and Savarese, S. and Saxena, A.},
  title     = {Structural-{RNN}: deep learning on spatio-temporal graphs},
  booktitle = {IEEE Conference on Computer Vision and Pattern Recognition (CVPR)},
  pages     = {5308--5317},
  year      = {2016},
  doi       = {10.1109/CVPR.2016.573}
}

@inproceedings{bengio2015scheduled,
  author    = {Bengio, S. and Vinyals, O. and Jaitly, N. and Shazeer, N.},
  title     = {Scheduled sampling for sequence prediction with recurrent neural networks},
  booktitle = {Advances in Neural Information Processing Systems (NeurIPS)},
  volume    = {28},
  pages     = {1171--1179},
  year      = {2015},
  url       = {https://proceedings.neurips.cc/paper_files/paper/2015/hash/e995f98d56967d946471af29d7bf99f1-Abstract.html}
}

@inproceedings{um2020solverintheloop,
  author    = {Um, K. and Brand, R. and Fei, Y. and Holl, P. and Thuerey, N.},
  title     = {Solver-in-the-loop: learning from differentiable physics to interact with iterative {PDE}-solvers},
  booktitle = {Advances in Neural Information Processing Systems (NeurIPS)},
  volume    = {33},
  pages     = {6111--6122},
  year      = {2020},
  url       = {https://proceedings.neurips.cc/paper_files/paper/2020/hash/43e4e6a6f341e00671e123714de019a8-Abstract.html}
}

@inproceedings{brandstetter2022message,
  author    = {Brandstetter, J. and Worrall, D. and Welling, M.},
  title     = {Message passing neural {PDE} solvers},
  booktitle = {International Conference on Learning Representations (ICLR)},
  year      = {2022},
  url       = {https://openreview.net/forum?id=vSix3HPYKSU}
}

@inproceedings{stachenfeld2022learned,
  author    = {Stachenfeld, K. and Fielding, D.B. and Kochkov, D. and Cranmer, M. and Pfaff, T. and Godwin, J. and Cui, C. and Ho, S. and Battaglia, P. and Sanchez-Gonzalez, A.},
  title     = {Learned simulators for turbulence},
  booktitle = {International Conference on Learning Representations (ICLR)},
  year      = {2022},
  url       = {https://openreview.net/forum?id=msRBojTz-Nh}
}

@article{mindlin1953elastic,
  author  = {Mindlin, R.D. and Deresiewicz, H.},
  title   = {Elastic spheres in contact under varying oblique forces},
  journal = {Journal of Applied Mechanics},
  volume  = {20},
  pages   = {327--344},
  year    = {1953},
  doi     = {10.1115/1.4010702}
}

@article{luding2008cohesive,
  author  = {Luding, S.},
  title   = {Cohesive, frictional powders: contact models for tension},
  journal = {Granular Matter},
  volume  = {10},
  pages   = {235--246},
  year    = {2008},
  doi     = {10.1007/s10035-008-0099-x}
}

@article{radjai1996force,
  author  = {Radjai, F. and Jean, M. and Moreau, J.-J. and Roux, S.},
  title   = {Force distributions in dense two-dimensional granular systems},
  journal = {Physical Review Letters},
  volume  = {77},
  pages   = {274--277},
  year    = {1996},
  doi     = {10.1103/PhysRevLett.77.274}
}

@article{majmudar2005contact,
  author  = {Majmudar, T.S. and Behringer, R.P.},
  title   = {Contact force measurements and stress-induced anisotropy in granular materials},
  journal = {Nature},
  volume  = {435},
  pages   = {1079--1082},
  year    = {2005},
  doi     = {10.1038/nature03805}
}

@article{anon2004dense,
  author  = {{GDR MiDi}},
  title   = {On dense granular flows},
  journal = {The European Physical Journal E},
  volume  = {14},
  pages   = {341--365},
  year    = {2004},
  doi     = {10.1140/epje/i2003-10153-0}
}

@article{jop2006constitutive,
  author  = {Jop, P. and Forterre, Y. and Pouliquen, O.},
  title   = {A constitutive law for dense granular flows},
  journal = {Nature},
  volume  = {441},
  pages   = {727--730},
  year    = {2006},
  doi     = {10.1038/nature04801}
}

@book{schofield1968critical,
  author    = {Schofield, A.N. and Wroth, C.P.},
  title     = {Critical state soil mechanics},
  publisher = {McGraw-Hill},
  address   = {London},
  year      = {1968}
}

@article{hochreiter1997long,
  author  = {Hochreiter, S. and Schmidhuber, J.},
  title   = {Long short-term memory},
  journal = {Neural Computation},
  volume  = {9},
  pages   = {1735--1780},
  year    = {1997},
  doi     = {10.1162/neco.1997.9.8.1735}
}

@article{wang2018multiscale,
  author  = {Wang, K. and Sun, W.},
  title   = {A multiscale multi-permeability poroplasticity model linked by recursive homogenizations and deep learning},
  journal = {Computer Methods in Applied Mechanics and Engineering},
  volume  = {334},
  pages   = {337--380},
  year    = {2018},
  doi     = {10.1016/j.cma.2018.01.036}
}

@article{zhang2021application,
  author  = {Zhang, N. and Shen, S.L. and Zhou, A. and Jin, Y.F.},
  title   = {Application of {LSTM} approach for modelling stress--strain behaviour of soil},
  journal = {Applied Soft Computing},
  volume  = {100},
  pages   = {106959},
  year    = {2021},
  doi     = {10.1016/j.asoc.2020.106959}
}

@article{qu2021towards,
  author  = {Qu, T. and Di, S. and Feng, Y.T. and Wang, M. and Zhao, T.},
  title   = {Towards data-driven constitutive modelling for granular materials via micromechanics-informed deep learning},
  journal = {International Journal of Plasticity},
  volume  = {144},
  pages   = {103046},
  year    = {2021},
  doi     = {10.1016/j.ijplas.2021.103046}
}

@article{karapiperis2021datadriven,
  author  = {Karapiperis, K. and Stainier, L. and Ortiz, M. and Andrade, J.E.},
  title   = {Data-driven multiscale modeling in mechanics},
  journal = {Journal of the Mechanics and Physics of Solids},
  volume  = {147},
  pages   = {104239},
  year    = {2021},
  doi     = {10.1016/j.jmps.2020.104239}
}

@inproceedings{cho2014learning,
  author    = {Cho, K. and van Merri\"enboer, B. and Gulcehre, C. and Bahdanau, D. and Bougares, F. and Schwenk, H. and Bengio, Y.},
  title     = {Learning phrase representations using {RNN} encoder--decoder for statistical machine translation},
  booktitle = {Conference on Empirical Methods in Natural Language Processing (EMNLP)},
  pages     = {1724--1734},
  year      = {2014},
  doi       = {10.3115/v1/D14-1179}
}

@inproceedings{vaswani2017attention,
  author    = {Vaswani, A. and Shazeer, N. and Parmar, N. and Uszkoreit, J. and Jones, L. and Gomez, A.N. and Kaiser, {\L}. and Polosukhin, I.},
  title     = {Attention is all you need},
  booktitle = {Advances in Neural Information Processing Systems (NeurIPS)},
  volume    = {30},
  pages     = {5998--6008},
  year      = {2017},
  url       = {https://proceedings.neurips.cc/paper_files/paper/2017/hash/3f5ee243547dee91fbd053c1c4a845aa-Abstract.html}
}

@inproceedings{velikovi2018graph,
  author    = {Veli\v{c}kovi\'c, P. and Cucurull, G. and Casanova, A. and Romero, A. and Li\`o, P. and Bengio, Y.},
  title     = {Graph attention networks},
  booktitle = {International Conference on Learning Representations (ICLR)},
  year      = {2018},
  url       = {https://openreview.net/forum?id=rJXMpikCZ}
}

@article{kumar2023gnsjoss,
  author  = {Kumar, K. and Vantassel, J.},
  title   = {{GNS}: a generalizable graph neural network-based simulator for particulate and fluid modeling},
  journal = {Journal of Open Source Software},
  volume  = {8},
  pages   = {5025},
  year    = {2023},
  doi     = {10.21105/joss.05025}
}

@inproceedings{kumar2023differentiable,
  author    = {Kumar, K. and Choi, Y.},
  title     = {Accelerating particle and fluid simulations with differentiable graph networks for solving forward and inverse problems},
  booktitle = {Workshops of the International Conference for High Performance Computing, Networking, Storage, and Analysis (SC-W)},
  pages     = {60--65},
  year      = {2023},
  doi       = {10.1145/3624062.3626082}
}

@article{choi2024surrogate,
  author  = {Choi, Y. and Kumar, K.},
  title   = {Graph neural network-based surrogate model for granular flows},
  journal = {Computers and Geotechnics},
  volume  = {166},
  pages   = {106015},
  year    = {2024},
  doi     = {10.1016/j.compgeo.2023.106015}
}

@inproceedings{choi2024threedimensional,
  author    = {Choi, Y. and Kumar, K.},
  title     = {Three-dimensional granular flow simulation using graph neural network-based learned simulator},
  booktitle = {Geo-Congress 2024 (ASCE)},
  pages     = {335--344},
  year      = {2024},
  doi       = {10.1061/9780784485347.034}
}

@article{choi2024inverse,
  author  = {Choi, Y. and Kumar, K.},
  title   = {Inverse analysis of granular flows using differentiable graph neural network simulator},
  journal = {Computers and Geotechnics},
  volume  = {171},
  pages   = {106374},
  year    = {2024},
  doi     = {10.1016/j.compgeo.2024.106374}
}

@article{choi2026multilayer,
  author  = {Choi, Y. and Macedo, J. and Liu, C.},
  title   = {Differentiable graph neural network simulator for forward and inverse modeling of multi-layered slope system with multiple material properties},
  journal = {Soils and Foundations},
  volume  = {66},
  pages   = {101773},
  year    = {2026},
  doi     = {10.1016/j.sandf.2026.101773}
}

@article{hsiao2025nerf,
  author  = {Hsiao, C.-H. and Kumar, K.},
  title   = {From images to properties: a {NeRF}-driven framework for granular material parameter inversion},
  journal = {EPJ Web of Conferences},
  volume  = {340},
  pages   = {10017},
  year    = {2025},
  doi     = {10.1051/epjconf/202534010017}
}

@misc{choi2026postliquefaction,
  author       = {Choi, Y. and Macedo, J.},
  title        = {Differentiable graph neural network simulator for the back-analysis of post-liquefaction residual strength from flow failure runout},
  howpublished = {arXiv:2602.11621},
  year         = {2026},
  doi          = {10.48550/arXiv.2602.11621}
}

@misc{mayr2021learning,
  author       = {Mayr, A. and Lehner, S. and Mayrhofer, A. and Kloss, C. and Hochreiter, S. and Brandstetter, J.},
  title        = {Learning {3D} granular flow simulations},
  howpublished = {arXiv:2105.01636},
  year         = {2021},
  doi          = {10.48550/arXiv.2105.01636}
}

@inproceedings{mayr2023bgnn,
  author    = {Mayr, A. and Lehner, S. and Mayrhofer, A. and Kloss, C. and Hochreiter, S. and Brandstetter, J.},
  title     = {Boundary graph neural networks for {3D} simulations},
  booktitle = {AAAI Conference on Artificial Intelligence (AAAI)},
  volume    = {37},
  pages     = {9099--9107},
  year      = {2023},
  doi       = {10.1609/aaai.v37i8.26092}
}

@article{li2024sgn,
  author  = {Li, S. and Sakai, M.},
  title   = {Advanced graph neural network-based surrogate model for granular flows in arbitrarily shaped domains},
  journal = {Chemical Engineering Journal},
  volume  = {500},
  pages   = {157349},
  year    = {2024},
  doi     = {10.1016/j.cej.2024.157349}
}

@article{alkin2024neuraldem,
  author  = {Alkin, B. and Kronlachner, T. and Papa, S. and Pirker, S. and Lichtenegger, T. and Brandstetter, J.},
  title   = {{NeuralDEM} for real time simulations of industrial particular flows},
  journal = {Communications Physics},
  volume  = {8},
  pages   = {440},
  year    = {2025},
  doi     = {10.1038/s42005-025-02342-4}
}

@article{jiang2024inversedesign,
  author  = {Jiang, Y. and Byrne, E. and Glassey, J. and Chen, X.},
  title   = {Integrating graph neural network-based surrogate modeling with inverse design for granular flows},
  journal = {Industrial \& Engineering Chemistry Research},
  volume  = {63},
  pages   = {9225--9235},
  year    = {2024},
  doi     = {10.1021/acs.iecr.4c00692}
}

@article{li2025graphdem,
  author  = {Li, B. and Du, B. and Liu, K. and Cheng, K. and Ye, J. and Feng, J. and Cui, X.},
  title   = {Graph-{DEM}: a graph neural network model for proxy and acceleration discrete element method},
  journal = {Applied Sciences},
  volume  = {15},
  pages   = {10432},
  year    = {2025},
  doi     = {10.3390/app151910432}
}

@article{garridonunez2026ballmill,
  author  = {Garrido Nu\~{n}ez, S. and Schott, D.L. and Padding, J.T.},
  title   = {Accelerating granular dynamics simulations: a graph neural network surrogate for complex high-energy ball milling},
  journal = {Powder Technology},
  volume  = {468},
  pages   = {121653},
  year    = {2026},
  doi     = {10.1016/j.powtec.2025.121653}
}

@article{le2026cconvdem,
  author  = {Le, D. and Delaney, G.W. and Nguyen, L. and Phung, T. and Howard, D. and Kahandawa, G. and Murshed, M.},
  title   = {A neural network surrogate for modelling granular flow dynamics in industrial applications with dynamic boundary conditions},
  journal = {Powder Technology},
  volume  = {476},
  pages   = {122258},
  year    = {2026},
  doi     = {10.1016/j.powtec.2026.122258}
}

@article{mandal2022forcechain,
  author  = {Mandal, R. and Casert, C. and Sollich, P.},
  title   = {Robust prediction of force chains in jammed solids using graph neural networks},
  journal = {Nature Communications},
  volume  = {13},
  pages   = {4424},
  year    = {2022},
  doi     = {10.1038/s41467-022-31732-3}
}

@article{li2023contactforce,
  author  = {Li, Z. and Li, X. and Zhang, H. and Huang, D. and Zhang, L.},
  title   = {The prediction of contact force networks in granular materials based on graph neural networks},
  journal = {The Journal of Chemical Physics},
  volume  = {158},
  pages   = {054905},
  year    = {2023},
  doi     = {10.1063/5.0122695}
}

@article{aminimajd2024suspension,
  author  = {Aminimajd, A. and Maia, J. and Singh, A.},
  title   = {Scalability of a graph neural network in accurate prediction of frictional contact networks in suspensions},
  journal = {Soft Matter},
  volume  = {21},
  pages   = {2826--2835},
  year    = {2025},
  doi     = {10.1039/D4SM01391C}
}

@article{aminimajd2025nearjamming,
  author  = {Aminimajd, A. and Maia, J. and Singh, A.},
  title   = {Robust prediction of frictional contact network in near-jamming suspensions employing deep graph neural networks},
  journal = {Physics of Fluids},
  volume  = {37},
  pages   = {073306},
  year    = {2025},
  doi     = {10.1063/5.0267708}
}

@article{haeri2024subspace,
  author  = {Haeri, A. and Holz, D. and Skonieczny, K.},
  title   = {Subspace graph networks for real-time granular flow simulation with applications to machine-terrain interactions},
  journal = {Engineering Applications of Artificial Intelligence},
  volume  = {135},
  pages   = {108765},
  year    = {2024},
  doi     = {10.1016/j.engappai.2024.108765}
}

@article{tuomainen2022manipulation,
  author  = {Tuomainen, N. and Blanco-Mulero, D. and Kyrki, V.},
  title   = {Manipulation of granular materials by learning particle interactions},
  journal = {IEEE Robotics and Automation Letters},
  volume  = {7},
  pages   = {5663--5670},
  year    = {2022},
  doi     = {10.1109/LRA.2022.3158382}
}

@inproceedings{wang2023pile,
  author    = {Wang, Y. and Li, Y. and Driggs-Campbell, K. and Fei-Fei, L. and Wu, J.},
  title     = {Dynamic-resolution model learning for object pile manipulation},
  booktitle = {Robotics: Science and Systems (RSS)},
  year      = {2023},
  doi       = {10.15607/RSS.2023.XIX.047}
}

@inproceedings{zhang2024adaptigraph,
  author    = {Zhang, K. and Li, B. and Hauser, K. and Li, Y.},
  title     = {{AdaptiGraph}: material-adaptive graph-based neural dynamics for robotic manipulation},
  booktitle = {Robotics: Science and Systems (RSS)},
  year      = {2024},
  doi       = {10.15607/RSS.2024.XX.010}
}

@inproceedings{orsula2025sim2dust,
  author    = {Orsula, A. and Geist, M. and Olivares-Mendez, M. and Martinez, C.},
  title     = {{Sim2Dust}: mastering dynamic waypoint tracking on granular media},
  booktitle = {International Conference on Space Robotics (iSpaRo)},
  pages     = {336--342},
  year      = {2025},
  doi       = {10.1109/iSpaRo66239.2025.11436990}
}

@inproceedings{cao2024neuma,
  author    = {Cao, J. and Guan, S. and Ge, Y. and Li, W. and Yang, X. and Ma, C.},
  title     = {{NeuMA}: neural material adaptor for visual grounding of intrinsic dynamics},
  booktitle = {Advances in Neural Information Processing Systems (NeurIPS)},
  volume    = {37},
  pages     = {65643--65669},
  year      = {2024},
  doi       = {10.52202/079017-2097}
}

@inproceedings{cai2024gic,
  author    = {Cai, J. and Yang, Y. and Yuan, W. and He, Y. and Dong, Z. and Bo, L. and Cheng, H. and Chen, Q.},
  title     = {{GIC}: {Gaussian}-informed continuum for physical property identification and simulation},
  booktitle = {Advances in Neural Information Processing Systems (NeurIPS)},
  volume    = {37},
  pages     = {75035--75063},
  year      = {2024},
  doi       = {10.52202/079017-2388}
}

@misc{manoharan2025filmgns,
  author       = {Manoharan, N.R. and Iqbal, H. and Kumar, K.},
  title        = {Parameter-efficient conditioning for material generalization in graph-based simulators},
  howpublished = {arXiv:2511.05456},
  year         = {2025},
  doi          = {10.48550/arXiv.2511.05456}
}

@misc{wang2026worldparticle,
  author       = {Wang, C. and Guo, M. and Chen, S. and Zhang, H. and Wang, M. and Ni, X. and Sun, H. and Wang, K. and Pan, Z. and Wu, K. and Liu, L. and Yang, Y. and Jiang, C. and Komura, T. and Matusik, W. and Chen, P.Y.},
  title        = {{WorldParticle}: unified world simulation of {Lagrangian} particle dynamics via transformer},
  howpublished = {arXiv:2605.15305},
  year         = {2026},
  doi          = {10.48550/arXiv.2605.15305}
}

\end{document}